\documentclass{article}

\usepackage[preprint]{neurips_2023}

\usepackage[utf8]{inputenc} 
\usepackage[T1]{fontenc}    
\usepackage{hyperref}       
\usepackage{url}            
\usepackage{booktabs}       
\usepackage{amsfonts}       
\usepackage{nicefrac}       
\usepackage{microtype}      
\usepackage{xcolor}         
\usepackage{graphicx} 

\usepackage{amsmath}
\usepackage{amsfonts}
\usepackage{bbm}
\usepackage{lineno}
\usepackage{upgreek}
\usepackage{amssymb}
\usepackage{epsfig}
\usepackage{dsfont}
\usepackage{indentfirst}
\usepackage{mathrsfs}
\usepackage{url}
\usepackage{authblk}
\usepackage{longtable}
\usepackage{xparse} 
\usepackage{bm}
\usepackage{mlmath}
\usepackage{amsthm}
\usepackage{cleveref}
\usepackage{subcaption}
\usepackage{ulem,xpatch}
\newtheorem{innercustomgeneric}{\customgenericname}
\providecommand{\customgenericname}{}

\newtheorem{lemma}{Lemma}
\newtheorem{definition}{Definition}
\newtheorem{remark}{Remark}
\newtheorem{assumption}{Assumption}

\newtheorem{proposition}{Proposition}
\newtheorem{theorem}{Theorem}

\title{Understanding the Initial Condensation of Convolutional Neural Networks}

\author{
Zhangchen Zhou\textsuperscript{\rm 1,2}, 
Hanxu Zhou\textsuperscript{\rm 1}\thanks{Corresponding author: zhouhanxu@sjtu.edu.cn}, 
Yuqing Li\textsuperscript{\rm 1,3}\thanks{Corresponding author: liyuqing$\_$551@sjtu.edu.cn.}, 
Zhi-Qin John Xu\textsuperscript{\rm 1,4,5}\thanks{Corresponding author: xuzhiqin@sjtu.edu.cn.}, 
\\
\textsuperscript{\rm 1} School of Mathematical Sciences, Shanghai Jiao Tong
University \\ 
\textsuperscript{\rm 2} Zhiyuan College, Shanghai Jiao Tong University \\
\textsuperscript{\rm 3} CMA-Shanghai, Shanghai Jiao Tong University \\
\textsuperscript{\rm 4} Institute of Natural Sciences, MOE-LSC, Shanghai Jiao Tong University \\
\textsuperscript{\rm 5} Qing Yuan Research Institute, Shanghai Jiao Tong University \\

}

\begin{document}

\maketitle

\begin{abstract}
Previous research has shown that fully-connected networks with small initialization and gradient-based training methods exhibit a phenomenon known as condensation during training. This phenomenon refers to the input weights of hidden neurons condensing into isolated orientations during training, revealing an implicit bias towards simple solutions in the parameter space. However, the impact of neural network structure on condensation has not been investigated yet. In this study, we focus on the investigation of convolutional neural networks (CNNs). Our experiments suggest that when subjected to small initialization and gradient-based training methods, kernel weights within the same CNN layer also cluster together during training, demonstrating a significant degree of condensation. Theoretically, we demonstrate that in a finite training period, kernels of a two-layer CNN with small initialization will converge to one or a few directions. This work represents a step towards a better understanding of the non-linear training behavior exhibited by neural networks with specialized structures.
\end{abstract}

\section{Introduction}
\begin{figure}[h]
	\centering
	\includegraphics[width=0.90\textwidth]{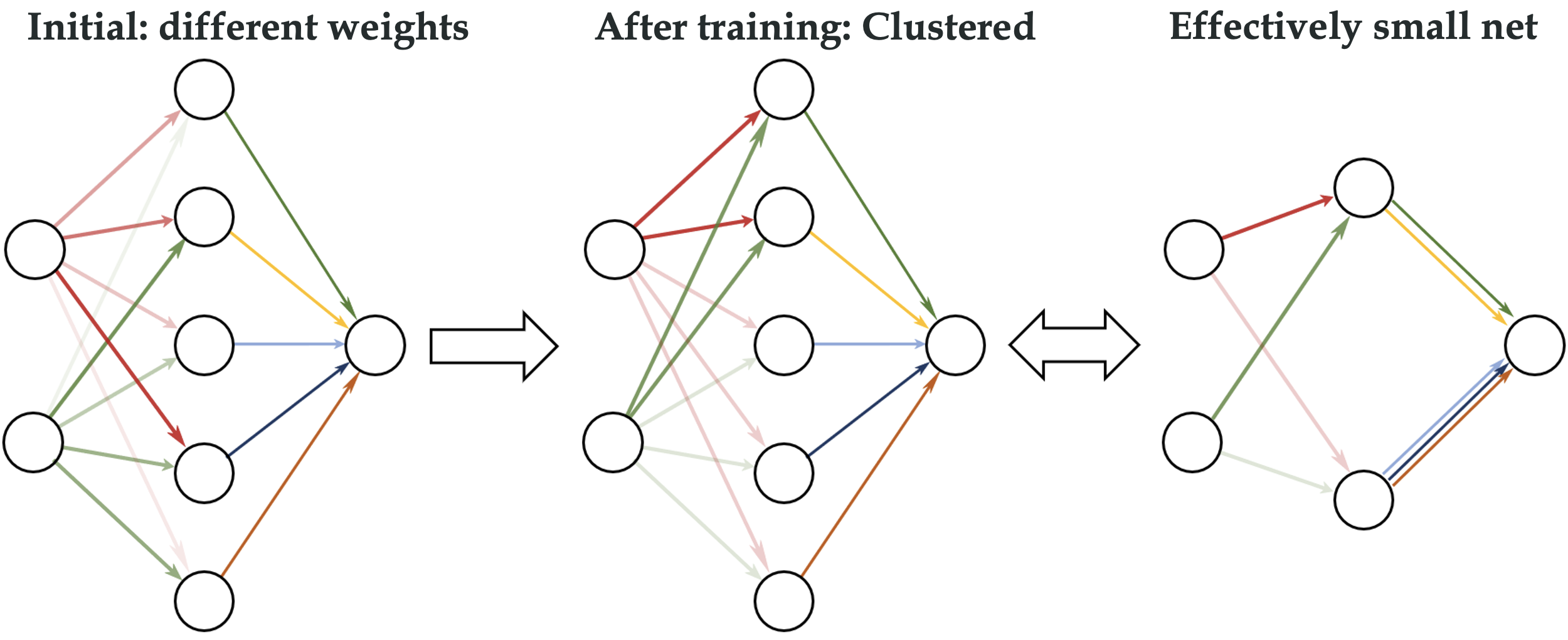}
	\caption{Illustration of condensation. The color and its intensity of a line indicate the strength of the weight. Initially, weights are random. Soon after training, the weights from an input node to all hidden neurons are clustered into two groups, i.e., condensation. Multiple hidden neurons can be replaced by an effective neuron with low complexity, which has the same input weight as original hidden neurons and the same output weight as the summation of all output weights of original hidden neurons.   \label{fig:condensetoy}}
	
\end{figure}

As large neural networks continue to demonstrate impressive performance in numerous practical tasks, a key challenge has come to understand the reasons behind the strong generalization capabilities often exhibited by over-parameterized networks \citep{breiman1995reflections,zhang2021understanding}. A commonly employed approach to understanding neural networks is to examine their implicit biases during the training process. Several studies have shown that neural networks tend to favor simple solutions. For instance, from a Fourier perspective, neural networks have a bias toward low-frequency functions, which is known as the frequency principle \citep{xu_training_2018,xu2019frequency} or spectral bias \citep{rahaman2018spectral}. In the parameter space, \cite{luo2021phase} observed a condensation phenomenon, i.e., the input weights of hidden neurons in two-layer $\rm{ReLU}$ neural networks  condense into isolated orientations during training in the non-linear regime, particularly with small initialization. Fig. \ref{fig:condensetoy} presents an illustrative example in which a large condensed network can be reduced to an effective smaller network with only two neurons. Based on complexity theory \citep{bartlett2002rademacher}, as the condensation phenomenon  reduces the network complexity, it might provide  insights into how over-parameterized neural networks achieve good generalization performance in practice. \cite{zhang2022implicit} drew inspiration from this phenomenon and found that dropout \citep{hinton2012improving,srivastava2014dropout}, a commonly used optimization technique for improving generalization, exhibits an implicit bias towards condensation through experiments and theory. Prior literature has predominantly centered on the study of fully-connected neural networks, thereby leaving the emergence and properties of the condensation phenomenon in neural networks with different structural characteristics inadequately understood.  Consequently, this paper aims to   investigate the occurrence of condensation in convolutional neural networks (CNNs).

The success of deep learning relies heavily on the structures used, such as convolution and attention. Convolution is an ideal starting point for investigating the impact of structure on learning outcomes, as it is widely used and has simple structure. To achieve a clear condensation phenomenon in CNNs, we adopt a strategy of initializing weights with small values. Small weight initialization can result in rich non-linearity of neural network (NN) training behavior \citep{mei2019mean, rotskoff2018parameters, chizat2018global, sirignano2018mean}. Over-parameterized NNs with small initialization can, for instance, achieve low generalization error \citep{advani2020high} and converge to a solution with maximum margin \citep{phuong2020inductive}. In contrast to the condensation in fully-connected networks, each kernel in CNNs is considered as a unit, and condensation is referred to the behavior, that a set of kernels in the same layer evolves towards the same direction. 

Understanding the initial condensation can benefit understanding subsequent training stages \citep{fort2020deep, hu2020surprising, luo2021phase, jiang2019fantastic, li2018visualizing}.
Previous research has shown how neural networks with small initialization can condense during the initial training stage in fully-connected networks \citep{maennel2018gradient,pellegrini2020analytic,zhou2022towards,chen2023phase}. This work aims to demonstrate the initial condensation in CNNs during training. A major advantage of studying the initial training stage of neural networks with small initialization is that the network can be approximated accurately by the leading-order Taylor expansion at zero weights. Further, the structure of CNNs may cause kernels in the same layer to exhibit similar dynamics. Through theoretical proof, we show that CNNs can condense into one or a few directions within a finite training period with small initialization. This initial condensation serves an important role in resetting the neural network of different initializations to a similar and simple state, thus reducing the sensitivity of initialization and facilitating the tuning of hyper-parameters of initialization.

\section{Related works}
For fully-connected neural networks, it has been generally studied that different initializations can lead to very different training behavior regimes \citep{luo2021phase}, including linear regime (similar to the lazy regime) ~\citep{jacot2018neural,arora2019exact,zhang2019type,weinan2020comparative,chizat2019note}, critical regime \citep{mei2019mean,rotskoff2018parameters,chizat2018global,sirignano2018mean} and condensed regime (non-linear regime). 
The relative change of input weights as the width approaches infinity is a critical parameter that distinguishes the different regimes, namely $0$, $O(1)$, and $+\infty$.  
\cite{zhou2022empirical} demonstrated that these regimes also exist for three-layer $\mathrm{ReLU}$ neural networks with infinite width.
Experiments suggest that condensation is a frequent occurrence in the non-linear regime.

\cite{zhang2021embedding,zhang2021hierarchical}discovered an embedding principle in loss landscapes between narrow and wide neural networks, based on condensation. This principle suggests that the loss landscape of a deep neural network (DNN) includes all critical points of narrower DNNs, which is also studied in \cite{fukumizu2000local,fukumizu2019semi,simsek2021geometry}.
The embedding structure indicates the existence of global minima where condensation can occur.
\cite{zhang2022linear} has demonstrated that NNs exhibiting condensation can achieve the desired function with a substantially lower number of samples compared to the number of parameters. However, these studies fail to demonstrate the training process's role in causing condensation.

CNN is one of the fundamental structures in deep learning \citep{gu2018recent}.
\cite{he2016deep} introduced the use of residual connections for training deep CNNs, which has greatly enhanced the performance of CNNs on complex practical tasks.
Recently, there are also many theoretical advances.  \cite{zhou2020universality} shows that CNN can be used to approximate any continuous function to an arbitrary accuracy when the depth of the neural network is large enough. \cite{arora2019exact} exactly compute the neural tangent kernel of CNN. 
Provided that the signal-to-noise ratio satisfies certain conditions, \cite{cao2022benign} have demonstrated that a two-layer CNN, trained through gradient descent, can obtain negligible training and test losses. 
In this work, we focus on the training process of CNNs.

\section{Preliminaries}
\subsection{Some Notations}\label{subsection...Notations-main}
For a matrix $\mathbf{A}$, we use $\mathbf{A}_{i, j}$ to denote its $(i, j)$-th entry. We  also use $\mathbf{A}_{i,:}$ to denote the $i$-th row vector of $\mathbf{A}$ and define $\mathbf{A}_{i, j: k}:=\left(\mathbf{A}_{i, j}, \mathbf{A}_{i, j+1}, \cdots, \mathbf{A}_{i, k}\right)$ as part of the vector. Similarly $\mathbf{A}_{:, k}$ is the $k$-th column vector and $\mathbf{A}_{i: j, k}:=\left(\mathbf{A}_{i, k}, \mathbf{A}_{i+1, k}, \cdots, \mathbf{A}_{j, k}\right)^\T$ is   part of  the $k$-th column vector. 

We let $[n]=\{1,2, \ldots, n\}$. We set   $\fN(\vmu, \Sigma)$ as the normal distribution with mean $\vmu$ and covariance $\Sigma$. We set a special vector $\mathbbm{1}:=(1,1,1,\dots,1)^\T$, whose dimension varies. For a vector $\mathbf{v}$, we use $\|\mathbf{v}\|_2$ to denote its Euclidean norm, and we use $\left<\cdot,\cdot\right>$ to denote the standard inner product between two vectors. Finally, for a matrix $\mathbf{A}$, we use   $\|\mathbf{A}\|_{2\to 2}$ to denote its operator norm.  
\subsection{Problem Setup}
We focus on the empirical risk minimization problem given by the quadratic loss:
\begin{equation}\label{eq...text...Intro...LossFunction-main}
\min_{\vtheta}R_S(\vtheta)=\frac{1}{2n}\sum_{i=1}^n\left({f(\vx_i,\vtheta)-y_i}\right)^2.
\end{equation}
In the above, $n$ is the total number of training samples,  $\{ \vx_i\}_{i=1}^n$ are the training inputs, $\{ y_i\}_{i=1}^n$ are the labels,  $f(\vx_i,\vtheta)$ is the prediction function, and $\vtheta$ are the parameters to be optimized, which is modeled by a $(L+1)$-layer CNN with filter size $m\times m$. We denote $\vx^{[l]}(i)$ as the output of the $l$-th layer with respect to the $i$-th sample for $l\geq 1$, and $\vx^{[0]}(i):=\vx_i$ is the $i$-th training input. For any $l\in[0:L]$, we denote the size of width, height, channel of $\vx^{[l]}$ as $W_l,~H_l,$ and $C_l$, respectively, i.e., $\{\vx^{[l]}(i)\}_{i=1}^n\subset\sR^{W_l \times H_l\times C_l}$. 
We introduce a filter operator $\chi(\cdot,\cdot)$, which maps the width and height indices of the output of all layers to a binary variable,    i.e.,  for a filter of size $m\times m$, the filter operator reads 
\begin{equation}\label{eq...text...Intro...FilterOperator-main}
    \chi(p, q)=\left\{\begin{array}{l}1,\ \ \text{for~~} 0\leqslant p, q \leqslant m-1 
    \\ 0, \ \ \text{otherwise,} \end{array}\right. 
\end{equation}
then the $(L+1)$-layer CNN with filter size $m\times m$ is recursively defined for  $l\in[2:L]$,
\begin{align*}
\vx_{u, v, \beta}^{[1]}&:=\left[\sum_{\alpha=1}^{C_{0}}\left(\sum_{p=-\infty}^{+\infty} \sum_{q=-\infty}^{+\infty} \vx^{[0]}_{u+p, v+q, \alpha} \cdot \mW_{p,q,\alpha,\beta}^{[1]} \cdot \chi(p, q)\right)\right]+\vb_{\beta}^{[1]}, \\
\vx_{u, v, \beta}^{[l]}&:=\left[\sum_{\alpha=1}^{C_{l-1}}\left(\sum_{p=-\infty}^{+\infty} \sum_{q=-\infty}^{\infty} \sigma\left(\vx_{u+p, v+q, \alpha}^{[l-1]} \right) \cdot \mW_{p,q,\alpha,\beta}^{[l]} \cdot \chi(p, q) \right)\right]+\vb_{\beta}^{[l]}, \\
f(\vx,\vtheta)& :=f_{\mathrm{CNN}}(\vx,\vtheta):=\left<\va, \sigma\left(\vx^{[L]}\right)\right>=\sum_{\beta=1}^{C_L}\sum_{u=1}^{W_L}\sum_{v=1}^{H_L} \va_{u, v, \beta} \cdot\sigma\left( \vx_{u, v, \beta}^{[L]}\right),
\end{align*} 
where $\sigma(\cdot)$ is the activation function applied coordinate-wisely to its input, and for each layer $l\in[L]$, all parameters belonging to this layer are initialized by: For $p,q\in[m-1]$,  $\alpha\in[C_{l-1}]$ and $\beta\in[C_l]$,
\begin{equation}
\mW_{p,q,\alpha,\beta}^{[l]}\sim\fN(0, \beta_1^2),\quad  \vb_{\beta}^{[l]}  \sim\fN(0, \beta_1^2).
\end{equation}
Note that for a pair of $\alpha$ and $\beta$, $\mW^{[l]}_{\cdot,\cdot,\alpha,\beta}$ is called a kernel.
Moreover, for $u\in[W_L]$ and $v\in[H_L]$,
\begin{equation}
\quad \va_{u, v, \beta}\sim\fN(0, \beta_2^2),
\end{equation}
 and for convenience in theory, we set  
$\beta_1=\beta_2=\eps,$ 
where $\eps>0$ is the scaling parameter. 

\textbf{Cosine similarity:} The cosine similarity between two vectors $\vu_1$ and $\vu_2$ is defined as
\begin{equation}
    D(\vu_1,\vu_2) = \frac{\vu_1^\T\vu_2}{(\vu_1^\T\vu_1)^{1/2}(\vu_2^{\T}\vu_2)^{1/2}}.
\end{equation}
We remark that  in order to compute the cosine similarity between two kernels,   each kernel $\mW^{[l]}_{\cdot,\cdot,\alpha,\beta}$ shall be vectorized.

In  the theoretical part, as we consider two-layer CNNs~($L=1$), thus  the upper case $[l]$ can be omitted since the number of weight vectors is equal to $1$, i.e., $\mW^{}_{\cdot,\cdot,\alpha,\beta}:=\mW^{[1]}_{\cdot,\cdot,\alpha,\beta}$.
We denote $M:=C_1$, the number of channels in $\vx^{[1]}(i)$,  which can be heuristically understood as the `width' of the hidden layer in the case of two-layer neural networks~(NNs). 

In experiments, the parameters are trained by either Adam or gradient descent~(GD), while in theory, only GD is used.
In the following, we demonstrate that kernels in multi-layer CNNs exhibit clear condensation in experiments during the initial training stage, with small initialization. We then provide theoretical evidence for two-layer CNNs.

\section{Condensation of the convolution kernels in experiments}

In this section, we will show condensation of convolution kernels in image datasets using different activation functions. 

\subsection{Experimental setup} 

For the CIFAR10 dataset: $500$ samples are randomly selected from CIFAR10 dataset for training.  We use the CNN with the structure: $n\times32C\text{-}(1024)\text{-}d$. The output dimension $d=10 \ \text{or} \ 1$ is used to classify the pictures or for regression respectively. The parameters of the convolution layer is initialized by the  $\mathcal{N}(0,\sigma_1^2)$, and the parameters of the linear layer is  $\mathcal{N}(0,\sigma_2^2)$. $\sigma_1$ is given by $(\frac{(c_{in}+c_{out})*m^2}{2})^{-\gamma}$  where $c_{in}$ and $c_{out}$ are the number of in channels and out channels respectively, $\sigma_2$ is given empirically by $0.0001$. The training method is GD or Adam with full batch. The training loss of each experiment is shown in Fig.\ref{Fig.loss} in Appendix.

\subsection{CIFAR10 examples}
\begin{figure}
     \centering
     \begin{subfigure}[b]{0.3\textwidth}
         \centering
         \includegraphics[width=\textwidth]{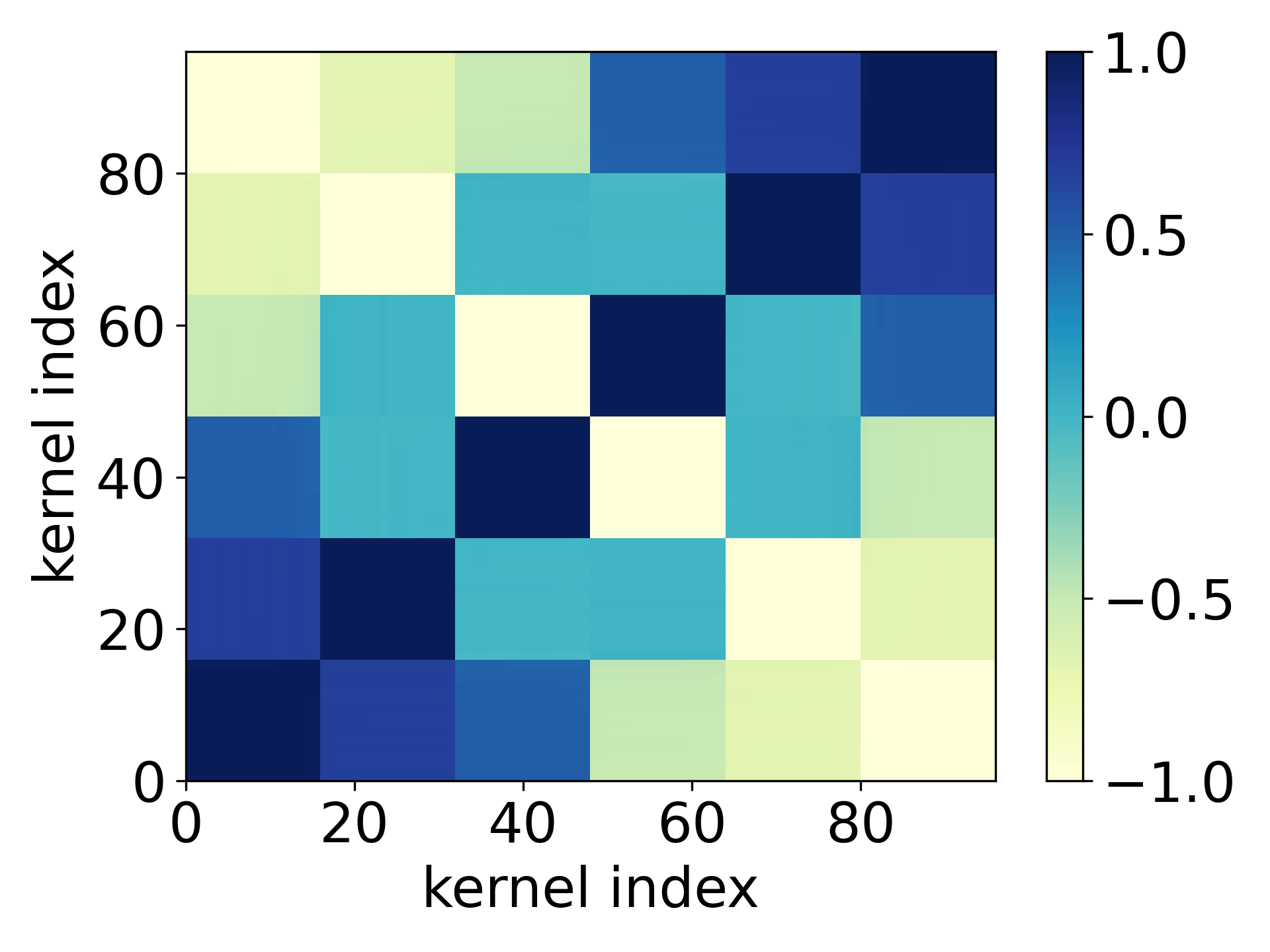}
         \caption{ layer $1$}
         \label{tanha1lccrossfinal}
     \end{subfigure}
     \hfill
     \begin{subfigure}[b]{0.3\textwidth}
         \centering
         \includegraphics[width=\textwidth]{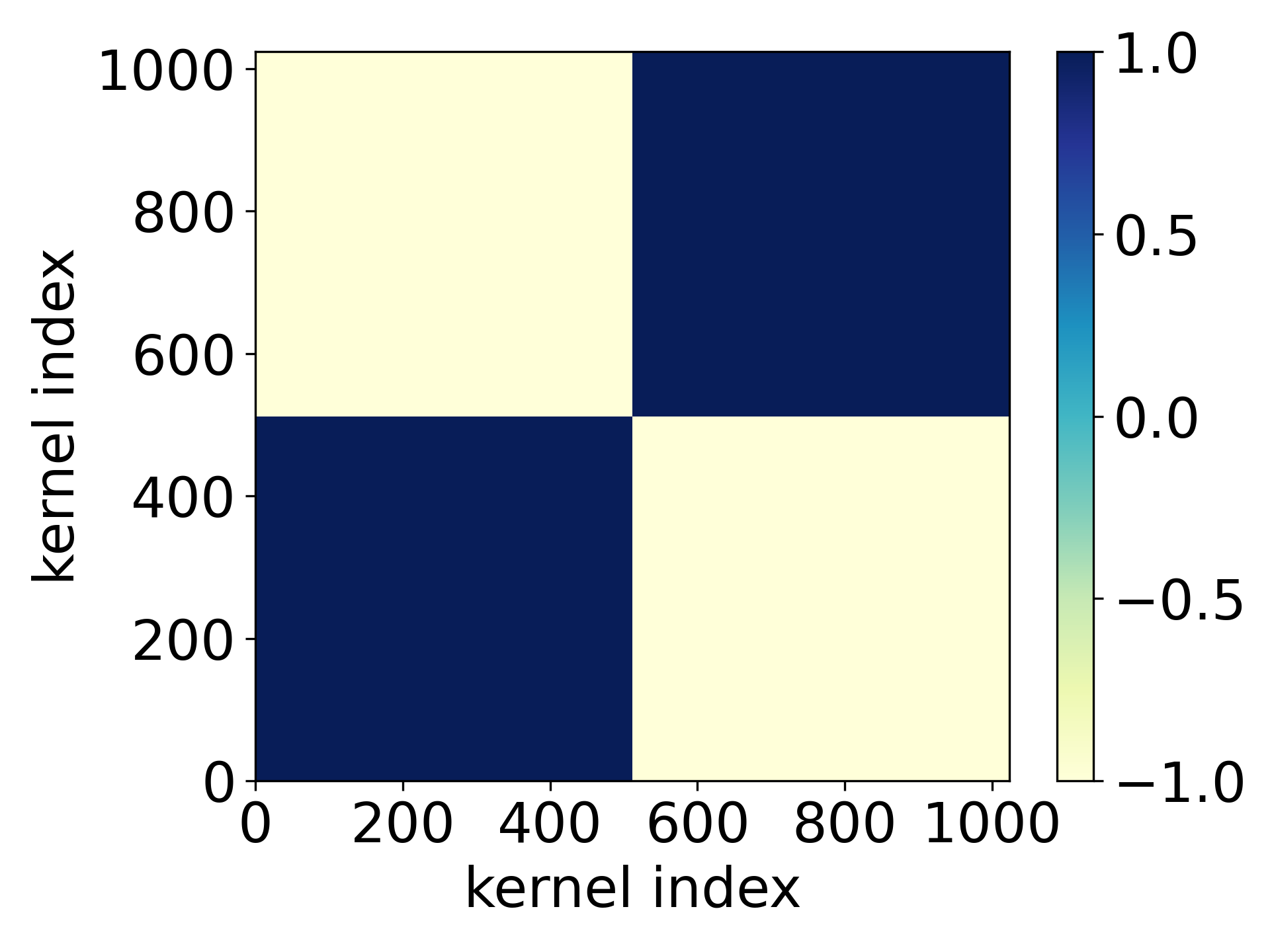}
         \caption{layer $2$}
         \label{tanha2lccrossfinal}
     \end{subfigure}
     \hfill
     \begin{subfigure}[b]{0.3\textwidth}
         \centering
         \includegraphics[width=\textwidth]{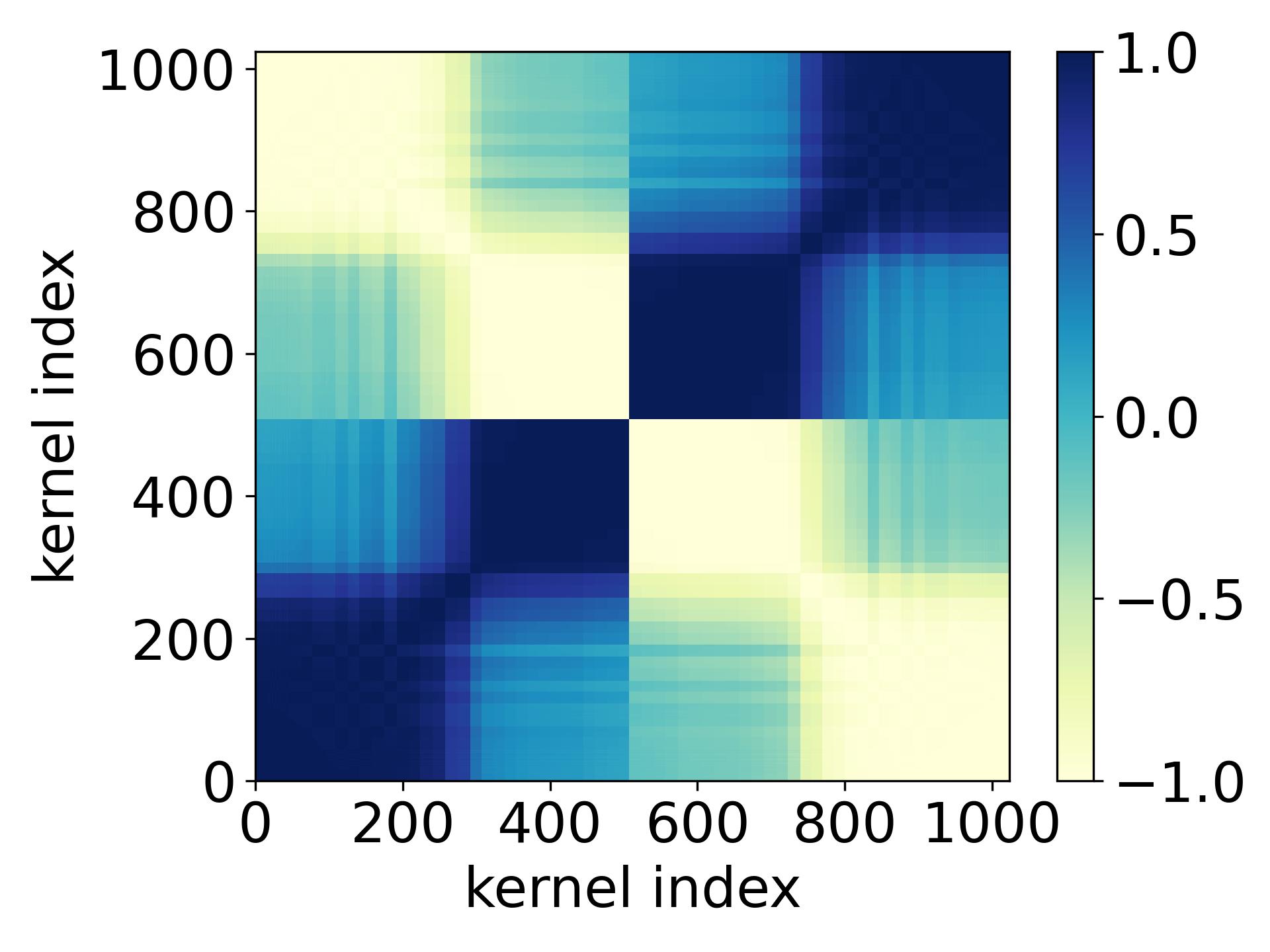}
         \caption{layer $3$}
         \label{tanha3lccrossfinal}
     \end{subfigure}
        \caption{Final condensation of a CNN with three convolution layers. The kernel size is $m=5$. The colors in the figure show the cosine similarity of weight vectors of each convolution kernel. The activation for all convolution layers are $\rm{tanh}(x)$. The number of the steps is all at the end of the training. The convolution kernels are initialized by $\gamma=2$. The learning rate is $2\times 10^{-6}$. We use cross-entropy (with softmax) as the criterion. The optimizer is full batch Adam on CIFAR10 dataset. }
        \label{fig:tanhathreeccrossfinal}
\end{figure}

We first show that when initialized with small weights, the convolutional kernels of a ${\rm tanh}(x)$ CNN undergo condensation during the training process.
As shown in Fig. \ref{fig:tanhathreeccrossfinal}, we train a CNN with three layers by cross-entropy loss until the training accuracy reaches $100\%$ (the accuracy during training is shown in Fig. \ref{fig:tanhathreeccrossfinalacc} in appendix). 
In each layer, we compute the cosine similarity between each pair of kernels. This reveals a clear condensation phenomenon after training.
 
Understanding the mechanism of condensation phenomenon is challenging.
To this end, we start by studying the initial training stage. We then study the initial condensation in CNNs. 

\begin{figure}
     \centering
     \begin{subfigure}[b]{0.3\textwidth}
         \centering
         \includegraphics[width=\textwidth]{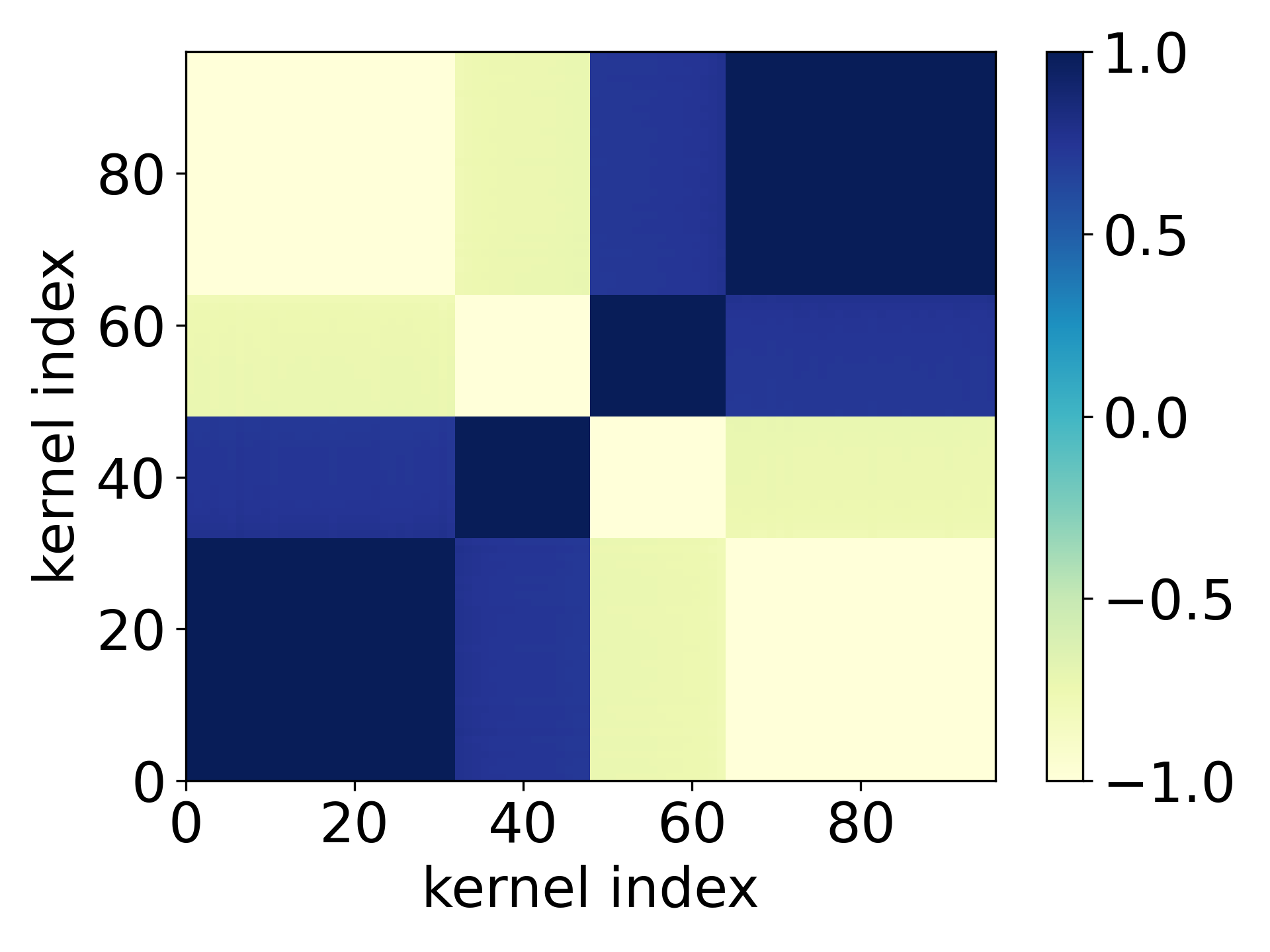}
         \caption{ layer $1$}
         \label{tanha1lccross}
     \end{subfigure}
     \hfill
     \begin{subfigure}[b]{0.3\textwidth}
         \centering
         \includegraphics[width=\textwidth]{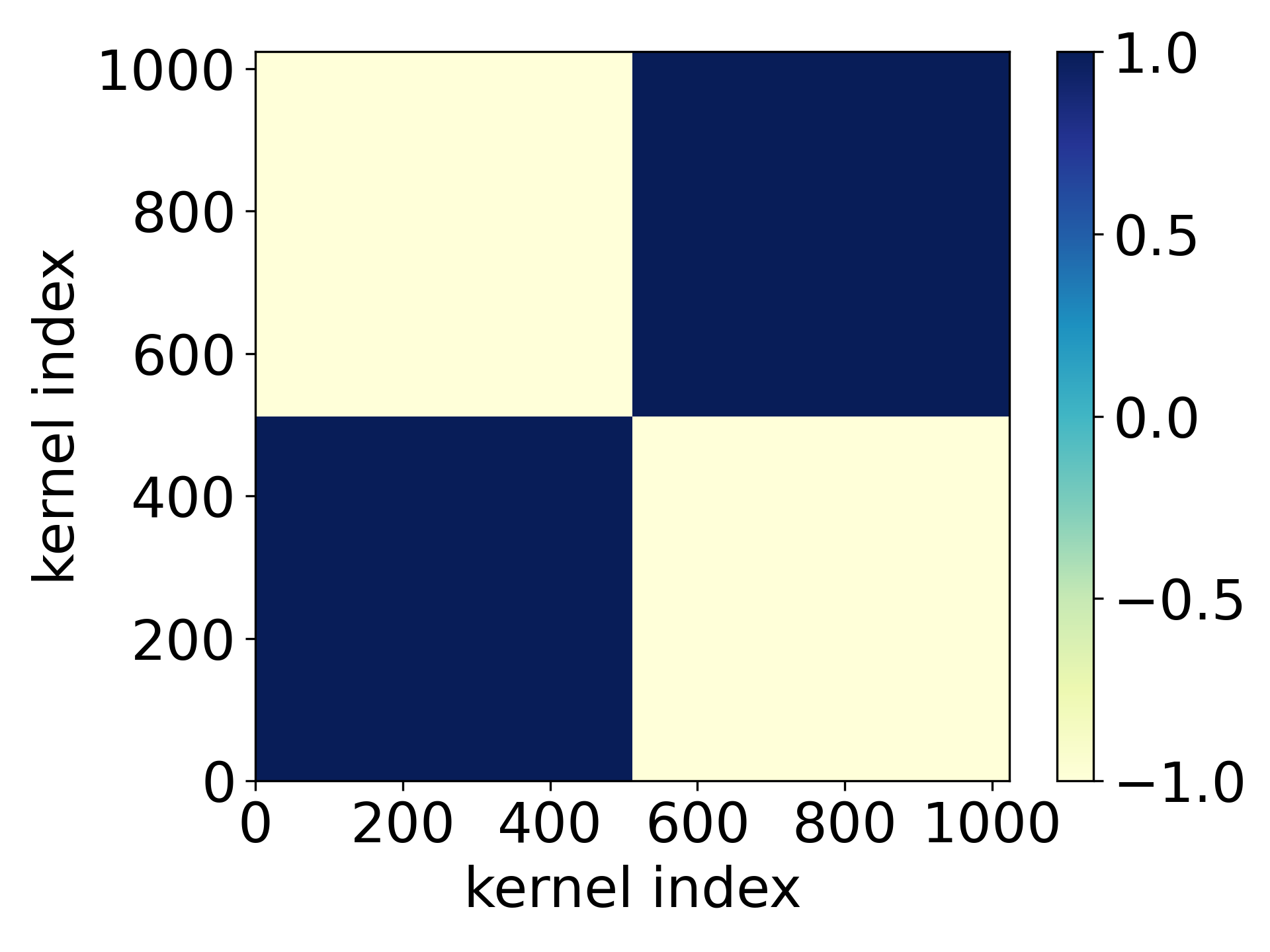}
         \caption{layer $2$}
         \label{tanha2lccross}
     \end{subfigure}
     \hfill
     \begin{subfigure}[b]{0.3\textwidth}
         \centering
         \includegraphics[width=\textwidth]{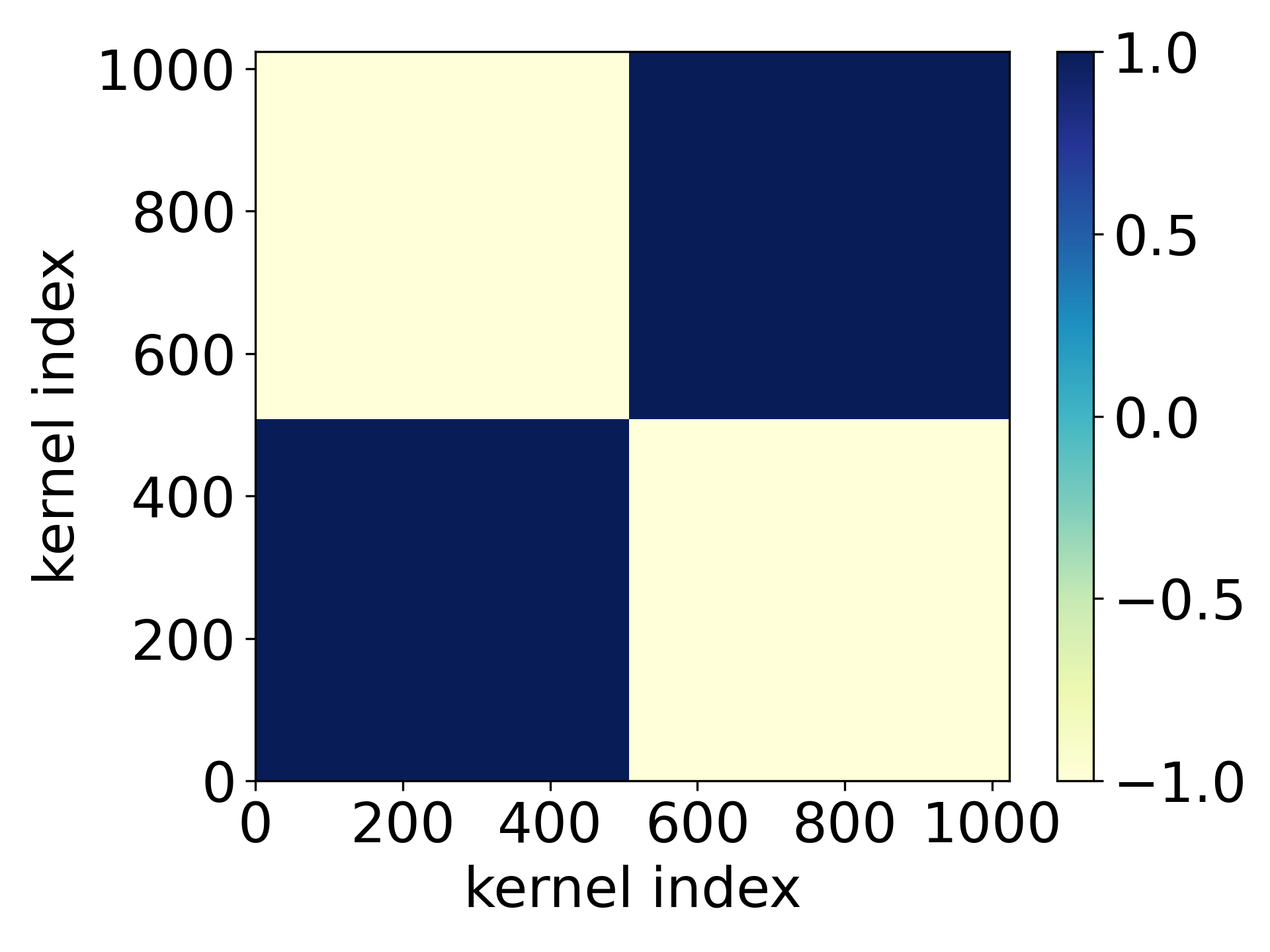}
         \caption{layer $3$}
         \label{tanha3lccross}
     \end{subfigure}
        \caption{Initial stage of Fig. \ref{fig:tanhathreeccrossfinal}, selected at epoch $=300$ with accuracy less than $20\%$.}
        \label{fig:tanhathreeccross}
\end{figure}

The initial stage (accuracy less than $20\%$) of Fig. \ref{fig:tanhathreeccrossfinal} is shown in Fig. \ref{fig:tanhathreeccross}. 
For each layer, nearly all kernels are condensed into two opposite directions in Fig. \ref{fig:tanhathreeccross}. In layer one, there are actually two pairs of opposite directions. This subtle phenomenon can be explained by our theory.
 
\begin{figure}
     \centering
    \begin{subfigure}[b]{0.3\textwidth}
        \centering
        \includegraphics[width=\textwidth]{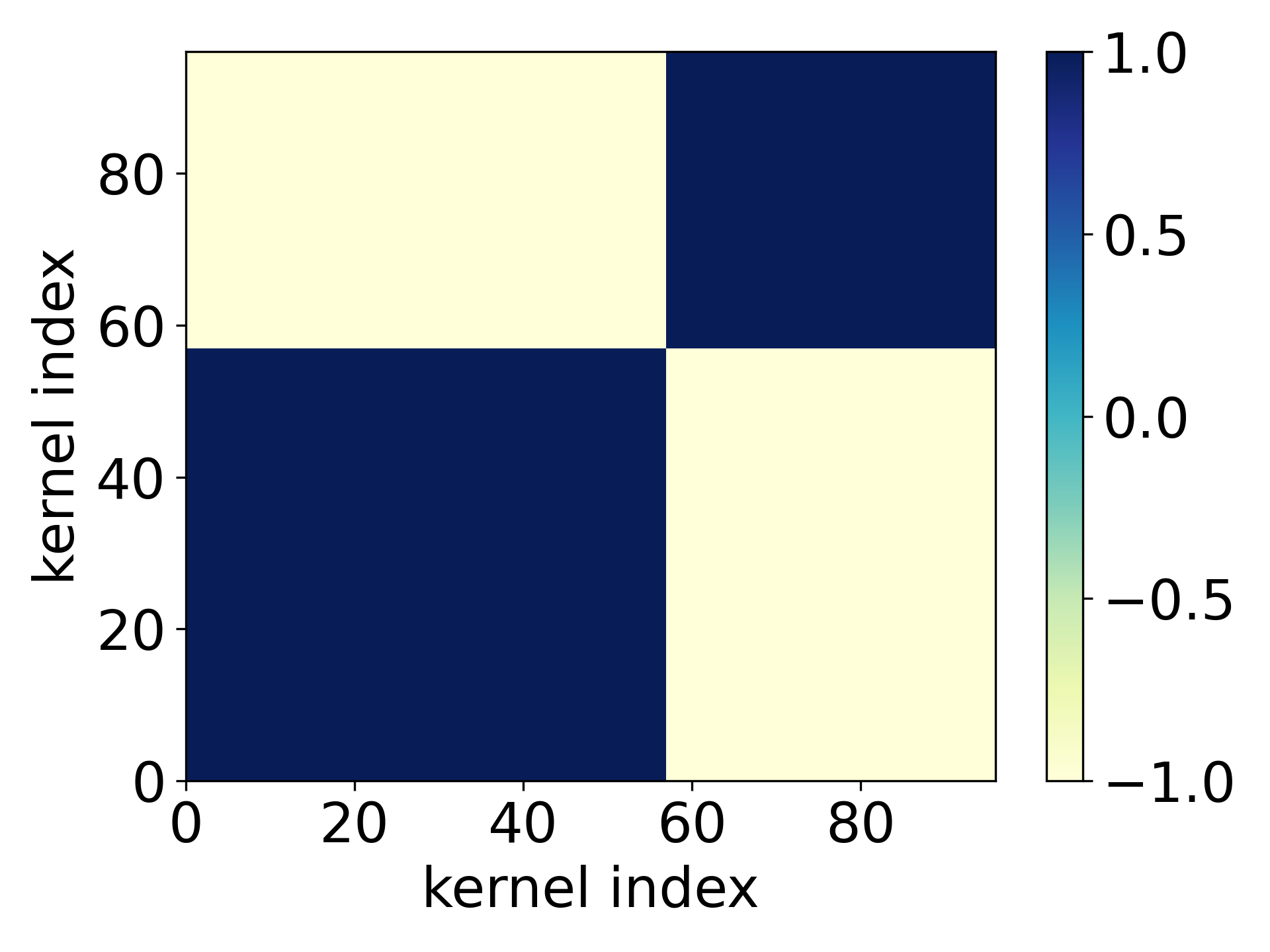}
        \caption{${\rm ReLU}(x)$}
        \label{relua2le}
    \end{subfigure}
    \hfill
     \begin{subfigure}[b]{0.3\textwidth}
         \centering
         \includegraphics[width=\textwidth]{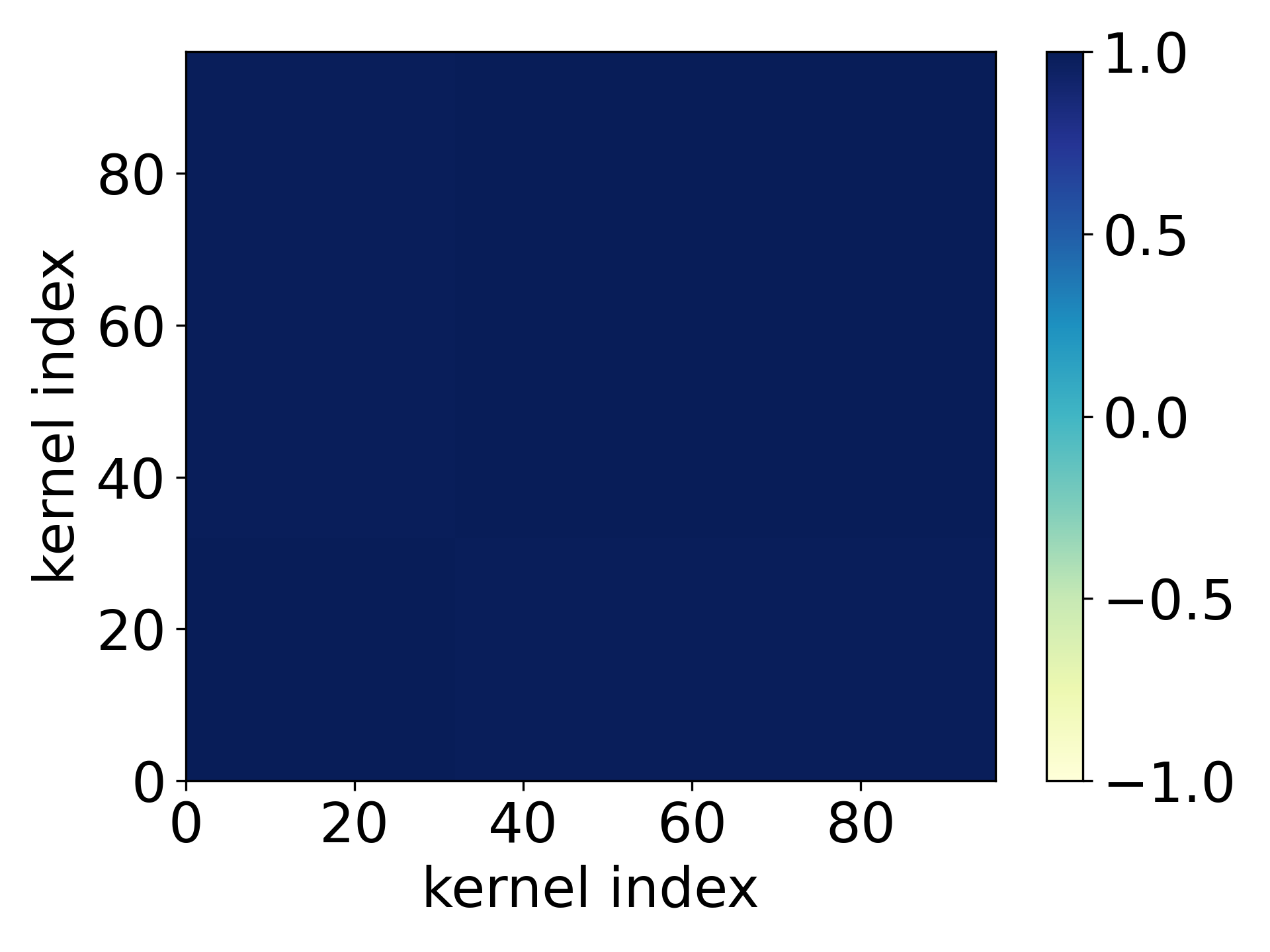}
         \caption{ $\rm{Sigmoid}(x)$}
         \label{sigmoida2le}
     \end{subfigure}
     \hfill
     \begin{subfigure}[b]{0.3\textwidth}
         \centering
         \includegraphics[width=\textwidth]{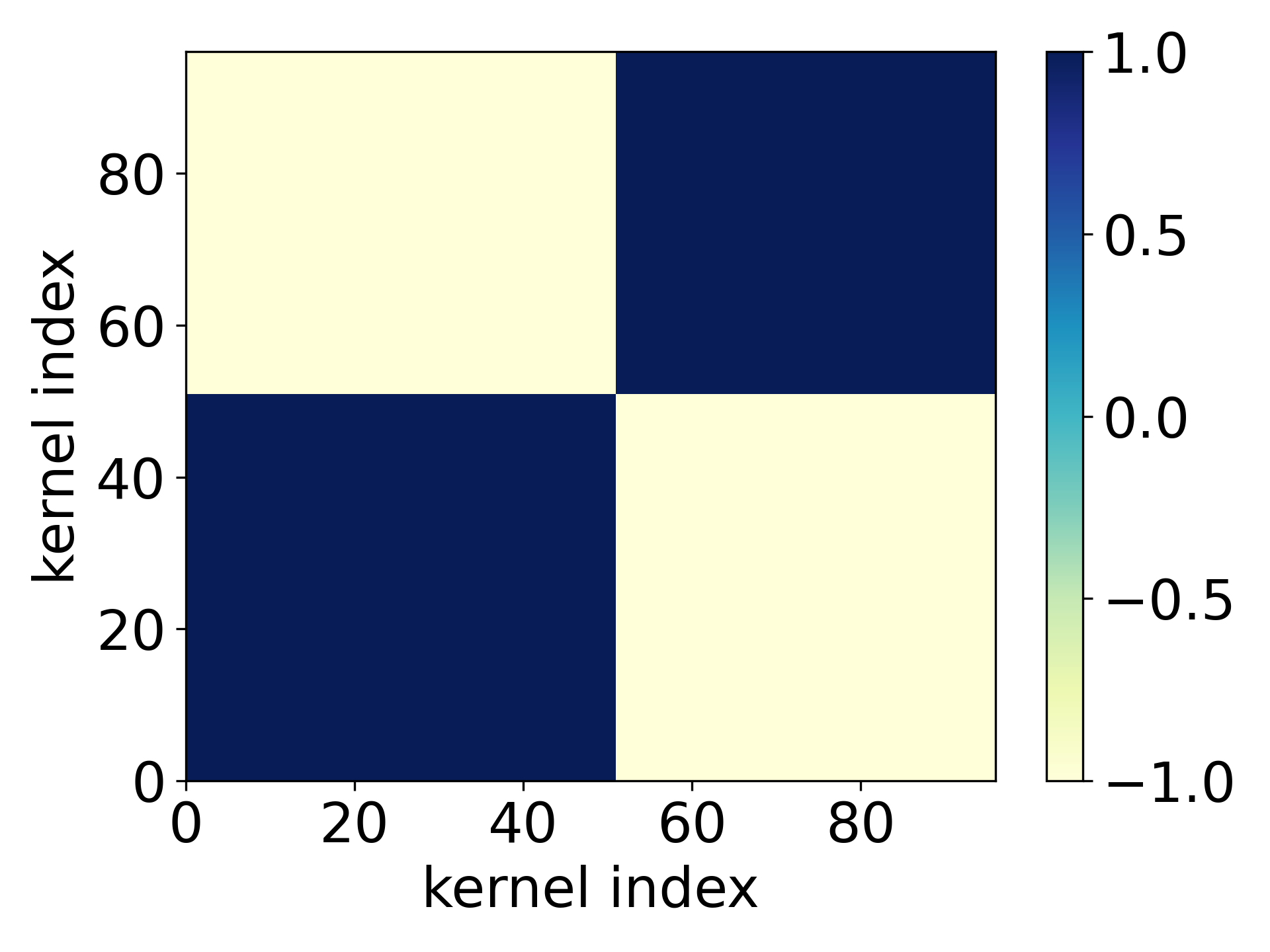}
         \caption{$\rm{tanh}(x)$}
         \label{tanha2le}
     \end{subfigure}
        \caption{Condensation of CNNs with different activations (indicated by sub-captions) for convolutional layers. The network has 32 kernels in the convolution layer, followed by a ReLU fully-connected layer with 1024 neurons and 10-dimensional output with softmax. The kernel size is $m=5$. The learning rate for all three experiments is $5\times 10^{-7}$. epoch$=1000$, epoch$=5000$ and epoch$=300$. The convolution layer is initialized by $\gamma=2$. We use full batch Adam with cross-entropy on CIFAR10 dataset. }
        \label{fig:tanh1c2l}
\end{figure}

We further examine the different activation functions. For illustration, we consider two-layer CNNs with activations $\mathrm{ReLU}(x)$, $\mathrm{Sigmoid}(x)$, or $\mathrm{tanh}(x)$. As shown in Fig. \ref{fig:tanh1c2l}, we can still see a very clear condensation phenomenon. 
Note that, as the learning rate is rather small to see the detailed training process, the epoch selected for studying the initial stage may appear  large.

\begin{figure}
     \centering
     \begin{subfigure}[b]{0.3\textwidth}
         \centering
         \includegraphics[width=\textwidth]{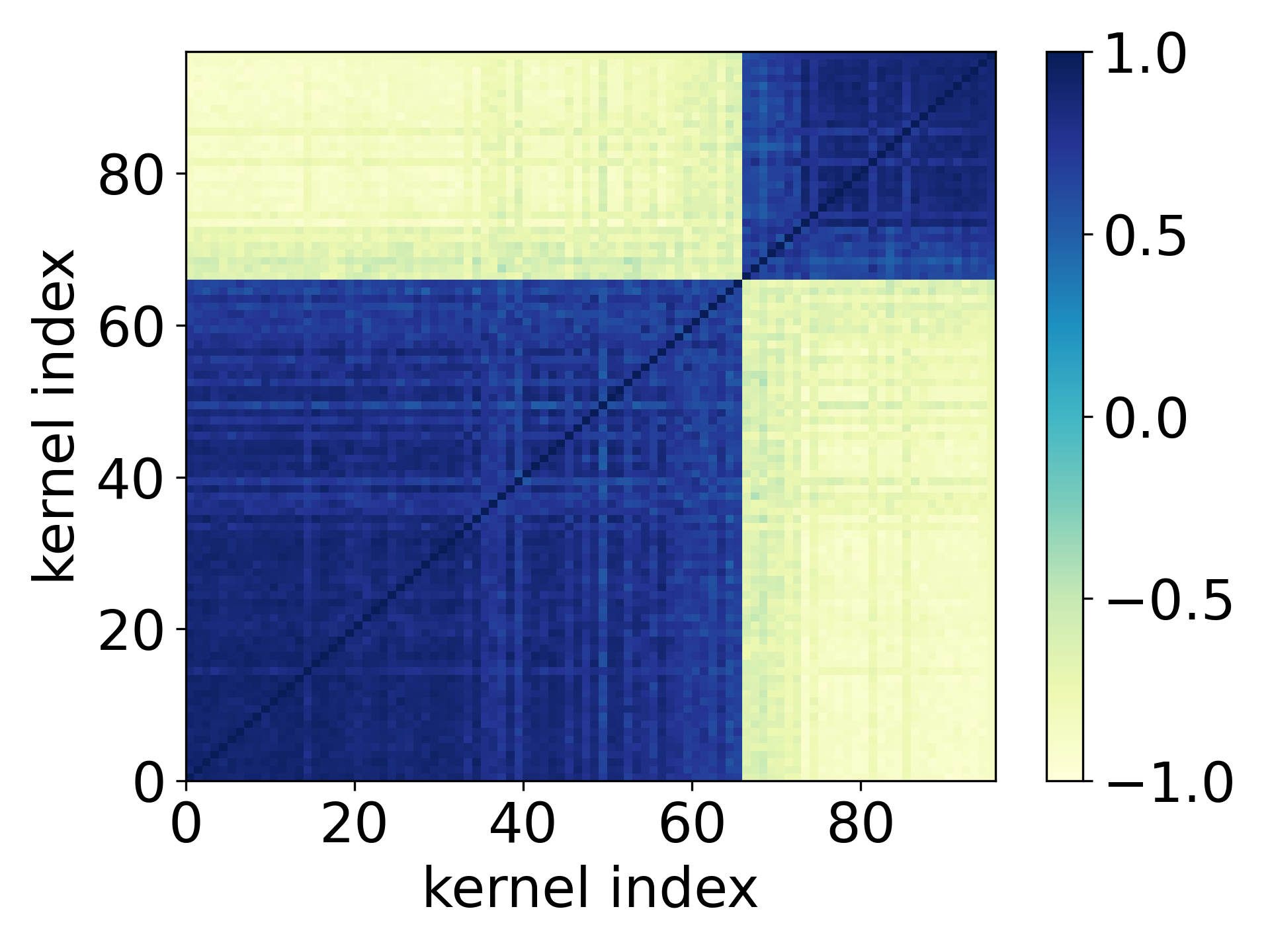}
         \caption{ layer $1$}
         \label{tanha1lc}
     \end{subfigure}
     \hfill
     \begin{subfigure}[b]{0.3\textwidth}
         \centering
         \includegraphics[width=\textwidth]{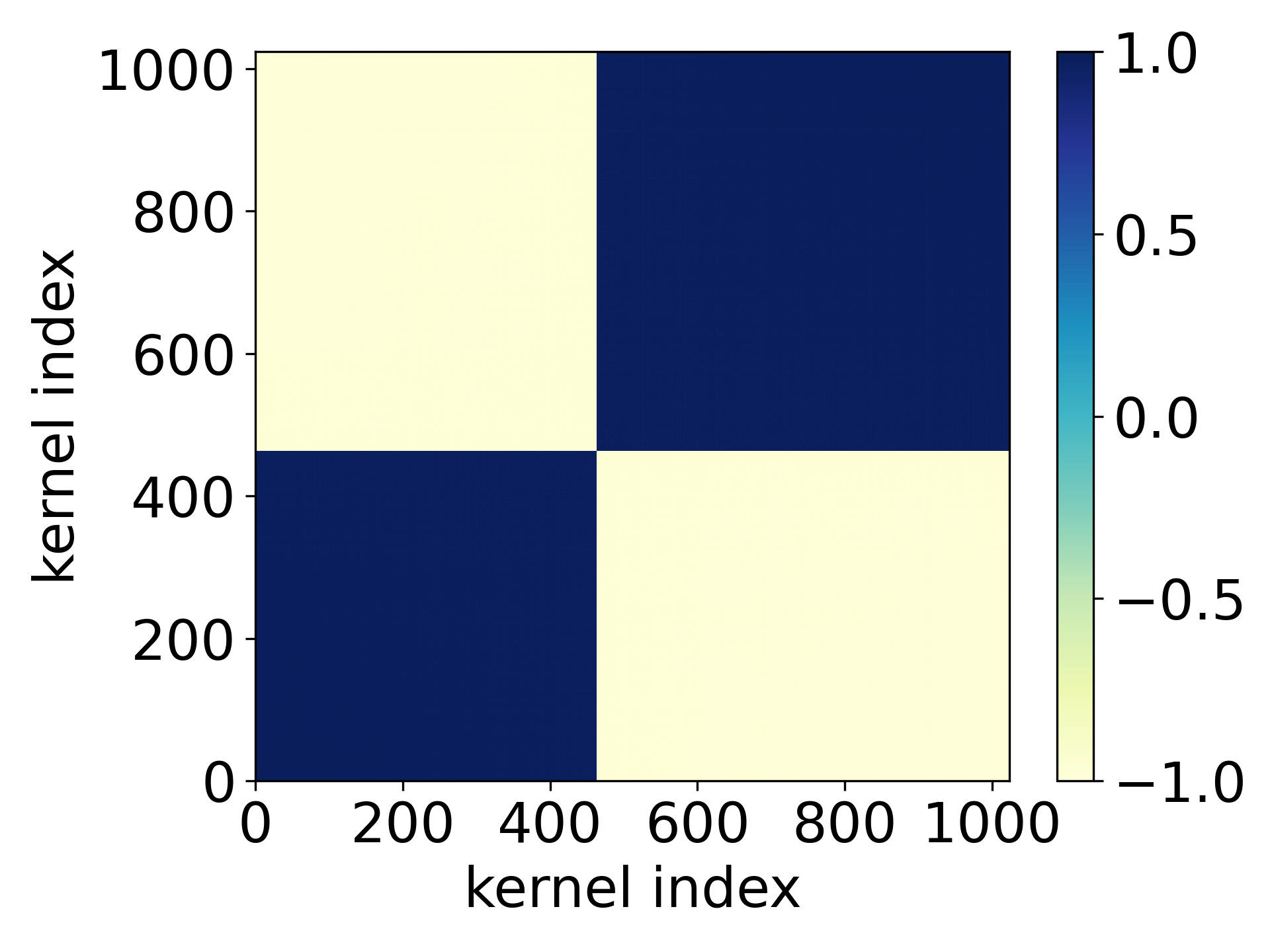}
         \caption{layer $2$}
         \label{tanha2lc}
     \end{subfigure}
     \hfill
     \begin{subfigure}[b]{0.3\textwidth}
         \centering
         \includegraphics[width=\textwidth]{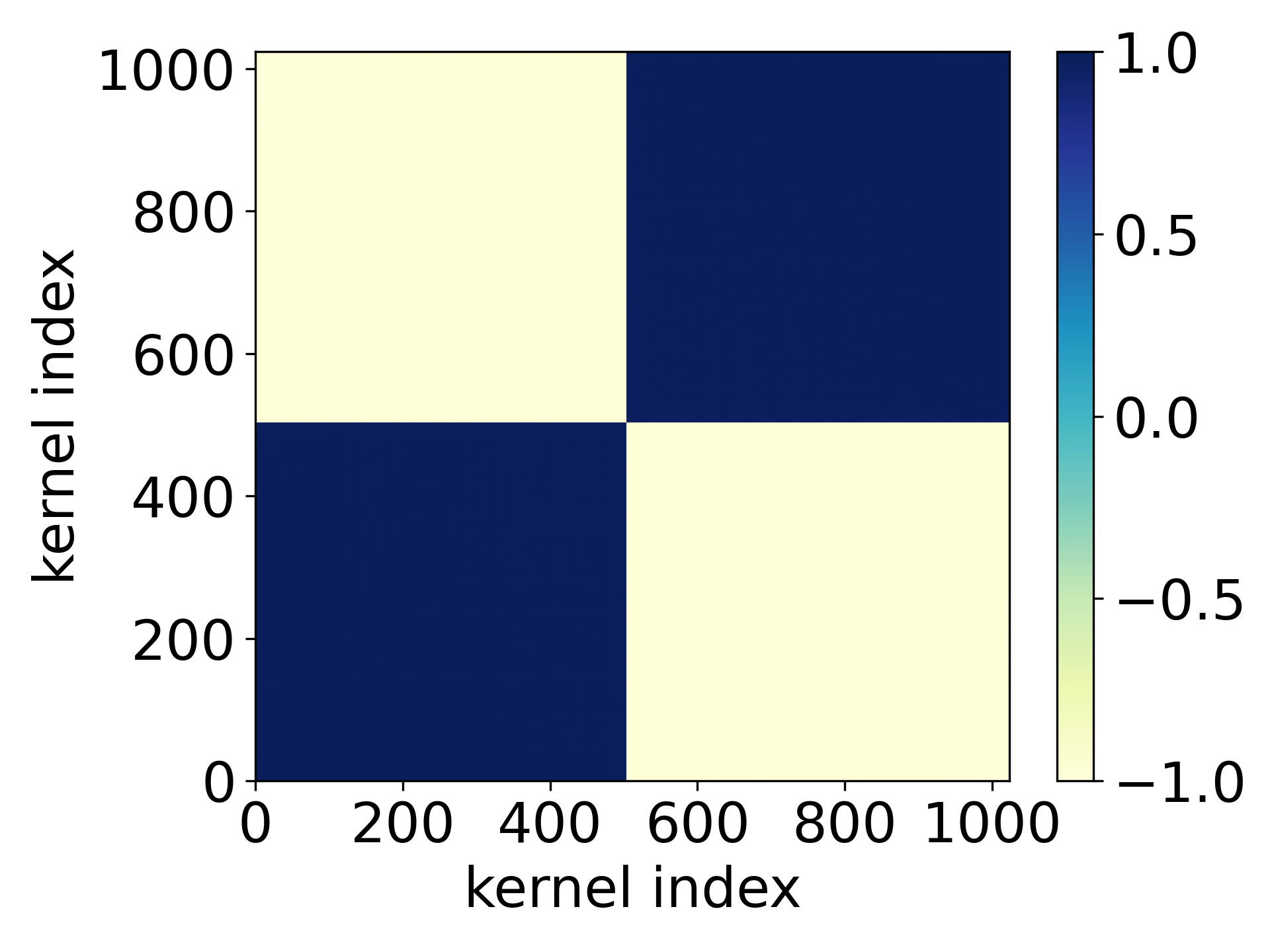}
         \caption{layer $3$}
         \label{tanha3lc}
     \end{subfigure}
        \caption{Condensation of $\rm{tanh}(x)$ CNN with three convolution layer by MSE loss. The kernel size is $m=5$. The color indicates the cosine similarity between kernels.  The number of the steps are at epoch $=200$, epoch $=300$ and epoch $=300$. The convolution kernels are initialized by $\gamma=1.2$. The learning rate is $1\times 10^{-6}$. We use one hot vector as label and use softmax. The optimizer is full-batch Adam on CIFAR10 dataset. }
        \label{fig:tanhathreecMSESoft}
\end{figure}

In our theoretical study, we consider the MSE loss. Therefore, we also show an example with MSE loss (softmax is also attached with the output layer) for activation $\rm{tanh}(x)$ in Fig. \ref{fig:tanhathreecMSESoft}. Similarly, condensation is clearly observed.  An MSE example without softmax is shown in \ref{Fig:tanhathreecMSE} in appendix.



\begin{figure}
    \centering
    \begin{subfigure}[b]{0.45\textwidth}
        \centering
        \includegraphics[width=\textwidth]{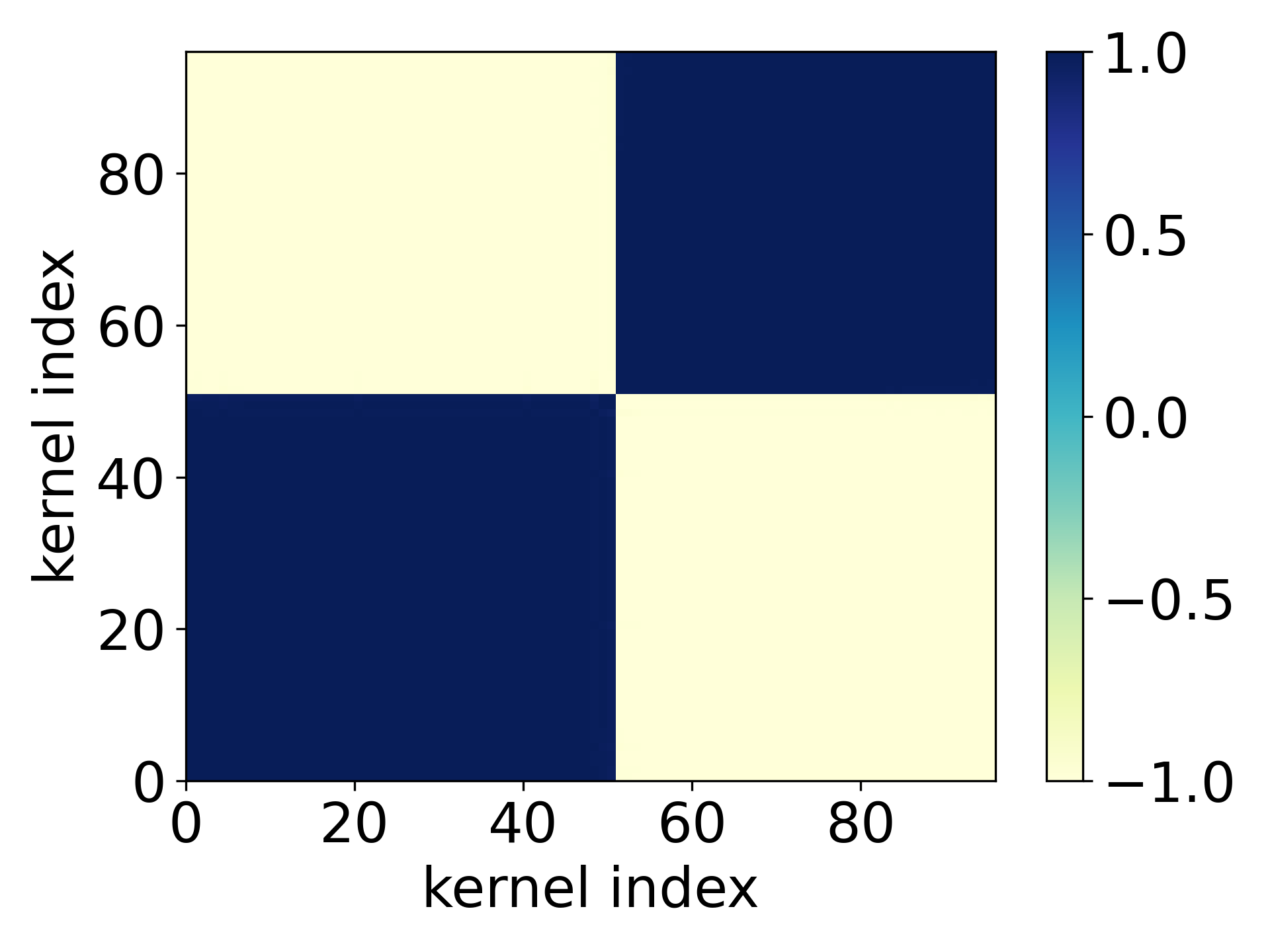}
        \caption{condensation}
        \label{GDCifar10c}
    \end{subfigure}
    \hfill
    \begin{subfigure}[b]{0.45\textwidth}
        \centering
        \includegraphics[width=\textwidth]{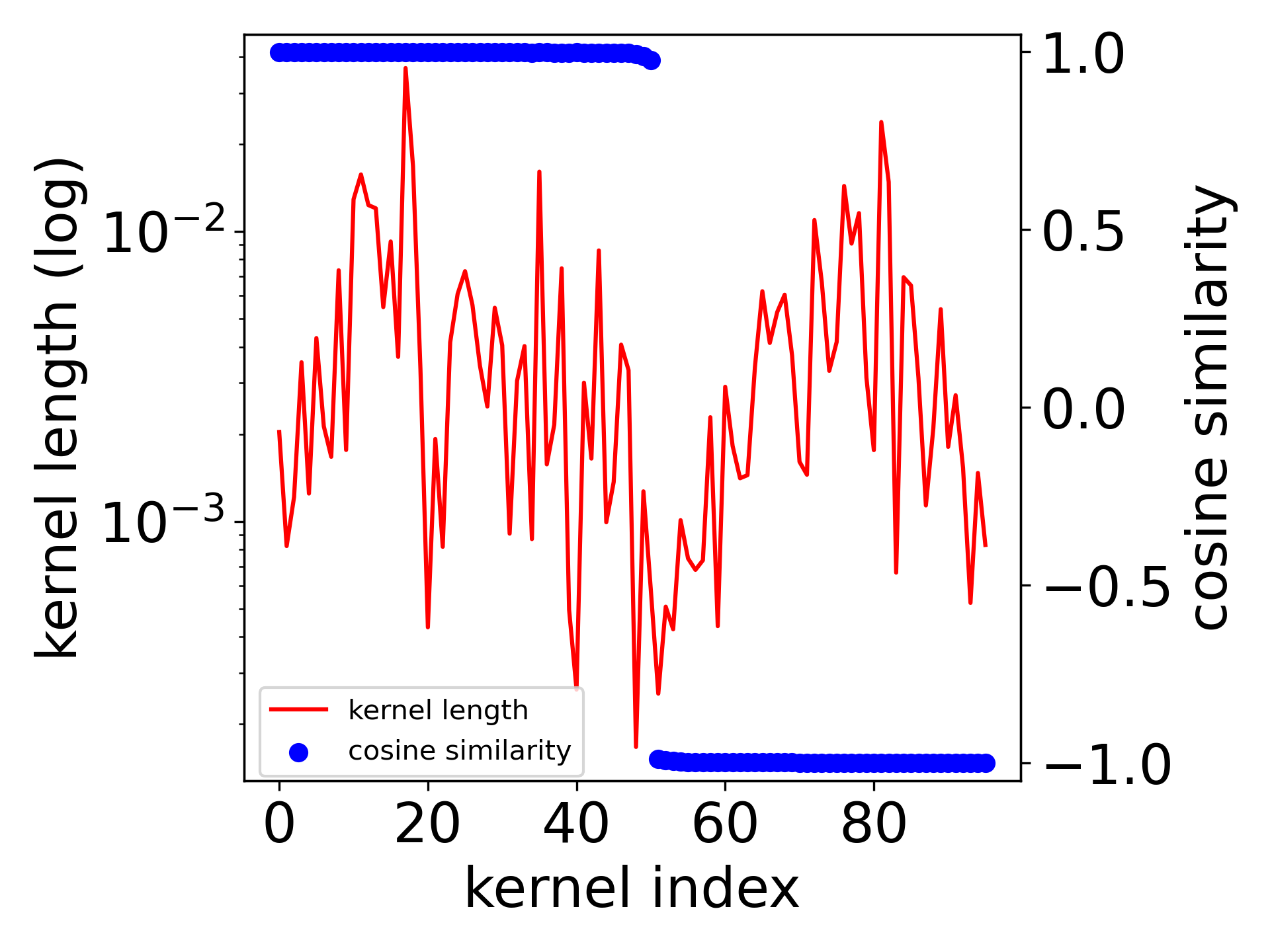}
        \caption{directions}
        \label{GDCifar10d}
    \end{subfigure}
    \caption{Condensation of two-layer CNN by GD and MSE training on CIFAR10 dataset with data size $n=500$. (a) cosine similarity. (b) left ordinate (red): the amplitude of each kernel; (b) right ordinate (blue): cosine similarity between kernel weight and $\mathbbm{1}$. The activation function of the convolution part is $\rm{tanh}(x)$. The kernel size is $m=3$. The learning rate is $5\times 10^{-6}$. The number of the selected steps is epoch $=4500$. The convolution layer is initialized by $\gamma=4$. }
    \label{fig:tanh1cGDCifar10}
\end{figure}

These examples demonstrate that condensation occurs when using the Adam method to train networks. We can also observe the condensation with the gradient descent (GD) method.
Fig. \ref{fig:tanh1cGDCifar10} illustrates that with GD, the kernel weights of a two-layer CNN still undergo condensation at two opposite directions during training.
Moreover, we observe that the direction of the condensation is $\boldsymbol{v}=\mathbbm{1}$. Fig. \ref{fig:tanh1cGDCifar10}(b) shows that the cosine similarity between each kernel weight and $\mathbbm{1}$ is almost equal to 1 or -1.

Similar results of MNIST dataset are also shown in Figs. \ref{Fig:1c} for CNNs with different activations, Fig. \ref{Fig:tanhathree} for multi-layer $\mathrm{tanh}$ CNN in appendix. Also, two-layer CNNs with $32$ and $320$ kernels trained by GD are shown in Fig.\ref{Fig:tanh1cGD} and Fig.\ref{Fig:tanh1cGD320}, respectively, in appendix. 

\section{Theory}
In the following section, we aim to provide theoretical evidence that kernels in the same layer tend to converge to one or a few directions when initialized with small values.
In this part, we shall impose some   technical conditions  on the activation function and input samples.
We start with a   technical condition~\cite[Definition 1]{zhou2022towards}  on the activation function $\sigma(\cdot)$.
\begin{definition}[Multiplicity $r$]\label{def...Multiplicity-main} 
   $\sigma(\cdot):\sR\to\sR$ has multiplicity $r$
    if there exists an integer $r\geq 1$, such that  for all $0\leq s\leq r-1$, the $s$-th order derivative satisfies $\sigma^{(s)}(0)=0$, and $\sigma^{(r)}(0) \neq 0$.
\end{definition}
\noindent We   list out some examples of activation functions with different multiplicities.
\begin{remark}
~\\
\begin{itemize}
\item $\rm{tanh}(x):=\frac{\exp(x)-\exp(-x)}{\exp(x)+\exp(-x)}$ is with multiplicity $r=1$;\\
\item $\rm{SiLU}(x):=\frac{x}{1+ \exp(-x)}$ is with multiplicity $r=1$;\\
\item $\rm{xtanh}(x):=\frac{x\exp(x)-x\exp(-x)}{\exp(x)+\exp(-x)}$ is with multiplicity $r=2$.
\end{itemize}
\end{remark}
\begin{assumption}[Multiplicity $1$]\label{Assumption....ActivationFunctions-main}
The activation function $\sigma\in\fC^2(\sR)$, and there exists a universal constant $C_L>0$, such that   its first and second  derivatives   satisfy 
\begin{equation} 
   \Norm{\sigma^{(1)}(\cdot)}_{\infty}\leq C_L,\quad \Norm{\sigma^{(2)}(\cdot)}_{\infty}\leq C_L.
\end{equation}
Moreover,
\begin{equation} 
\sigma(0)=0,\quad \sigma^{(1)}(0)=1.
\end{equation}
\end{assumption}
\begin{remark}
We remark that $\sigma$ has multiplicity $1$.  
$\sigma^{(1)}(0)=1$   can be replaced by  $\sigma^{(1)}(0)\neq 0$, and we set $\sigma^{(1)}(0)=1$ for simplicity, which can be easily satisfied by replacing the original activation $\sigma(\cdot)$ with $\frac{\sigma(\cdot)}{\sigma^{(1)}(0)}$.
\end{remark}

\begin{assumption} \label{assumption...Data-main}
The training inputs  $\{ \vx_i\}_{i=1}^n$  and labels  $\{ y_i\}_{i=1}^n$ satisfy  that   there exists a universal constant $c>0$, such that given any $i\in[n]$, then for each $u\in[W_0]$, $v\in[H_0]$ and $\alpha\in[C_0]$, the following holds
\[  \frac{1}{c}\leq \Abs{\vx_{u,v,\alpha}(i)}, \quad\Abs{y_{i}}\leq c.\] 
\end{assumption}
\noindent
We assume further that 
\begin{assumption}\label{assump...LimitExistence-main}
The following limit exists
\begin{equation} \label{eq...assump...definition...LimitExistence-main}
{\gamma}:=\lim_{M\to\infty} -\frac{\log \eps^2}{\log M}.
\end{equation}

\end{assumption}
As $M\rightarrow\infty$, then $\epsilon\rightarrow0$ and since   multiplicity $r=1$, we can use the first-order Taylor expansion to approximate the network output and obtain a linear dynamics 
\begin{equation}
   \frac{\D \vtheta_\beta}{\D t}=\mA\vtheta_\beta,  
\end{equation}
where 
\begin{align*}
\vtheta_\beta:=\Big(&\mW_{0,0,1,\beta},\mW_{0,1,1,\beta},\cdots,\mW_{0,m-1,1,\beta}; \mW_{1,0,1,\beta},\cdots,\mW_{1,m-1,1,\beta};\cdots\cdots\mW_{m-1,m-1,1,\beta};\\
&\mW_{0,0,2,\beta},\mW_{0,1,2,\beta},\cdots,\mW_{0,m-1,2,\beta}; \mW_{1,0,2,\beta},\cdots,\mW_{1,m-1,2,\beta};\cdots\cdots\mW_{m-1,m-1,2,\beta};\\
&\cdots\cdots\cdots\cdots\cdots\cdots\cdots\cdots\cdots\cdots\cdots\cdots\cdots\cdots\cdots\cdots\cdots\cdots\cdots\cdots\cdots\cdots\cdots\cdots\cdots\\
&\mW_{0,0,C_0,\beta},\mW_{0,1,C_0,\beta},\cdots,\mW_{0,m-1,C_0,\beta};  \cdots,\mW_{1,m-1,C_0,\beta};\cdots\cdots\mW_{m-1,m-1,C_0,\beta};\vb_{\beta};\\
&{\va}_{1,1,\beta}, {\va}_{1,2,\beta},\cdots,{\va}_{1,H_1,\beta};{\va}_{2,1,\beta},\cdots,{\va}_{2,H_1,\beta};\cdots\cdots{\va}_{W_1,H_1,\beta}\Big)^\T,
\end{align*} 
and 
\begin{equation}
    \mA:=\left[\begin{array}{cc}
\mzero_{(C_0 m^2+1)\times (C_0m^2+1)} & {\mZ}^\T \\
\mZ & \mzero_{W_1H_1\times W_1H_1} 
\end{array}\right],
\end{equation}
where $\mZ$ is detailed described by multi channel \eqref{multichannel_matrix} and single channel \eqref{Singlechannel_matrix} in appendix . 
We perform singular value decomposition~(SVD) on $\mZ$, i.e., 
\begin{equation}\label{eq...text...E-LinearDynamics....SVD-main}
\mZ=\mU\Lambda\mV^\T,    
\end{equation}
where
\[
\mU=\left[\vu_1,\vu_2,\cdots,\vu_{W_1H_1}\right],\quad\mV=\left[\bm{v}_1,\bm{v}_2,\cdots,\bm{v}_{m^2+1}\right],
\]
and as we denote $r:=\mathrm{rank}(\mZ)$, naturally, $r\leq \min\{W_1H_1, m^2+1\}$,  we have $r$ singular values,
\[\lambda_1\geq\lambda_2\geq \cdots \geq\lambda_r> 0,\]
 and WLOG, we assume that  
 \begin{assumption}[Spectral Gap of $\mZ$]\label{assump...SpectralGap-main}
The singular values $\{\lambda_k\}_{k=1}^r$ of $\mZ$ satisfy that
\begin{equation} \lambda_1>\lambda_2\geq \cdots \geq\lambda_r> 0,\end{equation}     
and we denote the spectral gap between $\lambda_1$ and $\lambda_2$ by 
\[\Delta\lambda:= \lambda_1-\lambda_2.\]
 \end{assumption}
We denote   that
\begin{align*}
\vtheta_{\mW,\beta}&:=\left(\mW_{0,0,\beta},\mW_{0,1,\beta},\cdots,\mW_{0,m-1,\beta}; \mW_{1,0,\beta},\cdots,\mW_{1,m-1,\beta};\cdots\cdots\mW_{m-1,m-1,\beta};{\vb}_{\beta}  \right)^\T,\\
\vtheta_{\va,\beta}&:= \left({\va}_{1,1,\beta}, {\va}_{1,2,\beta},\cdots,{\va}_{1,H_1,\beta};{\va}_{2,1,\beta},\cdots,{\va}_{2,H_1,\beta};\cdots\cdots{\va}_{W_1,H_1,\beta}\right)^\T,
\end{align*}
hence 
\[
\vtheta_\beta=\left(\vtheta_{\mW,\beta}^\T, \vtheta_{\va,\beta}^\T\right)^\T.
\]
In order to study the phenomenon  of condensation, we  concatenate the vectors $\left\{ \vtheta_{\mW,\beta}\right\}_{\beta=1}^M$ into
\[
\vtheta_{\mW}:=\mathrm{vec}\left(\left\{ \vtheta_{\mW,\beta}\right\}_{\beta=1}^M\right),
\]
and we denote further that 
\[
\vtheta_{\mW,\bm{v}_1}:=\fP_{1}\vtheta_{\mW}:= \left( \left<\vtheta_{\mW,1},\bm{v}_1\right>, \left<\vtheta_{\mW,2},\bm{v}_1\right>, \cdots \left<\vtheta_{\mW,M},\bm{v}_1\right>\right)^\T,
\]
where $\bm{v}_1$ is the  eigenvector of the largest eigenvalue of $\mZ^\T\mZ$, or the first column vector of $\mV$ in \eqref{eq...text...E-LinearDynamics....SVD}.   

In appendix, we prove that  for any $\eta_0>\frac{\gamma-1}{100}>0$, there exists 
\begin{equation}
T_{\mathrm{eff}}>\frac{1}{\lambda_1}\left[ \log\left(\frac{1}{4}\right)+\left({\frac{\gamma-1}{4}}-\eta_0\right)\log(M)\right],
\end{equation}
leading to the following theorem.

\begin{theorem}\label{thm..Condense-main}
Given any $\delta\in(0,1)$, under \Cref{Assumption....ActivationFunctions-main}, \Cref{assumption...Data-main}, \Cref{assump...LimitExistence-main} and \Cref{assump...SpectralGap-main}, if $\gamma> 1$,
then	with probability at least $1-\delta$ over the choice of $\vtheta^0$,
we have  
\begin{equation}\label{eq...thm...Condense...PartOne-main}
\lim_{M\to+\infty} \sup\limits_{t\in[0,T_{\mathrm{eff}}]} \frac{\Norm{\vtheta_{\mW}(t)-\vtheta_{\mW}(0)}_2}{\Norm{\vtheta_{\mW}(0)}_2}=+\infty,
\end{equation}
and
\begin{equation}\label{eq...thm...Condense...PartTwo-main}
\lim_{M\to+\infty} \sup\limits_{t\in[0,T_{\mathrm{eff}}]}\frac{\Norm{\vtheta_{\mW, \bm{v}_1}(t) }_2}{\Norm{\vtheta_{\mW}(t)}_2} =1.
\end{equation}
\end{theorem}

The theorem shows that as the number of channels  increases,  two implications arise within a finite training time in the   small initialization scheme. Firstly, the relative change of kernel weight becomes considerably large, demonstrating significant non-linear behavior. Secondly, the kernel weight concentrates on one direction. 

\begin{figure}[htbp]
    \centering
    \begin{subfigure}[b]{0.6\textwidth}
        \centering
        \includegraphics[width=\textwidth]{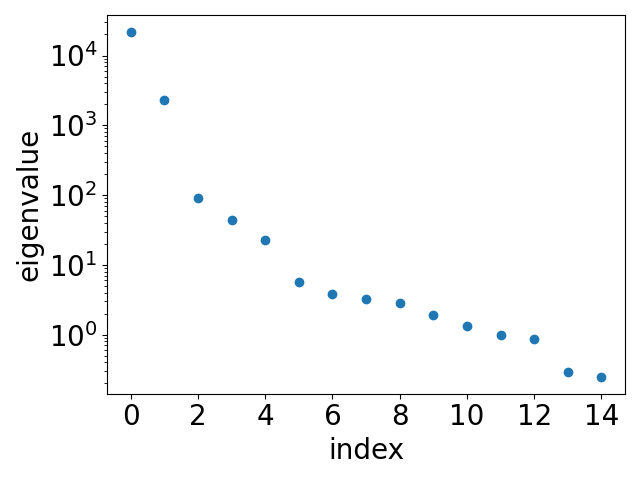}
        \caption{eigenvalues}
        \label{eigCifar10}
    \end{subfigure}
    \caption{The largest $15$ eigenvalues of $\mZ^\T\mZ$ in tanh(x) CNN on CIFAR10 dataset. A clear spectral gap ($\Delta\lambda := \lambda_1-\lambda_2$) could be observed, which satisfies \Cref{assump...SpectralGap-main} and leads to condensing into one direction or a pair of opposite directions. }
    \label{CIFAR10verify}
\end{figure}

The mechanism of the  concentration effect, or the phenomenon of initial condensation,  arises from  the spectral gap in \Cref{assump...SpectralGap-main}. The original definition of spectral gap is a fundamental concept in the analysis of Markov chains, which refers to the minimum difference between the two largest eigenvalues of the transition matrix of a Markov chain, and is closely related to the convergence rate of the chain to its stationary distribution~\cite{olla2012fluctuations}.
A large spectral gap indicates that the Markov chain converges quickly to its stationary distribution, while a small spectral gap suggests a slower convergence rate. 
Similar to that, one may observe from
relation \eqref{equ:...spectral gap}   that   the larger the spectral gap is, the faster the kernel weight concentrates onto the direction of $\bm{v}_1$. 
Fig. \ref{CIFAR10verify} shows an example of eigenvalues of $\mZ^\T\mZ$, in which a huge spectral gap can be observed. We then study the eigenvector of the maximum eigenvalue. In this example, since the input has 3 channels, we decompose the eigenvector into corresponding three parts and find the inner product of each part with $\mathbbm{1}$ is close to 1, i.e., $0.9891\pm0.0349$, $0.9982\pm0.0009$, and $0.9992\pm0.0003$, respectively, computed from 50 independent trials, where in each trial we randomly selected 500 images. 
The   spectral gap for MNIST is also displayed in Fig. \ref{MNISTverify} and the inner product between the eigenvector of the maximum eigenvalue and $\mathbbm{1}$ is shown in the appendix. Also note that the second eigen direction can sometimes be observed, such as the case in Fig. \ref{fig:tanhathreeccross}(a).

\section{Conclusion}
In this work, our experiments, supported by theoretical analysis, demonstrate the phenomenon of condensation in CNNs during training with small initialization. Specifically, we observe that the kernels in the same layer evolve towards one or few directions.
This initial condensation of kernels resets the neural network to a simpler state with few distinct kernels. This process plays a crucial role in the subsequent learning process of the neural network.

\section{Acknowledgements}
This work is sponsored by the National Key R\&D Program of China  Grant No. 2022YFA1008200, the National Natural Science Foundation of China Grant No. 62002221, Shanghai Municipal of Science and Technology Major Project No. 2021SHZDZX0102, and the HPC of School of Mathematical Sciences and the Student Innovation Center, and the Siyuan-1 cluster supported by the Center for High Performance Computing at Shanghai Jiao Tong University.





\bibliographystyle{elsarticle-num-names}
\bibliography{DLRef.bib}

\clearpage

\appendix
\section{experiments}
\subsection{Experimental setup}

For the MNIST dataset: We use the convolution neural network with the structure: $n\times32C\text{-}(1024)\text{-}d$. The output dimension $d$ is $1$ or $10$, with respect to the loss function MSE loss or cross-entropy. The parameters of the convolution layer is initialized by the Gaussian $(0,\sigma_1^2)$, and the parameters of the linear layer is Gaussian ($0,\sigma_2^2)$. $\sigma_1$ is given by $(\frac{(c_{in}+c_{out})*m^2}{2})^{-\gamma}$ where $c_{in}$ and $c_{out}$ are the number of in channels and out channels respectively, $\sigma_2$ is given empirically by $0.0001$. The data size is $n$ which is randomly chosen from the MNIST dataset. The training method is GD or Adam with full batch.
\subsection{CIFAR10 examples}

\begin{figure}[htb]
    \centering
    \includegraphics[width=0.8\textwidth]{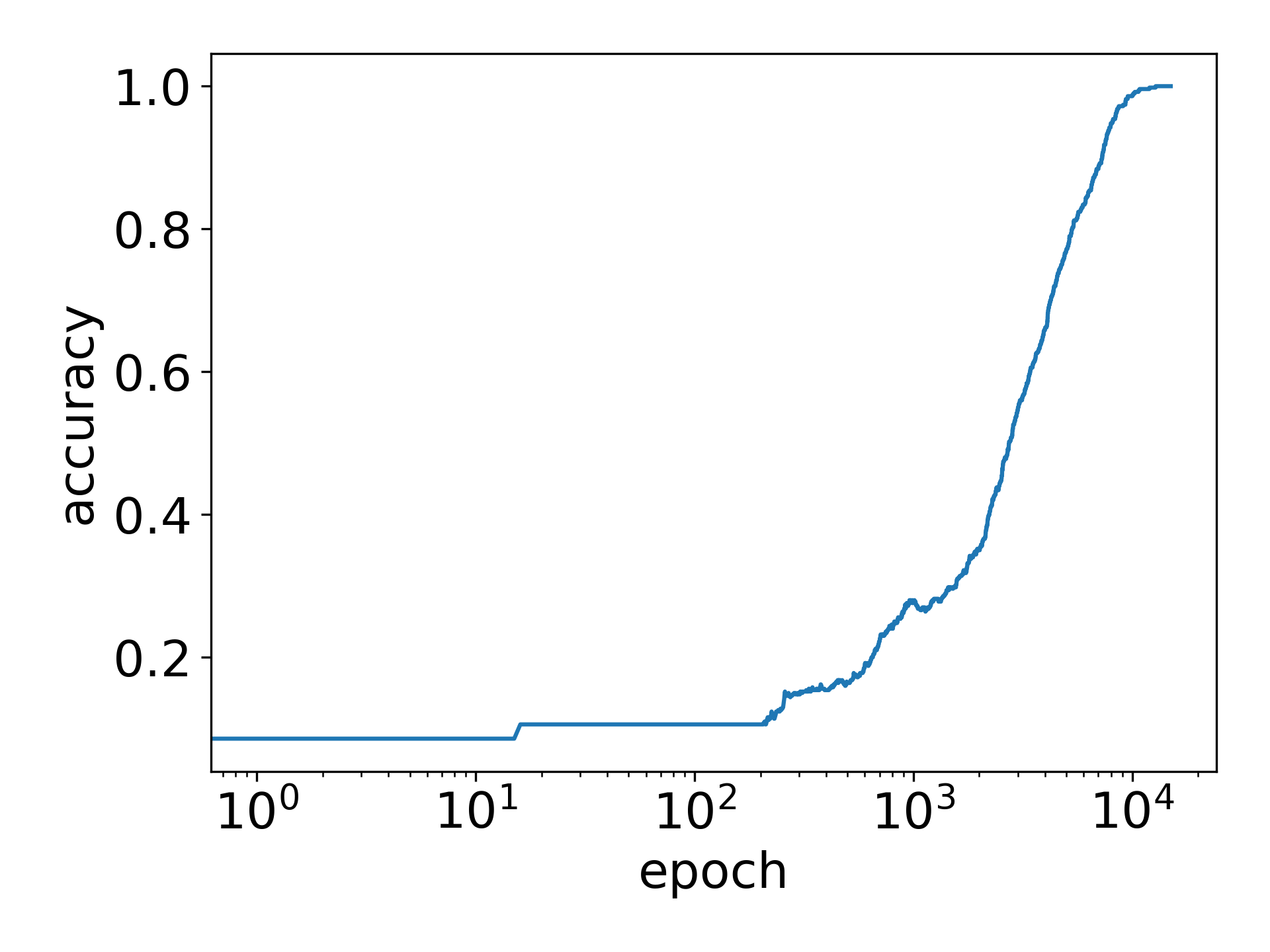}
    \caption{The training accuracy of the experiment in Fig: \ref{fig:tanhathreeccrossfinal}}
    \label{fig:tanhathreeccrossfinalacc}
\end{figure}


\begin{figure}
     \centering
     \begin{subfigure}[b]{0.3\textwidth}
         \centering
         \includegraphics[width=\textwidth]{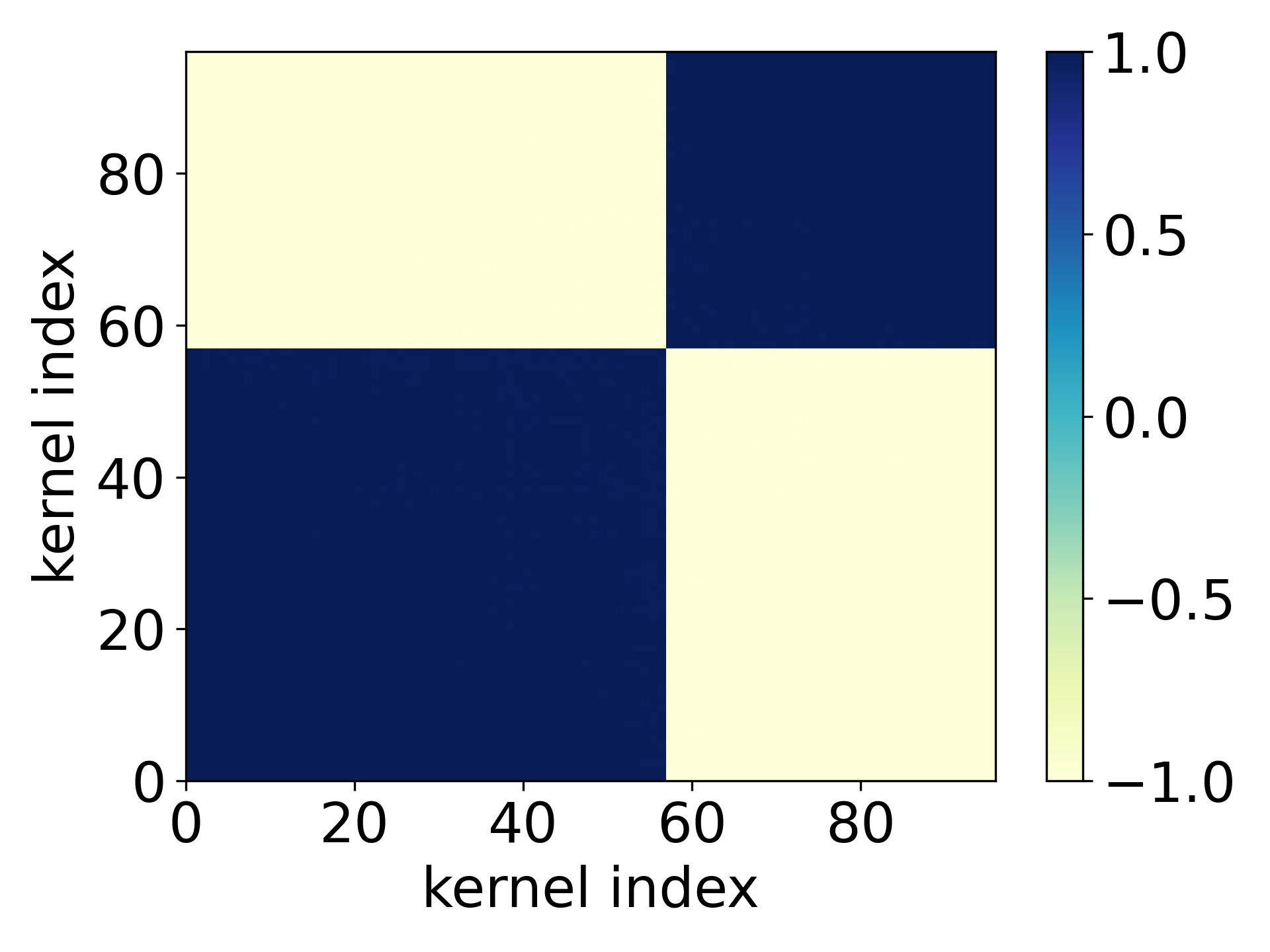}
         \caption{ layer $1$}
         \label{tanha1lcMSE}
     \end{subfigure}
     \hfill
     \begin{subfigure}[b]{0.3\textwidth}
         \centering
         \includegraphics[width=\textwidth]{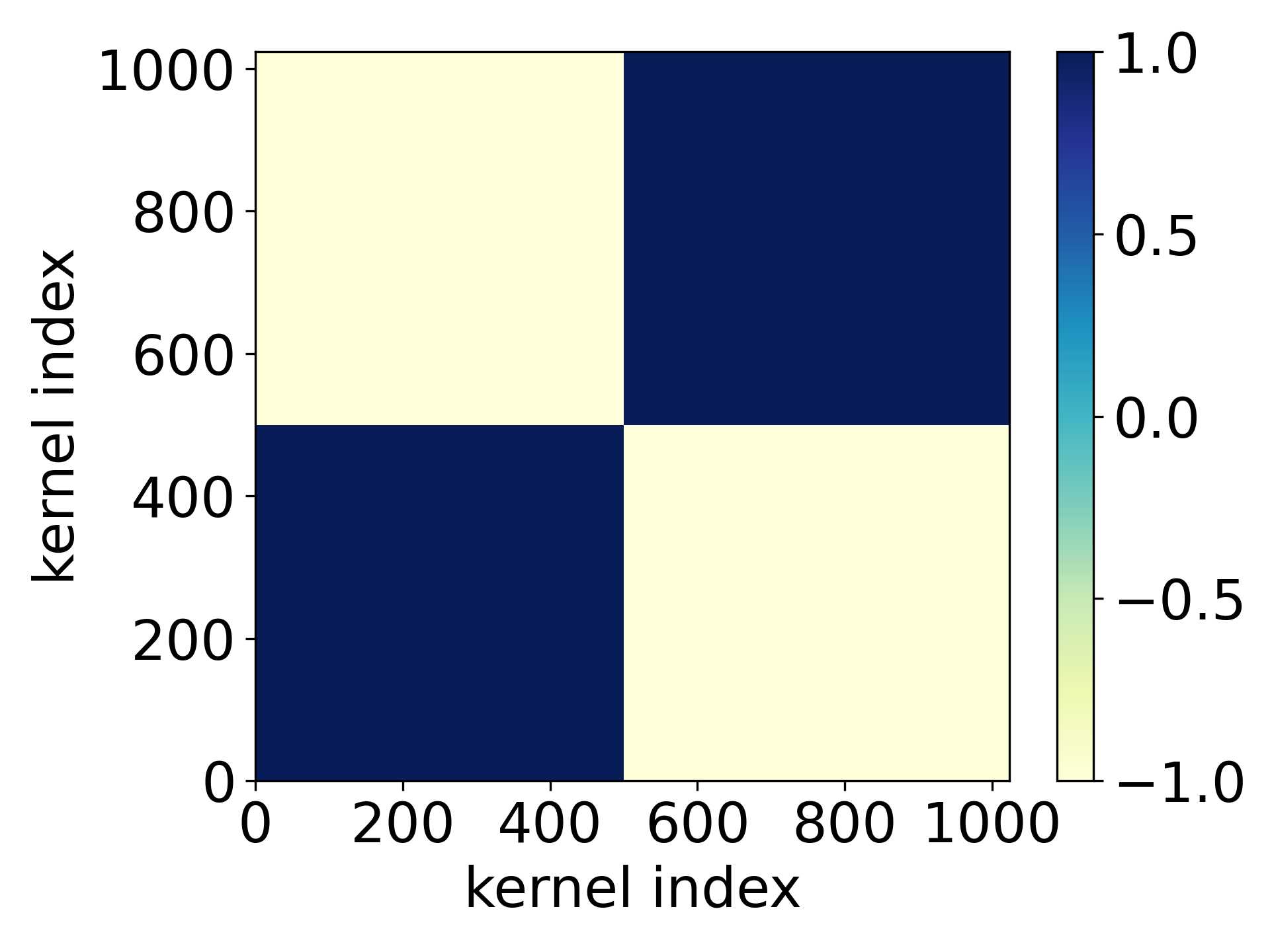}
         \caption{layer $2$}
         \label{tanha2lcMSE}
     \end{subfigure}
     \hfill
     \begin{subfigure}[b]{0.3\textwidth}
         \centering
         \includegraphics[width=\textwidth]{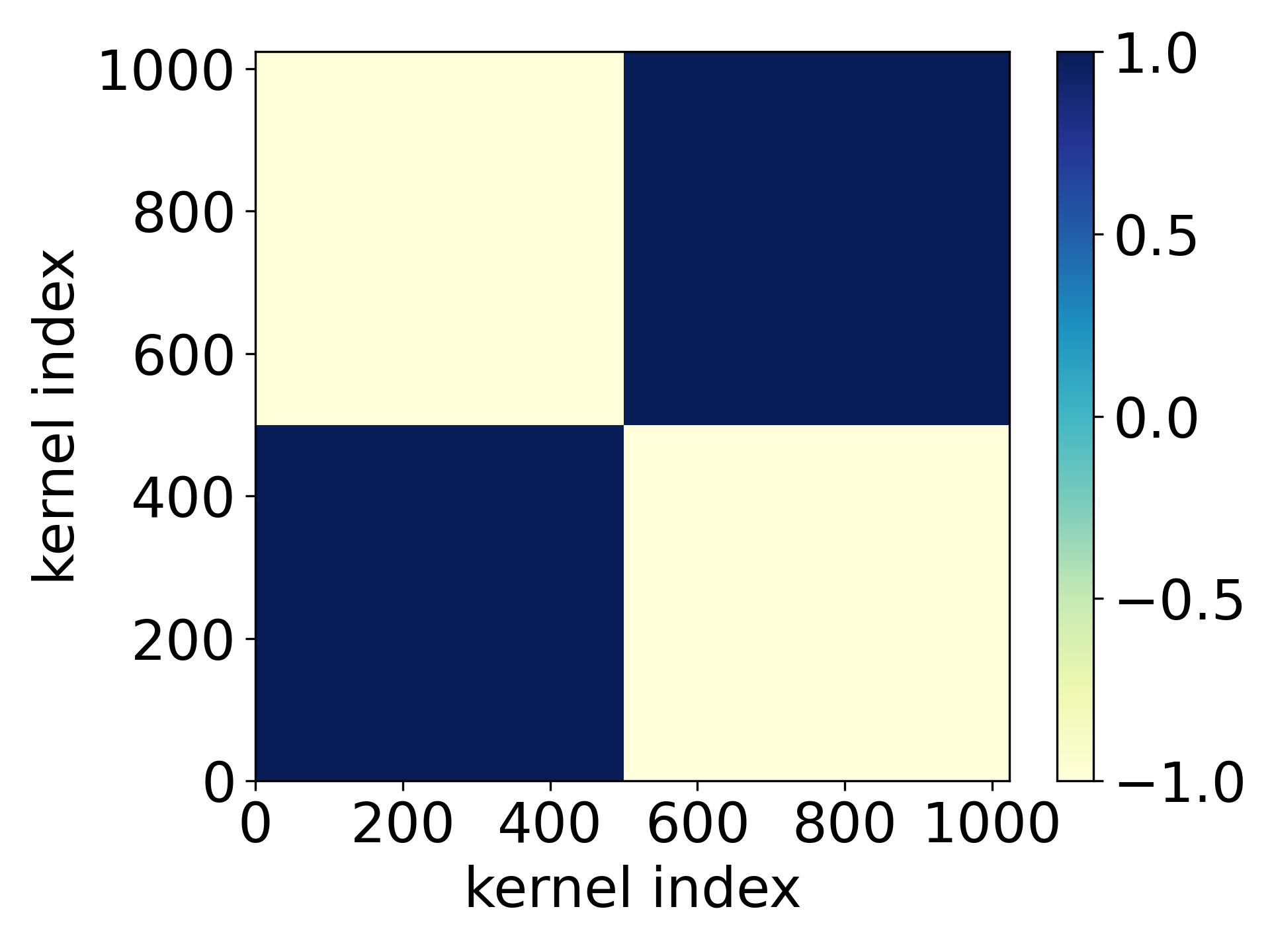}
         \caption{layer $3$}
         \label{tanha3lcMSE}
     \end{subfigure}
        \caption{Condensation of convolution network with three convolution kernel training on CIFAR10 dataset with data size= $500$. The kernel size $m=3$. The colors in the figure show the cosine similarity of normalized weight vectors of each convolution kernel. The activation for all convolution layers are $\rm {tanh}(x)$. The number of the steps are at epoch $=200$, epoch $=200$ and epoch $=200$. The convolution kernels are initialized by $\gamma=2$. The learning rate is $5\times 10^{-6}$. We use one dimension output instead of label and use MSE loss. The optimizer is full batch Adam. }
        \label{Fig:tanhathreecMSE}
\end{figure}
\subsection{MNIST examples}
When we are training the MNIST dataset, we use MSE loss with one dimension output as the criterion.

\begin{figure}
     \centering
     \begin{subfigure}[b]{0.3\textwidth}
         \centering
         \includegraphics[width=\textwidth]{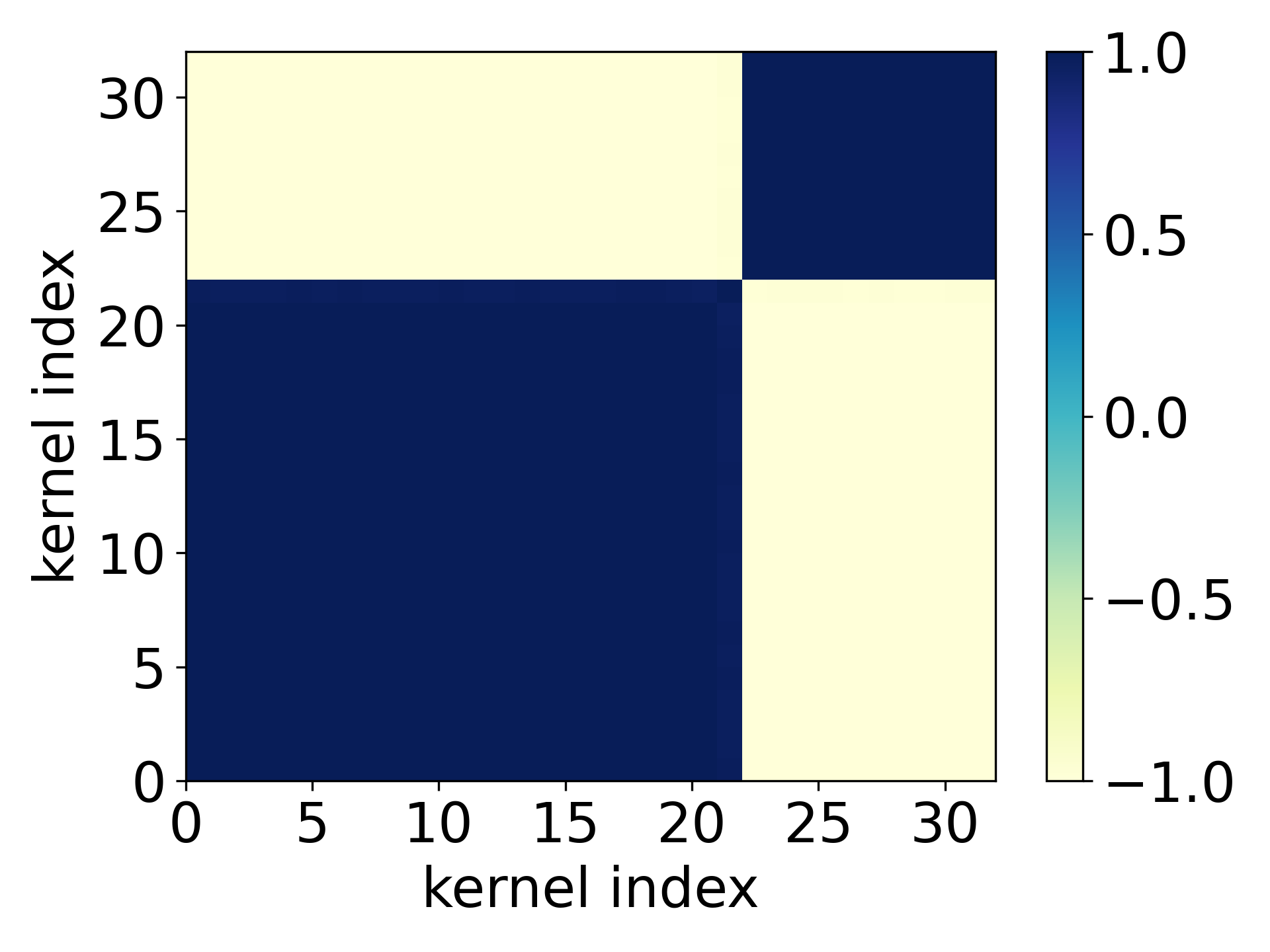}
         \caption{${\rm ReLU}(x)$}
         \label{ReLU1c}
     \end{subfigure}
     \hfill
     \begin{subfigure}[b]{0.3\textwidth}
         \centering
         \includegraphics[width=\textwidth]{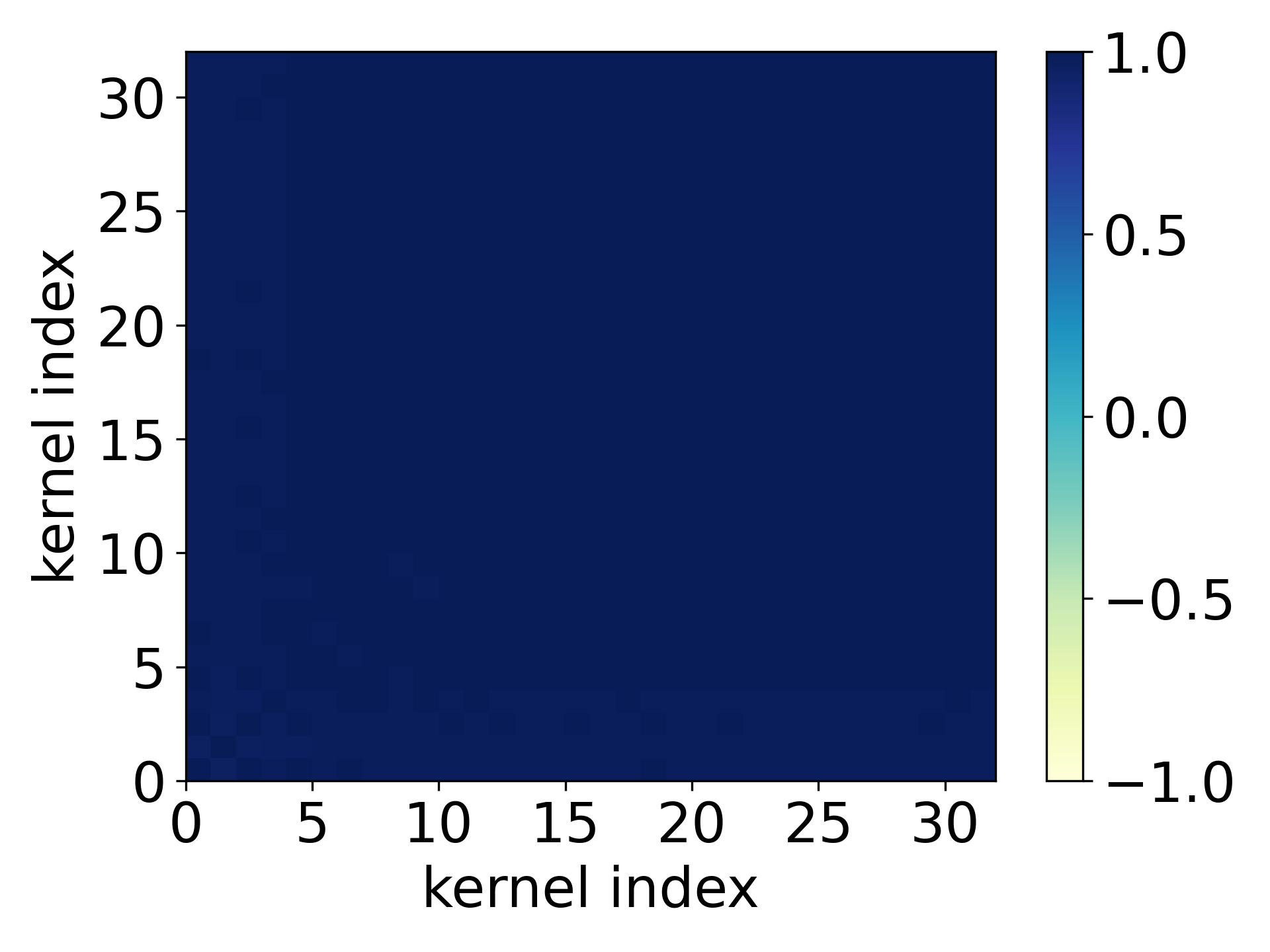}
         \caption{$\rm{Sigmoid}(x)$}
         \label{sigmoid1c}
     \end{subfigure}
     \hfill
     \begin{subfigure}[b]{0.3\textwidth}
         \centering
         \includegraphics[width=\textwidth]{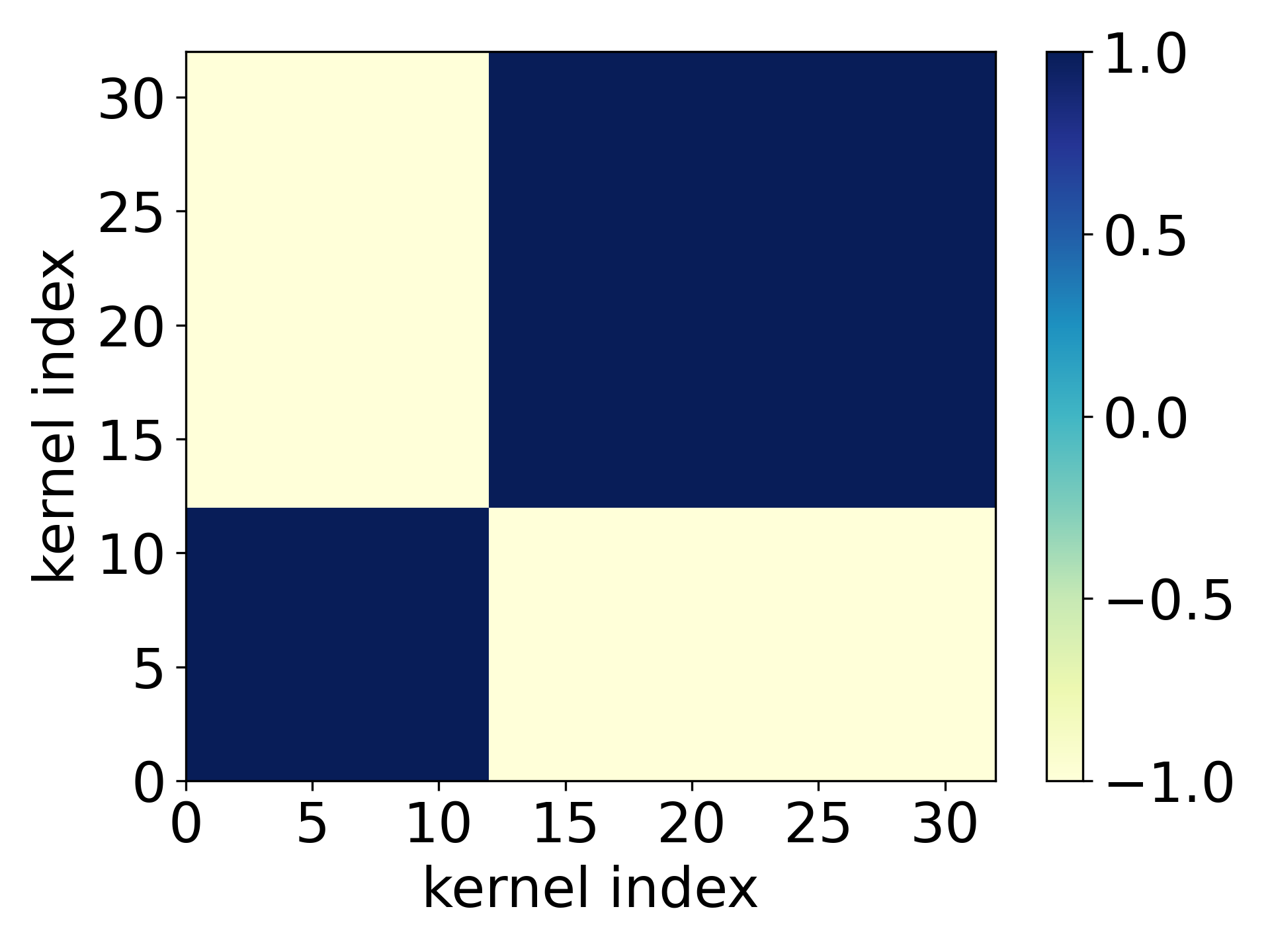}
         \caption{$\rm{tanh}(x)$}
         \label{tanh1c}
     \end{subfigure}
        \caption{Condensation of CNNs with different activations (indicated by sub-captions) for convolutional layers. The network has $32$ kernels in the convolution layer, followed by $1$-dimensional output. The kernel size is  $m=5$. The learning rate is $5\times 10^{-7}$, $5\times 10^{-7}$ and $5\times 10^{-6}$ separately. The number of the selected steps are at epoch $=252$, epoch $=302$, and epoch $=200$. The convolution layer is initialized by $\gamma=2$. We use full batch Adam optimizer with MSE loss on MNIST dataset.}
        \label{Fig:1c}
\end{figure}

We display the initial condensation of the convolution neural network with one convolution layer on the MNIST dataset whose data size is $n=500$. The colors in Fig.\ref{Fig:1c} show the cosine similarity $D(u,v)$ of different kernel in one convolution layer. Yellow $D(u,v)\sim 1$(purple $D(u,v)\sim -1$), means the kernel weight vectors are at the same(opposite) directions,  The activation function of the convolution layer are $\rm{Sigmoid}(x)$, ${\rm ReLU}(x)$ and $\rm{tanh}(x)$. The Fig.\ref{tanh1c} implies that during the initial stage of the training, the convolution kernel in each layer will condense at two opposite directions when the activation function is $\rm{tanh}(x)$. This phenomenon will also happen when the activation functions are ${\rm ReLU}(x)$ and $\rm{Sigmoid}(x)$ which is shown in Fig.\ref{ReLU1c} and Fig.\ref{sigmoid1c}.

\begin{figure}
     \centering
     \begin{subfigure}[b]{0.3\textwidth}
         \centering
         \includegraphics[width=\textwidth]{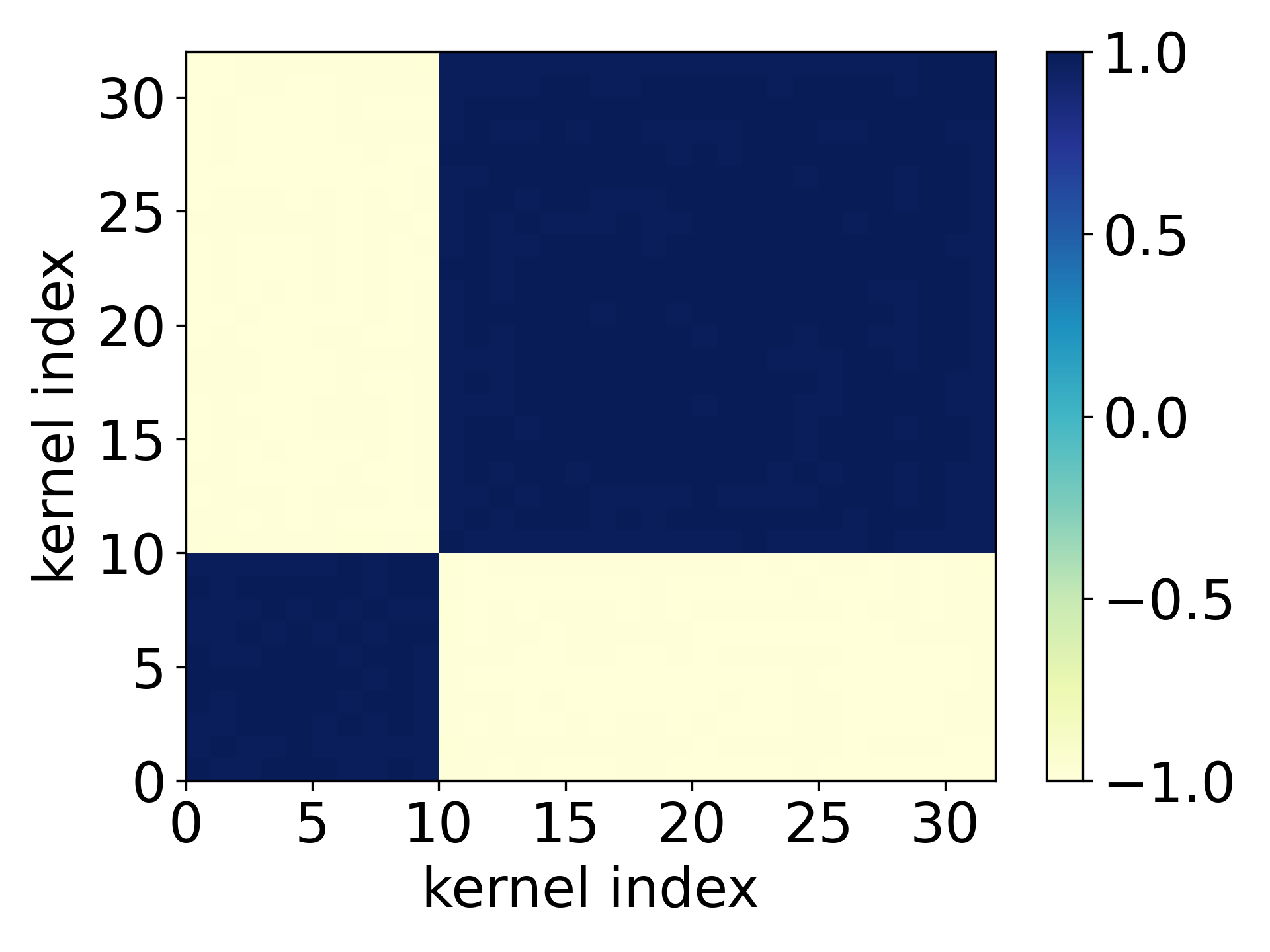}
         \caption{ layer $1$}
         \label{tanha1l}
     \end{subfigure}
     \hfill
     \begin{subfigure}[b]{0.3\textwidth}
         \centering
         \includegraphics[width=\textwidth]{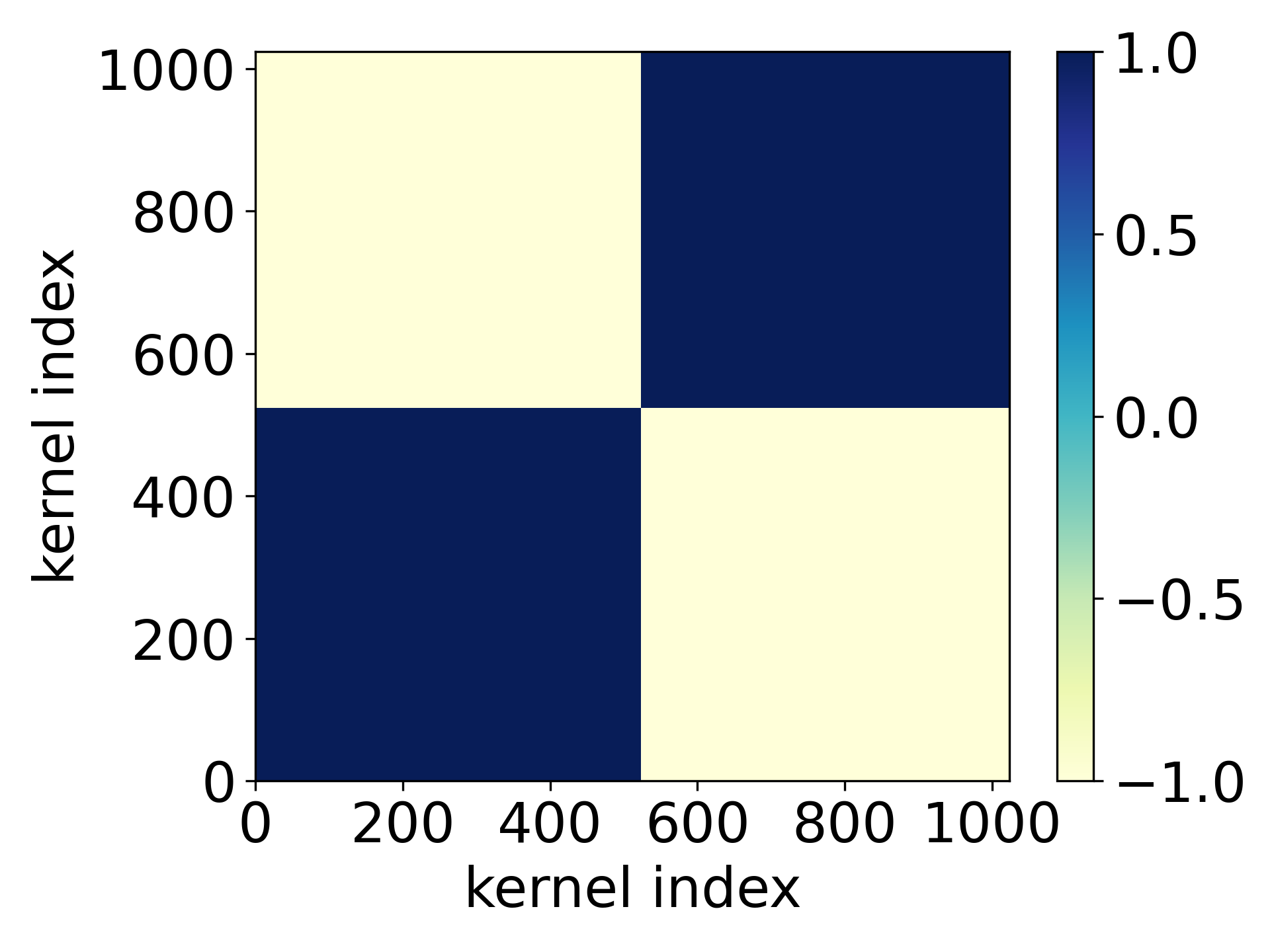}
         \caption{layer $2$}
         \label{tanha2l}
     \end{subfigure}
     \hfill
     \begin{subfigure}[b]{0.3\textwidth}
         \centering
         \includegraphics[width=\textwidth]{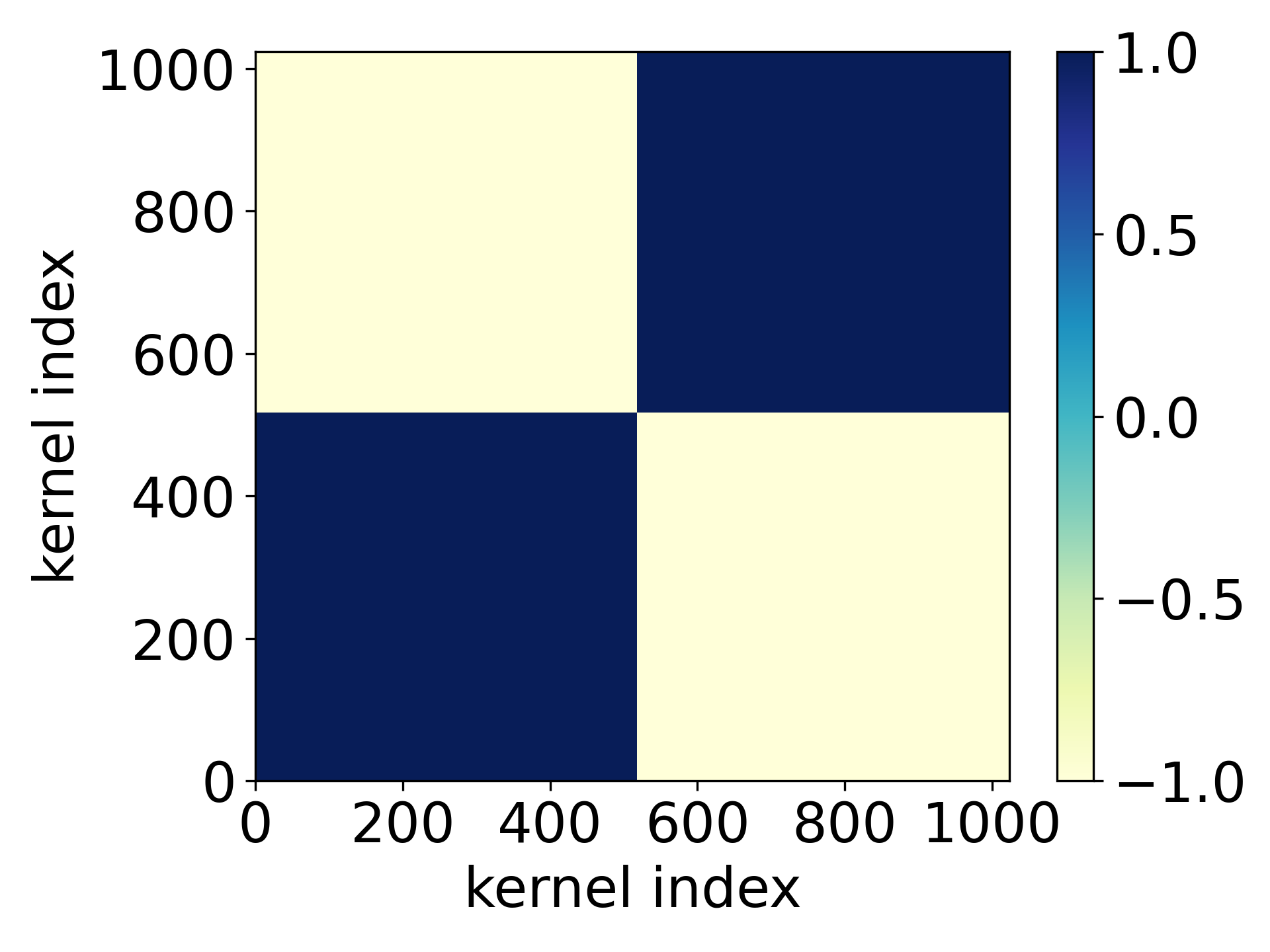}
         \caption{layer $3$}
         \label{tanha3l}
     \end{subfigure}
        \caption{Condensation of ${\rm tanh}$ CNN  with three convolution layer by MSE loss. The kernel size is $m=5$. The color indicates the cosine similarity between kernels. The number of the steps are all at epoch $=100$. The learning rate is $5\times10^{-7}$. The convolution layer is initialized by $\gamma=2$. We use $1$-dimension output. The optimizer is full batch Adam on MNIST dataset.}
        \label{Fig:tanhathree}
\end{figure}

Condensation of multi-layer convolution neural network on MNIST dataset is shown in Fig.\ref{Fig:tanhathree}. The activation function between each convolution layer is $\rm{tanh}(x)$. We have the kernel weight of different layer are all condensation on two opposite directions, i.e. one line.

\begin{figure}
    \centering
    \begin{subfigure}[b]{0.4\textwidth}
        \centering
        \includegraphics[width=\textwidth]{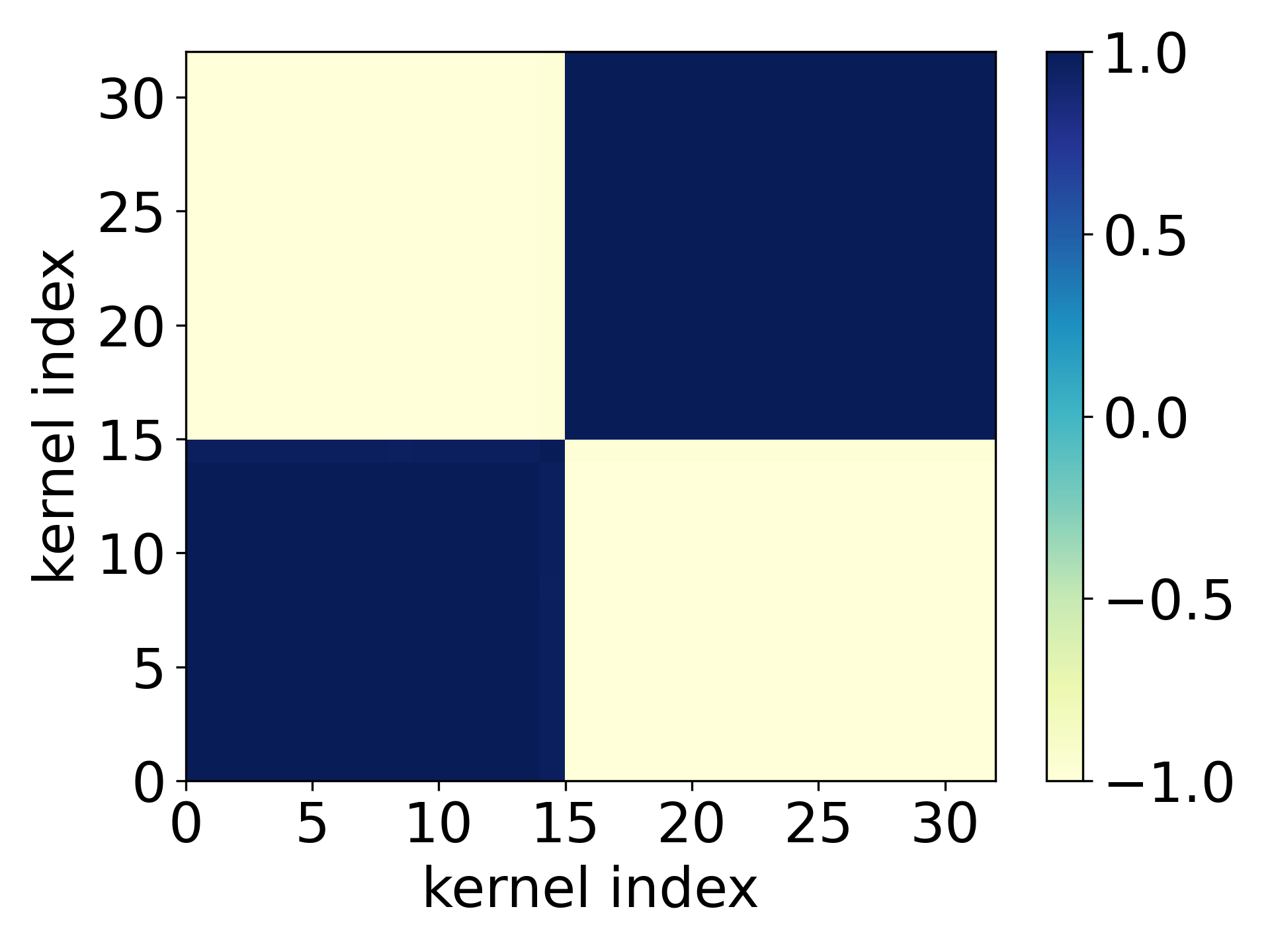}
        \caption{condensation}
        \label{GDMNISTc}
    \end{subfigure}
    \hfill
    \begin{subfigure}[b]{0.4\textwidth}
        \centering
        \includegraphics[width=\textwidth]{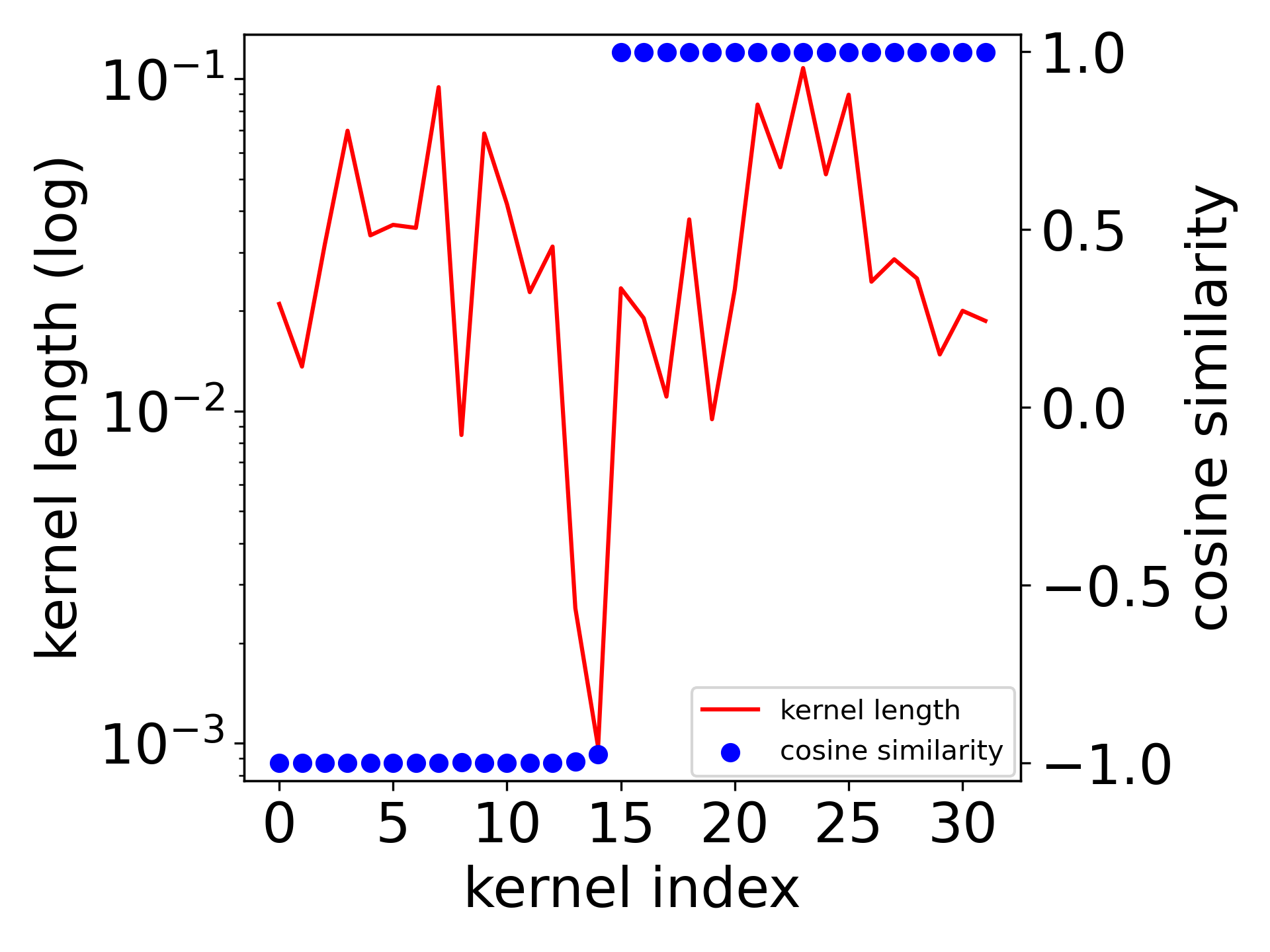}
        \caption{directions}
        \label{GDMNISTd}
    \end{subfigure}
    \caption{Condensation of two-layer CNN by GD and MSE training on MNIST dataset with data size $n=500$. The network has $32$ kernels. (a) consine similarity. (b) left ordinate (red): the amplitude of each kernel; (b) right ordinate: cosine similarity between each kernel and $\mathbbm{1}$. The activation function of the convolution part is ${\rm tanh}(x)$. The kernel size is $m=3$. The learning rate is $5\times 10^{-6}$. The number of the selected steps is epoch $=3600$. The convolution layer is initialized by $\gamma=2$. }
    \label{Fig:tanh1cGD}
\end{figure}

\begin{figure}
    \centering
    \begin{subfigure}[b]{0.4\textwidth}
        \centering
        \includegraphics[width=\textwidth]{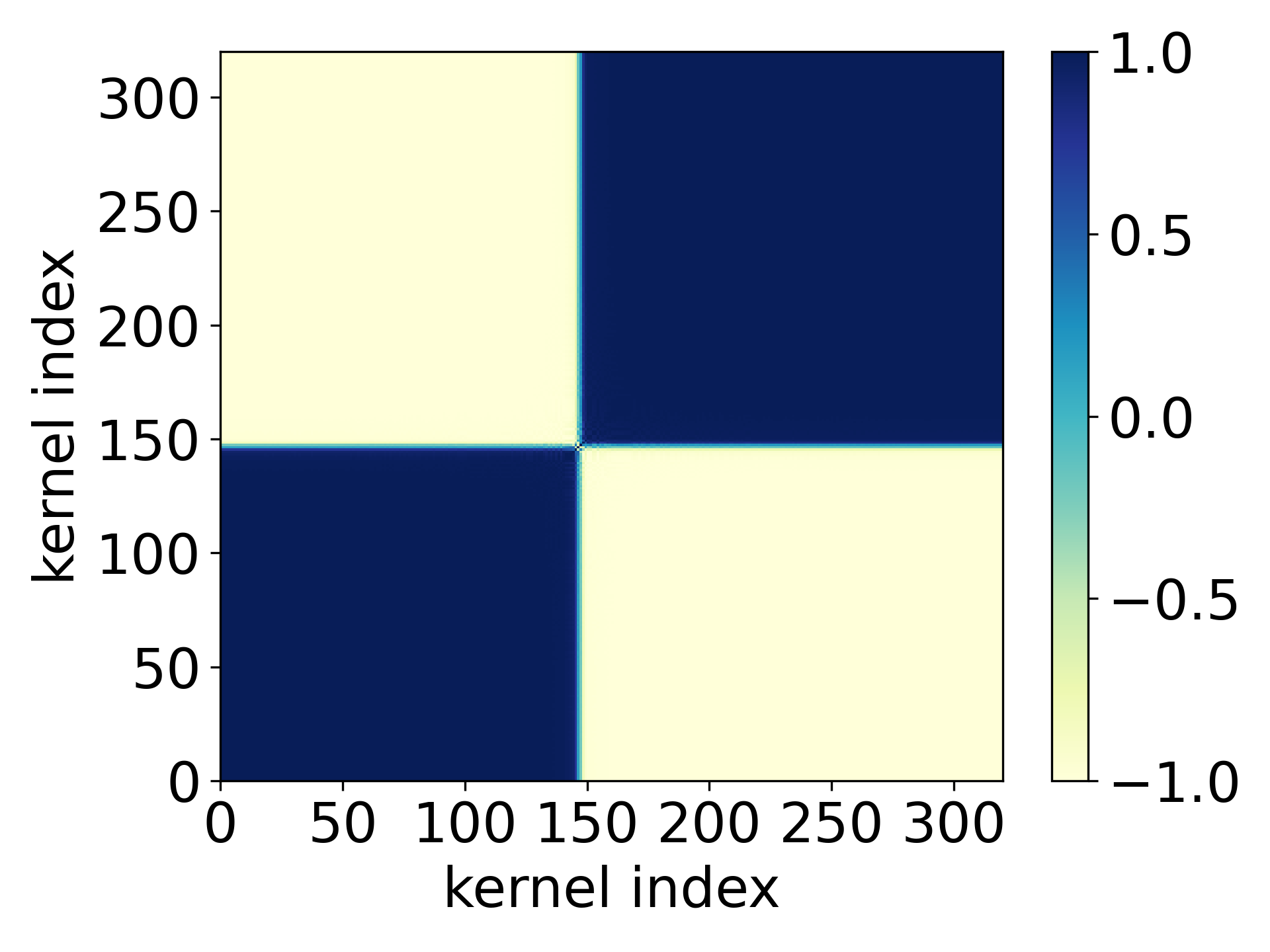}
        \caption{condensation}
        \label{GDMNISTc320}
    \end{subfigure}
    \hfill
    \begin{subfigure}[b]{0.4\textwidth}
        \centering
        \includegraphics[width=\textwidth]{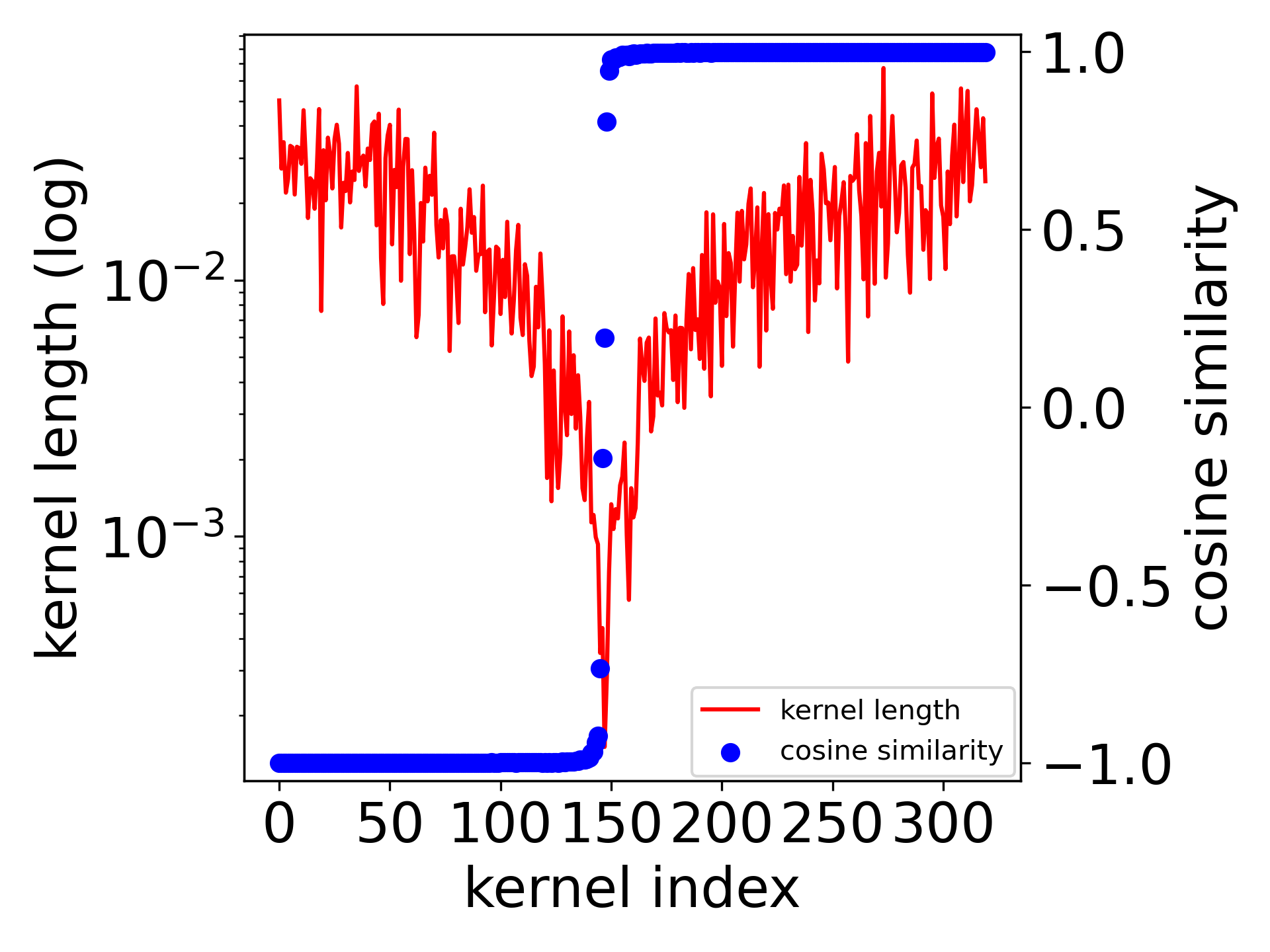}
        \caption{directions}
        \label{GDMNISTd320}
    \end{subfigure}
    \caption{Condensation of two-layer CNN by GD and MSE training on MNIST dataset with data size $n=500$.The network has $320$ kernels. The activation function of the convolution part is ${\rm tanh}(x)$. The kernel size is $m=3$. The learning rate is $5\times 10^{-6}$. The number of the selected steps is epoch $=7000$. The convolution layer is initialized by $\gamma=2$. }
    \label{Fig:tanh1cGD320}
\end{figure}

\begin{figure}[htb]
    \centering
        \centering
        \includegraphics[width=0.6\textwidth]{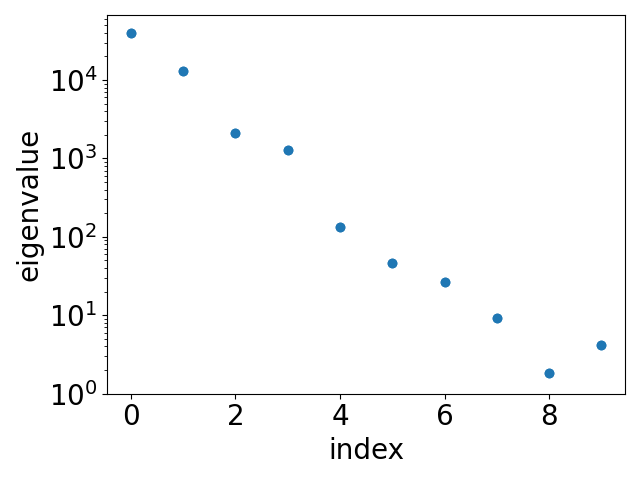}

        \label{eig}

    \caption{The largest $10$ eigenvalues of $\mZ^\T\mZ$ in tanh(x) CNN on MNIST dataset. A clear spectral gap ($\Delta\lambda := \lambda_1-\lambda_2$) could be observed, which satisfies \Cref{assump...SpectralGap-main}. }
    \label{MNISTverify}
\end{figure}

In the MNIST example, it has only one input channel, thus we directly project the first eigenvector to $\mathbbm{1}$ and find the inner product is close to $1$, i.e. $0.99974\pm0.00003$, computed from 50 independent trials, where each trail randomly selected 500 images. 



The loss at the initial stage is shown as follows:

\begin{figure}[htbp]
     \centering
    \begin{subfigure}[b]{0.3\textwidth}
        \centering
        \includegraphics[width=\textwidth]{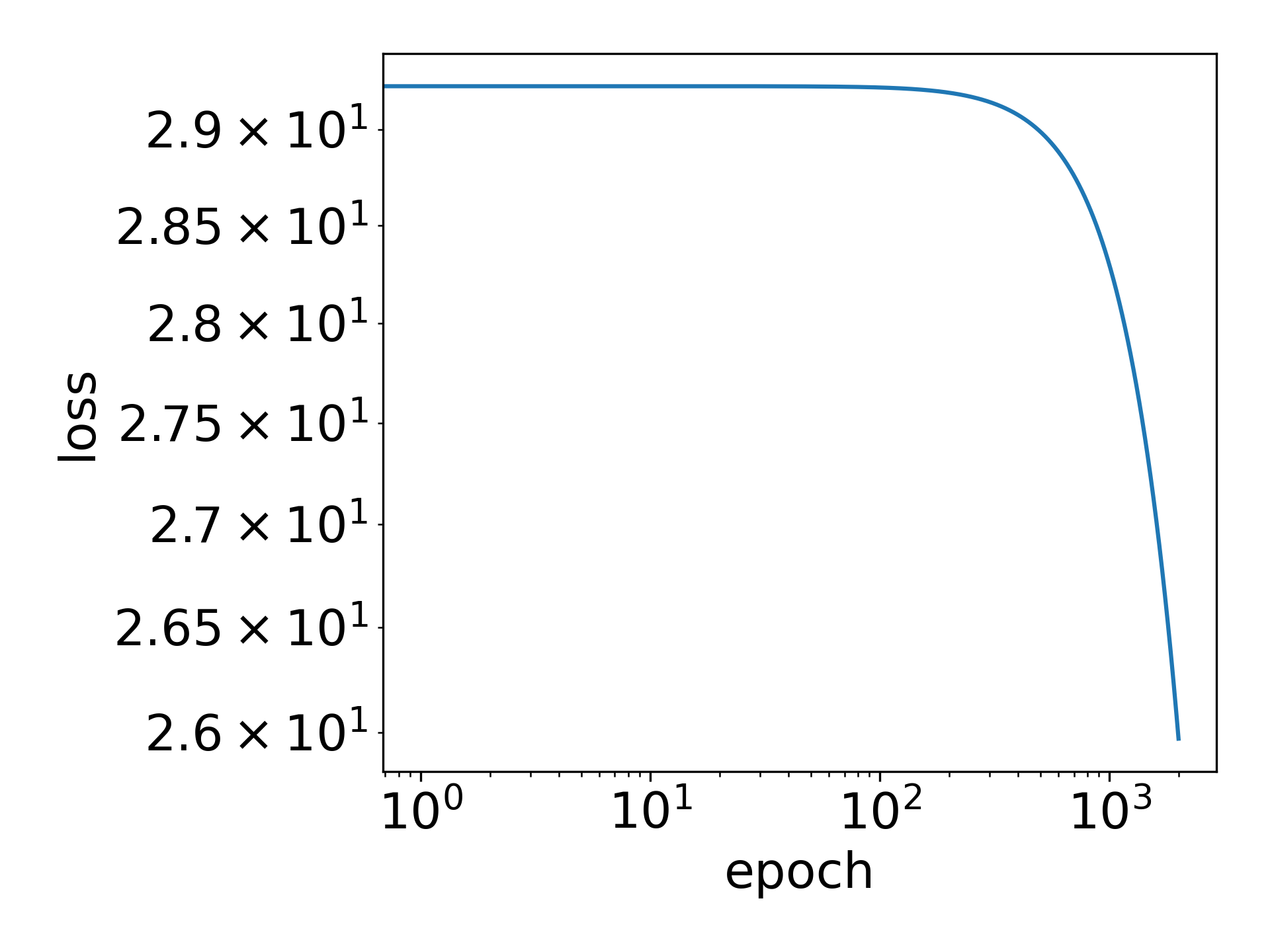}
        \caption{Fig.\ref{ReLU1c}}
    \end{subfigure}
    \hfill
     \begin{subfigure}[b]{0.3\textwidth}
         \centering
         \includegraphics[width=\textwidth]{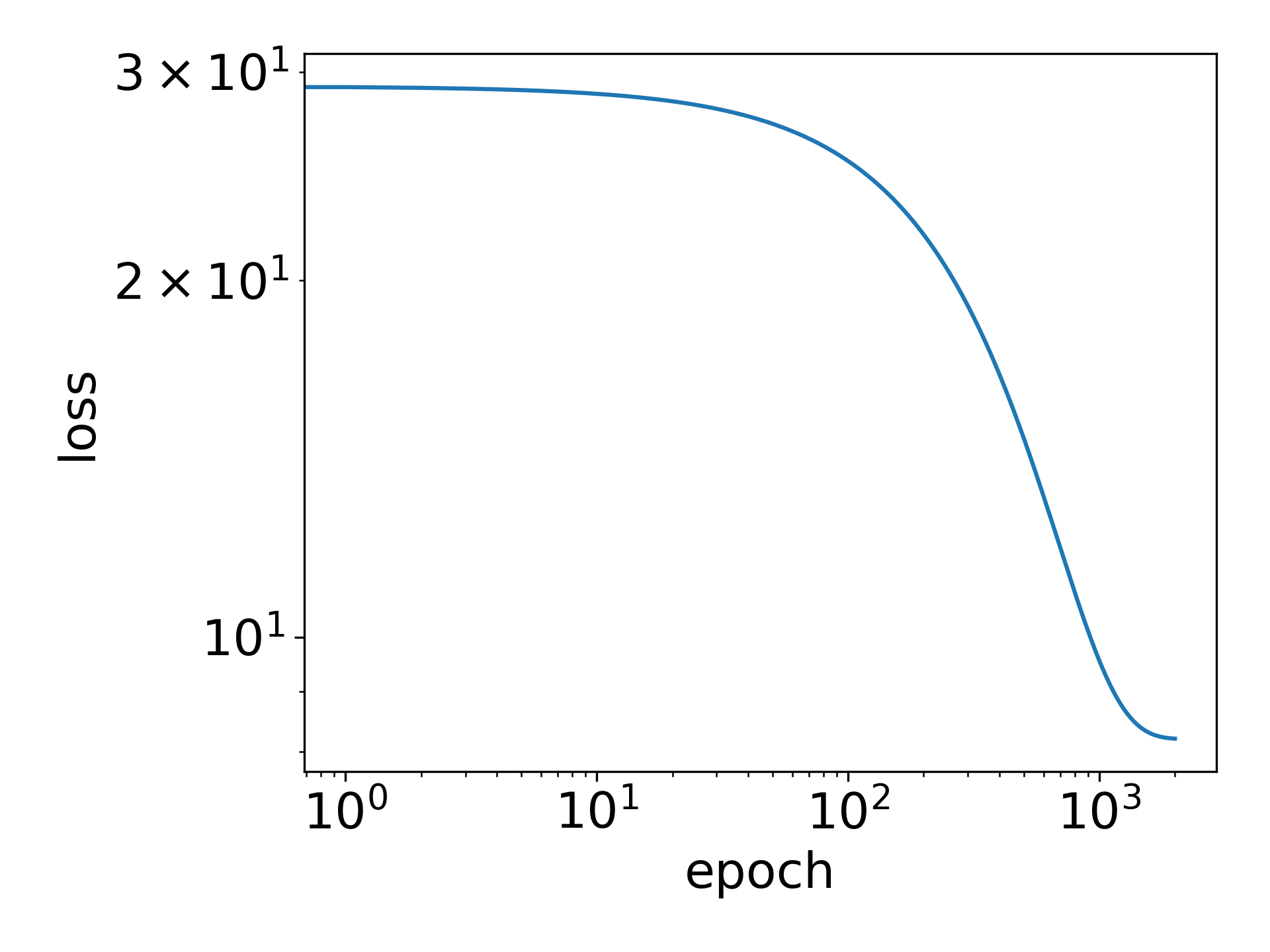}
         \caption{Fig.\ref{sigmoid1c}}
     \end{subfigure}
     \hfill
     \begin{subfigure}[b]{0.3\textwidth}
         \centering
         \includegraphics[width=\textwidth]{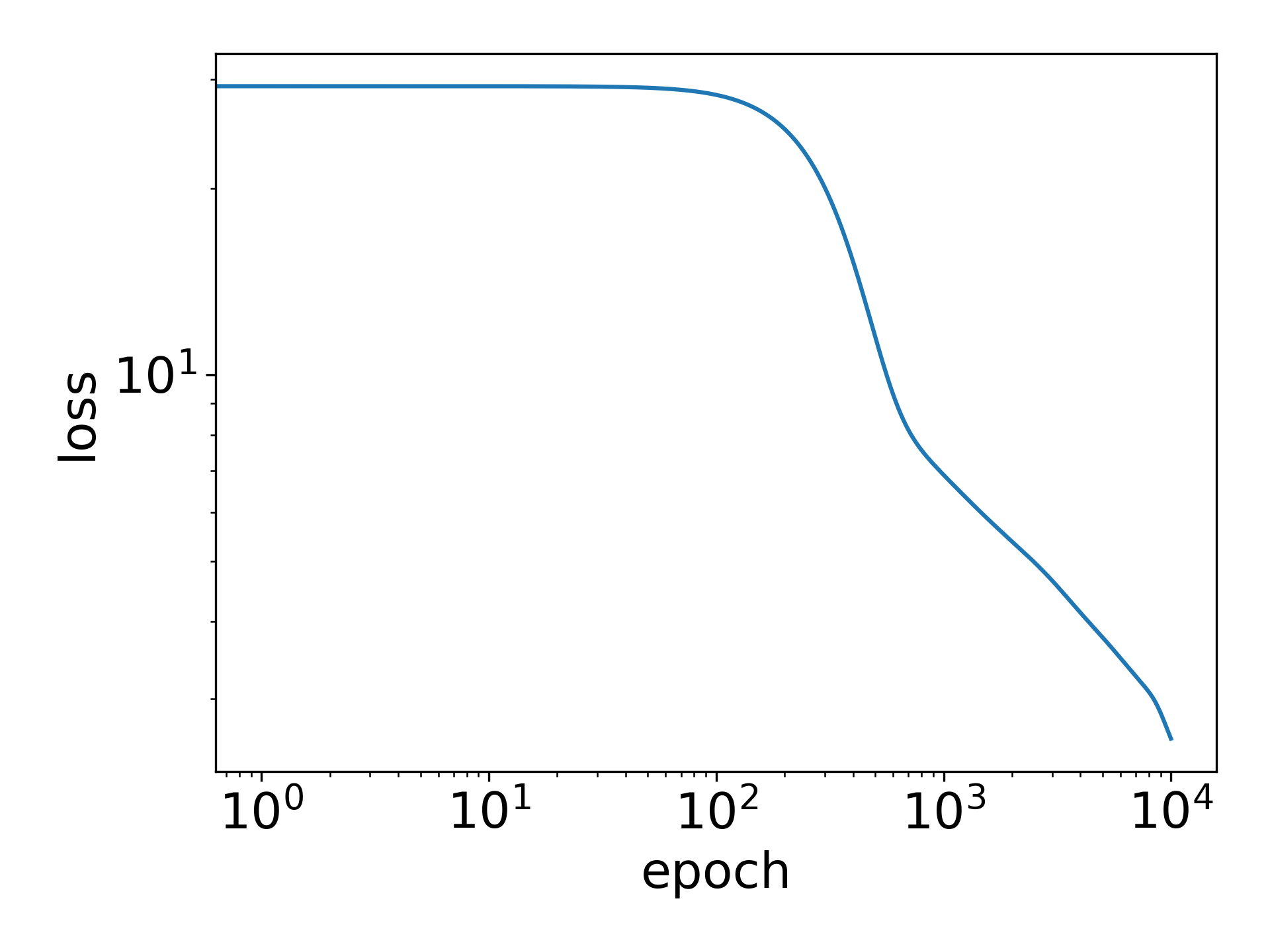}
         \caption{Fig.\ref{tanh1c}}
     \end{subfigure}

     \begin{subfigure}[b]{0.3\textwidth}
        \centering
         \includegraphics[width=\textwidth]{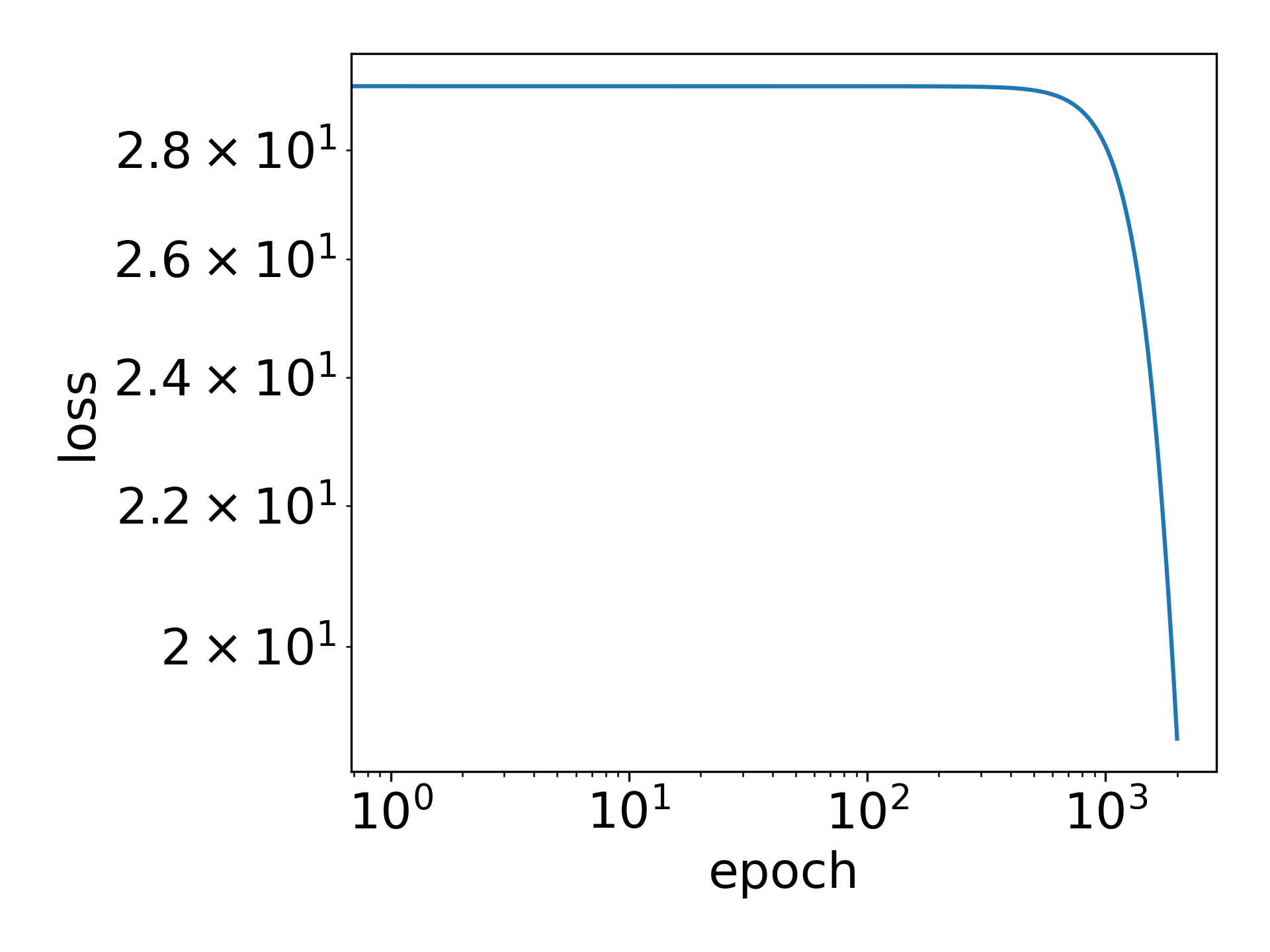}
         \caption{Fig.\ref{Fig:tanhathree}}
     \end{subfigure}
     \hfill
     \begin{subfigure}[b]{0.3\textwidth}
         \centering
         \includegraphics[width=\textwidth]{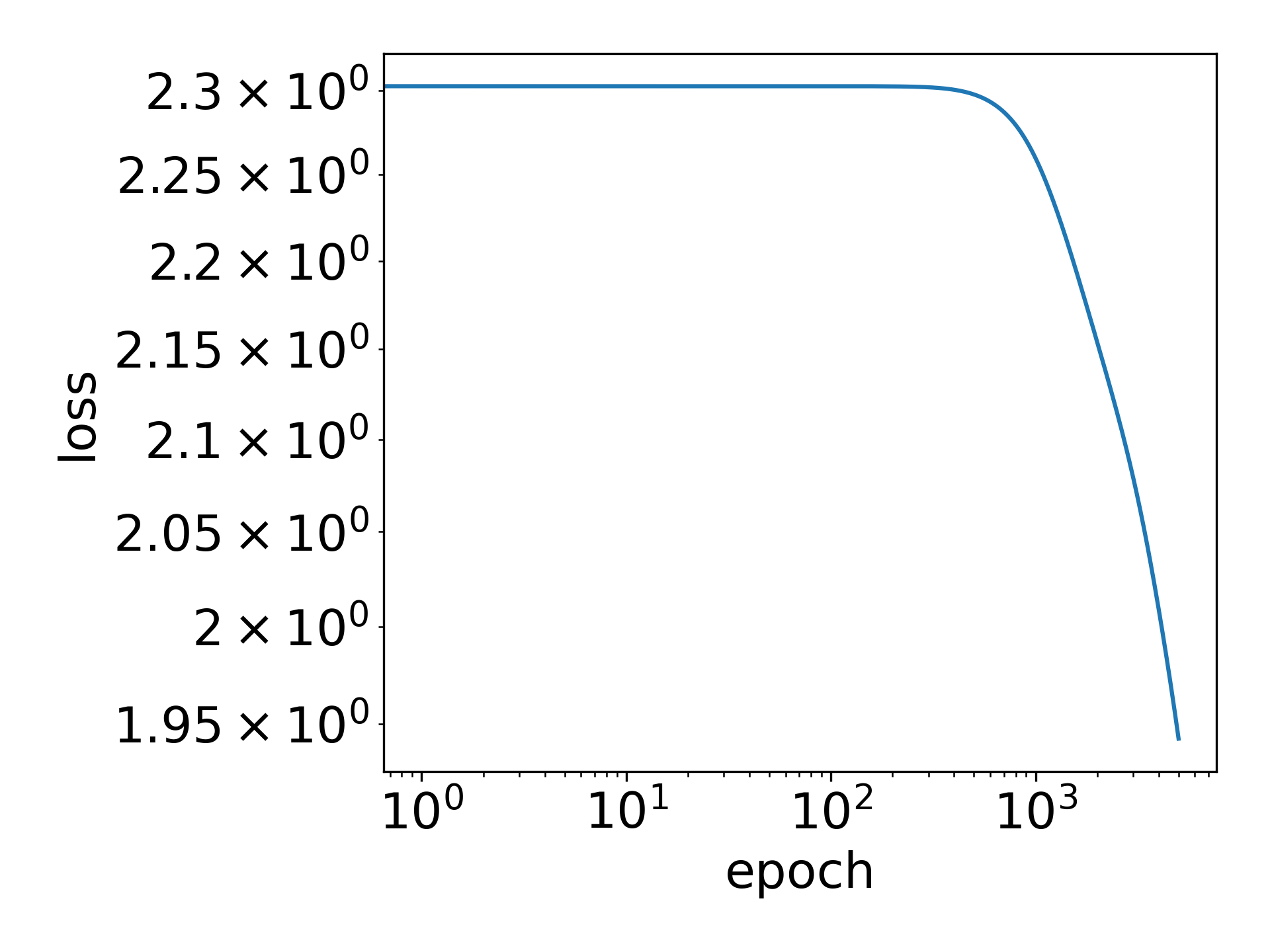}
         \caption{Fig.\ref{relua2le}}
     \end{subfigure}
     \hfill
     \begin{subfigure}[b]{0.3\textwidth}
         \centering
         \includegraphics[width=\textwidth]{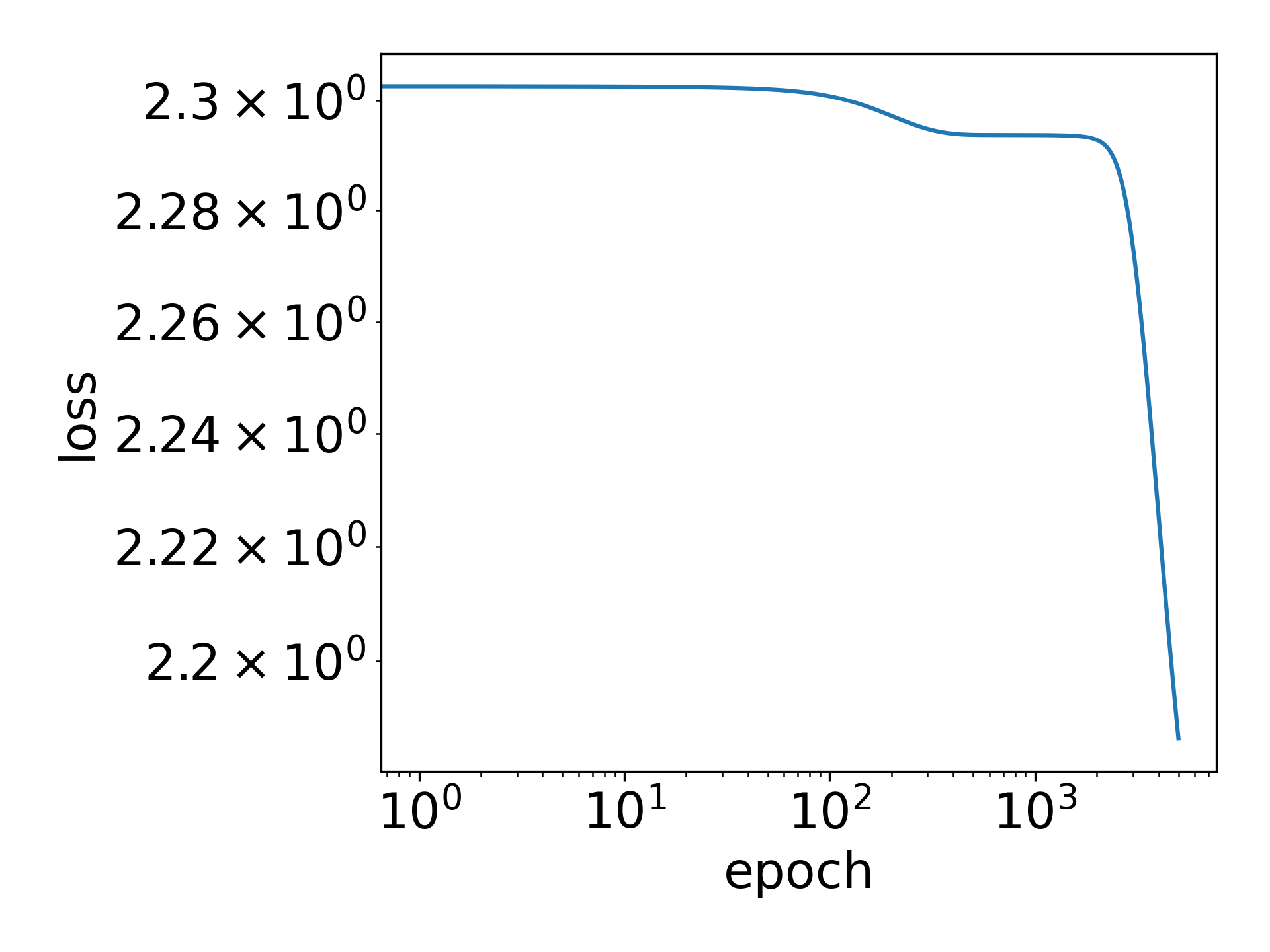}
         \caption{Fig.\ref{sigmoida2le}}
     \end{subfigure}

     \begin{subfigure}[b]{0.3\textwidth}
         \centering
         \includegraphics[width=\textwidth]{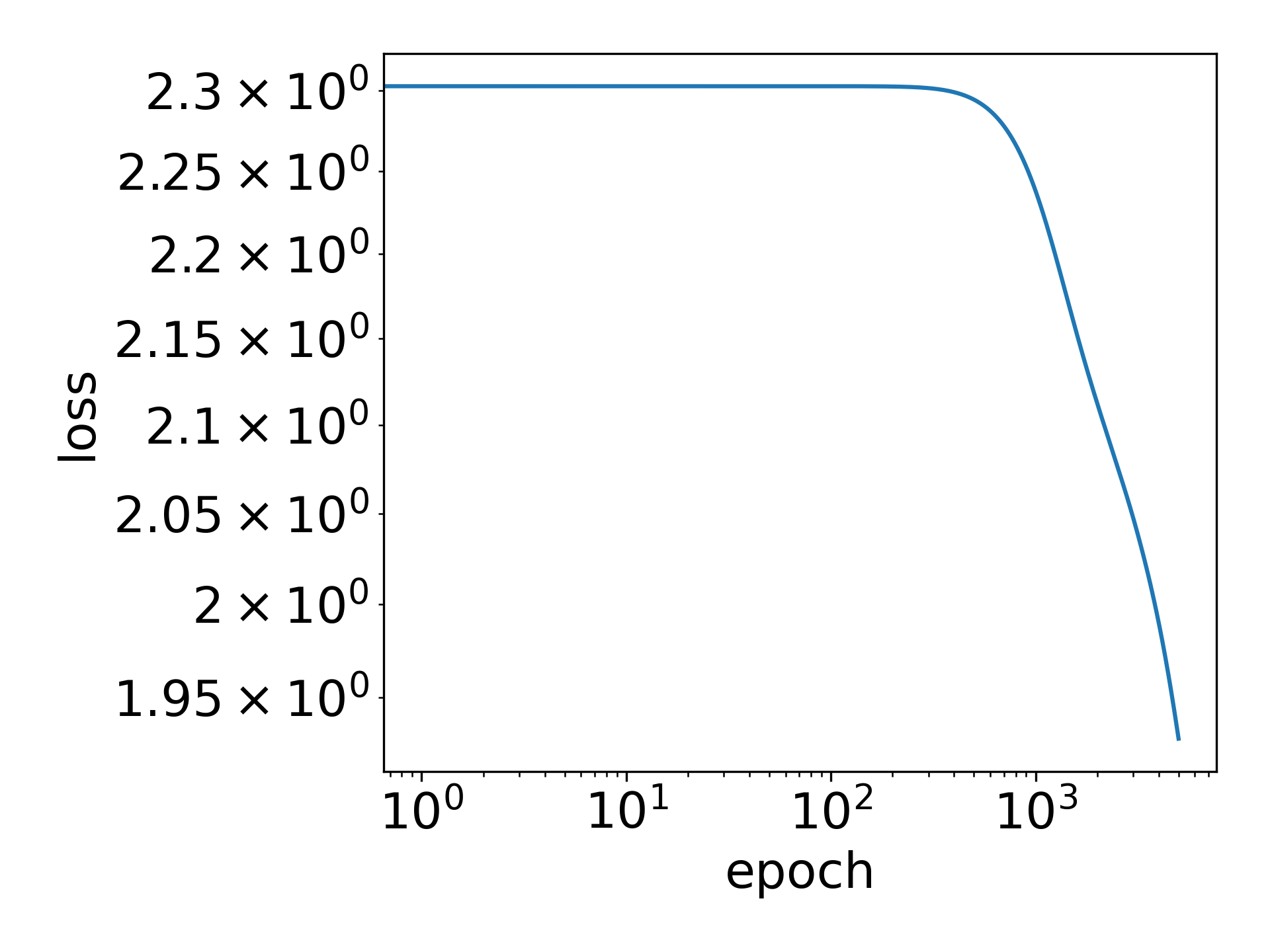}
         \caption{Fig.\ref{tanha2le}}
     \end{subfigure}
     \hfill
     \begin{subfigure}[b]{0.3\textwidth}
         \centering
         \includegraphics[width=\textwidth]{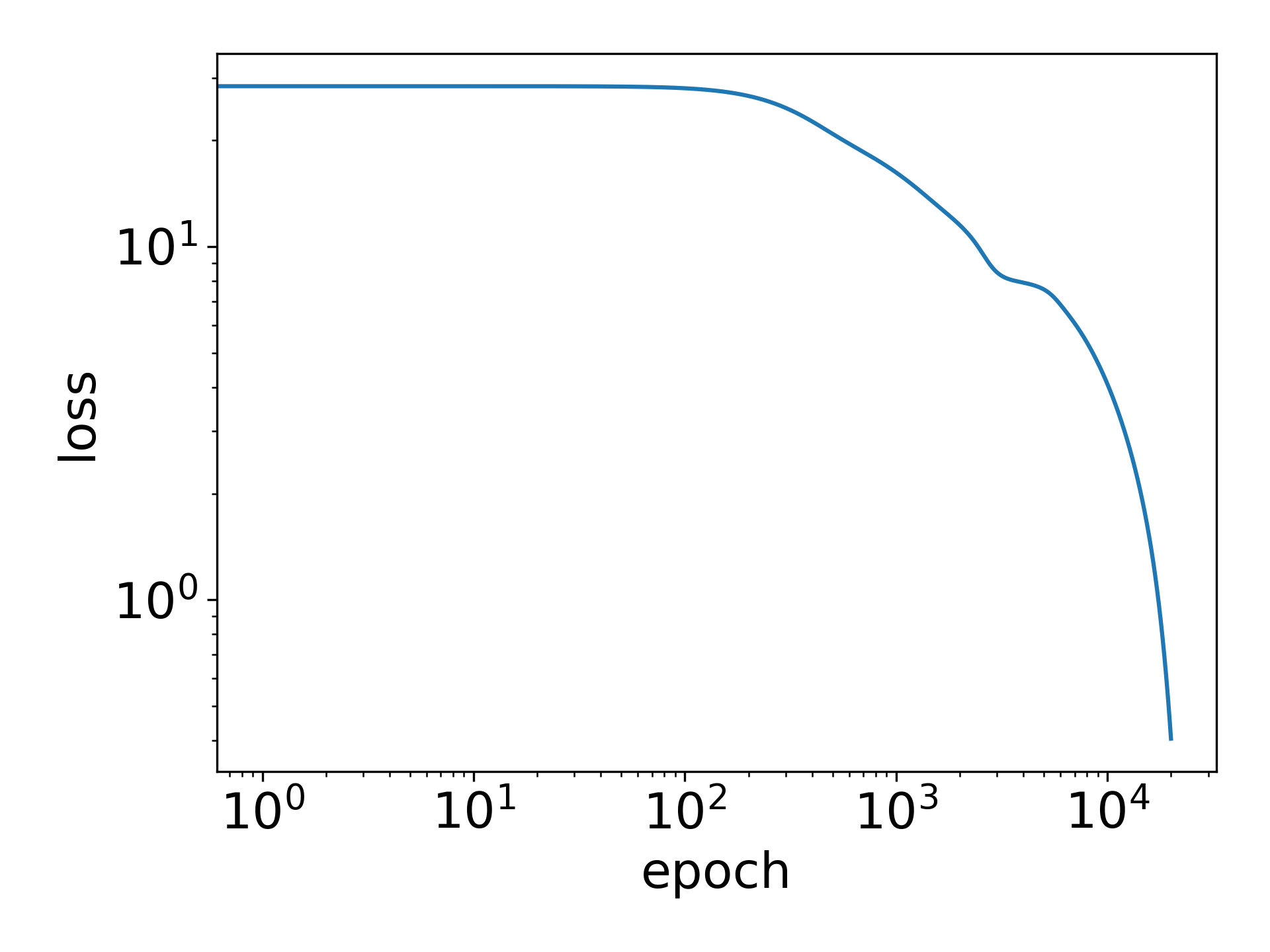}
         \caption{Fig.\ref{Fig:tanhathreecMSE}}
     \end{subfigure}
     \hfill
        \begin{subfigure}[b]{0.3\textwidth}
         \centering
         \includegraphics[width=\textwidth]{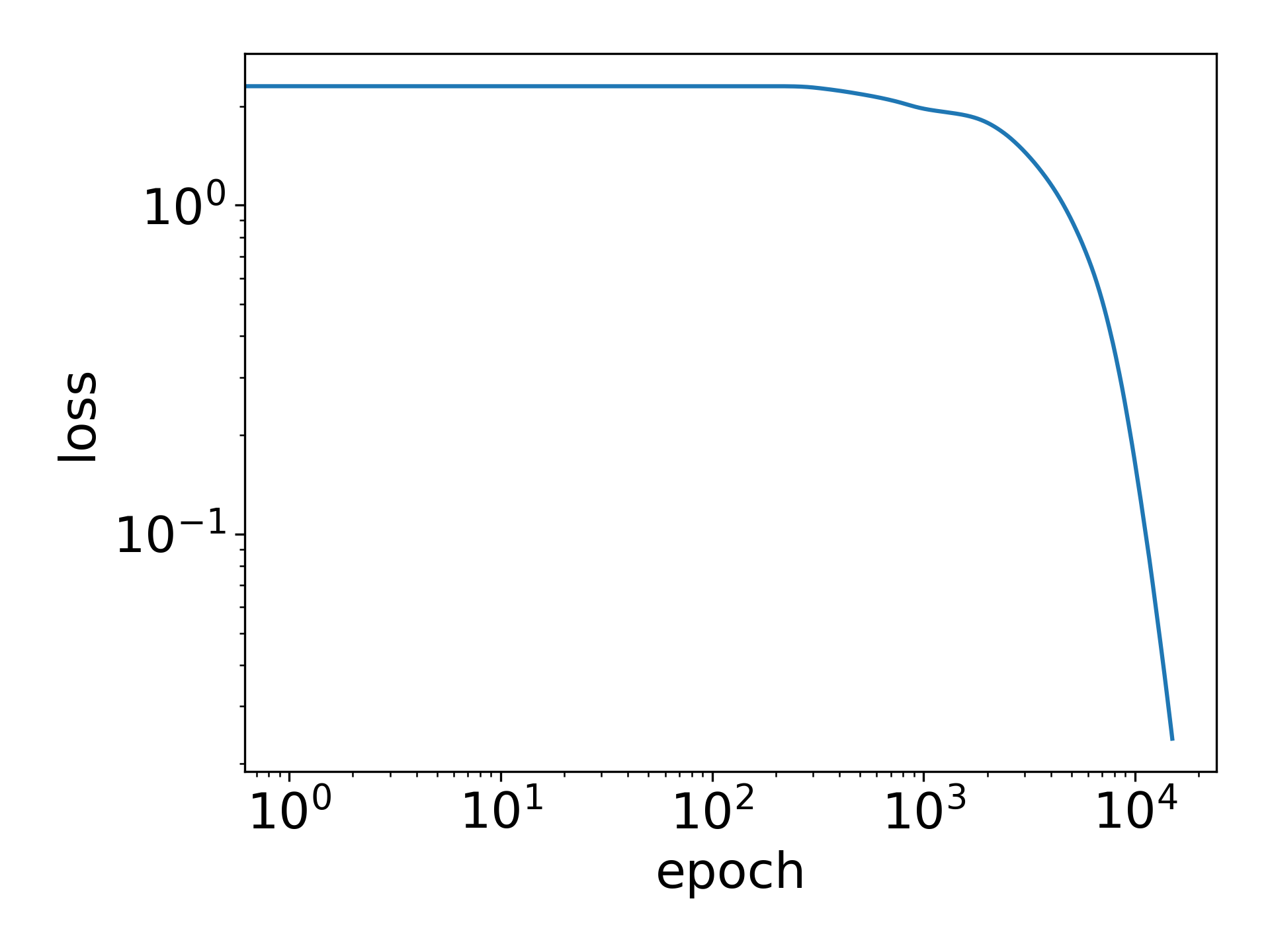}
         \caption{Fig.\ref{fig:tanhathreeccross}}
     \end{subfigure}
     
     \begin{subfigure}[b]{0.3\textwidth}
         \centering
         \includegraphics[width=\textwidth]{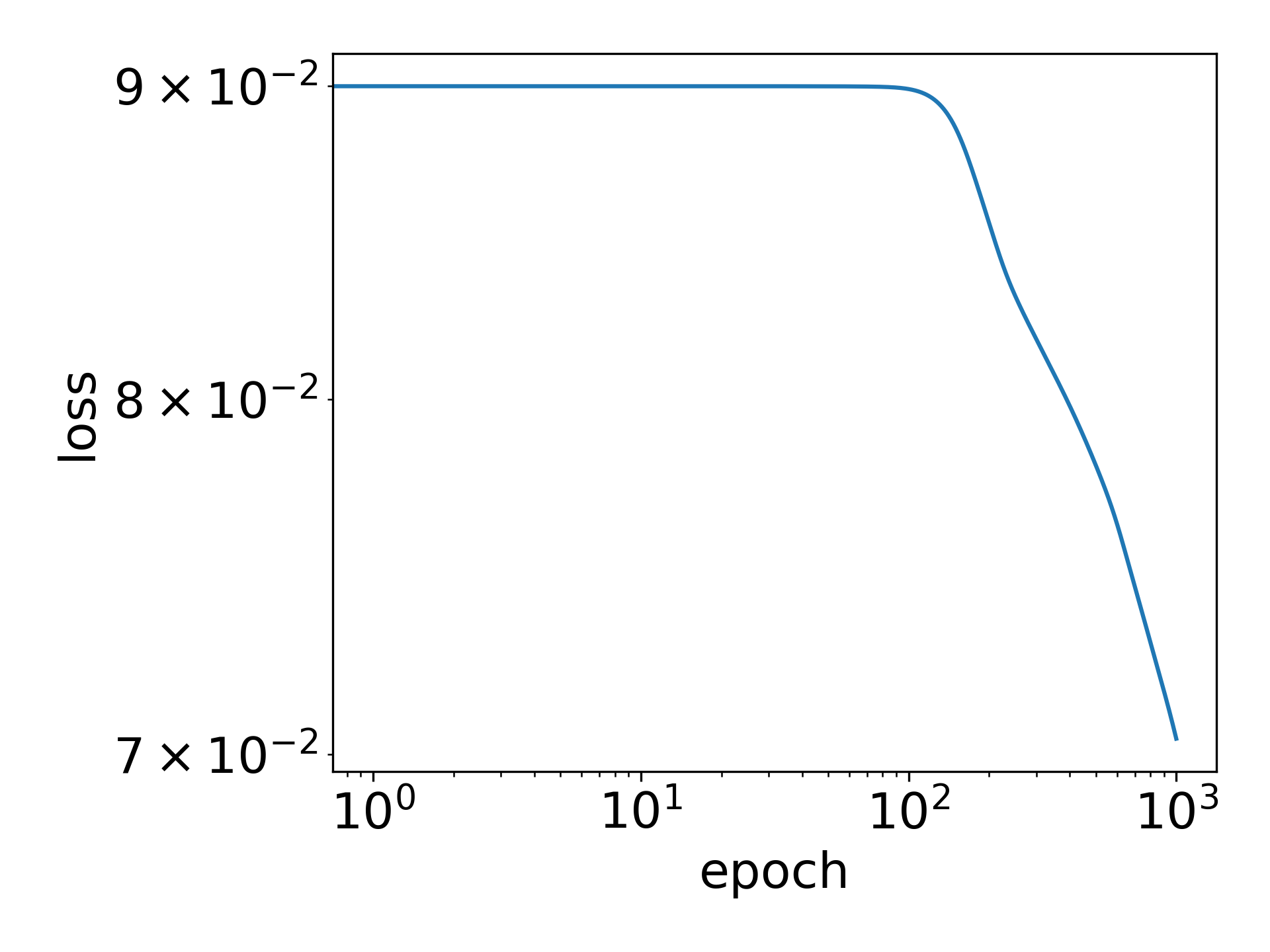}
         \caption{Fig.\ref{fig:tanhathreecMSESoft}}
     \end{subfigure}
     \hfill
     \begin{subfigure}[b]{0.3\textwidth}
         \centering
         \includegraphics[width=\textwidth]{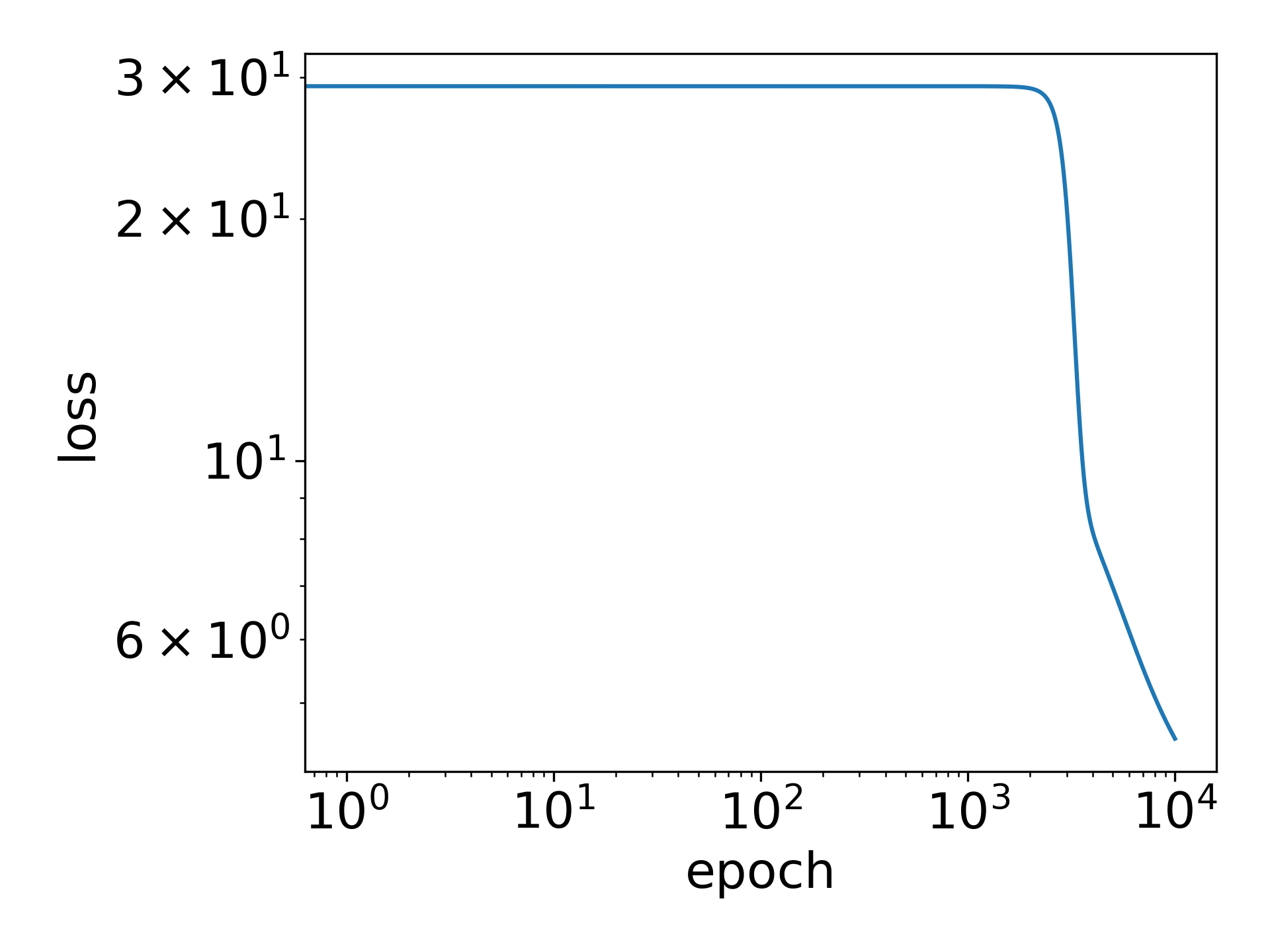}
        \caption{Fig.\ref{GDMNISTc}}
     \end{subfigure}
     \hfill
    \begin{subfigure}[b]{0.3\textwidth}
        \centering
        \includegraphics[width=\textwidth]{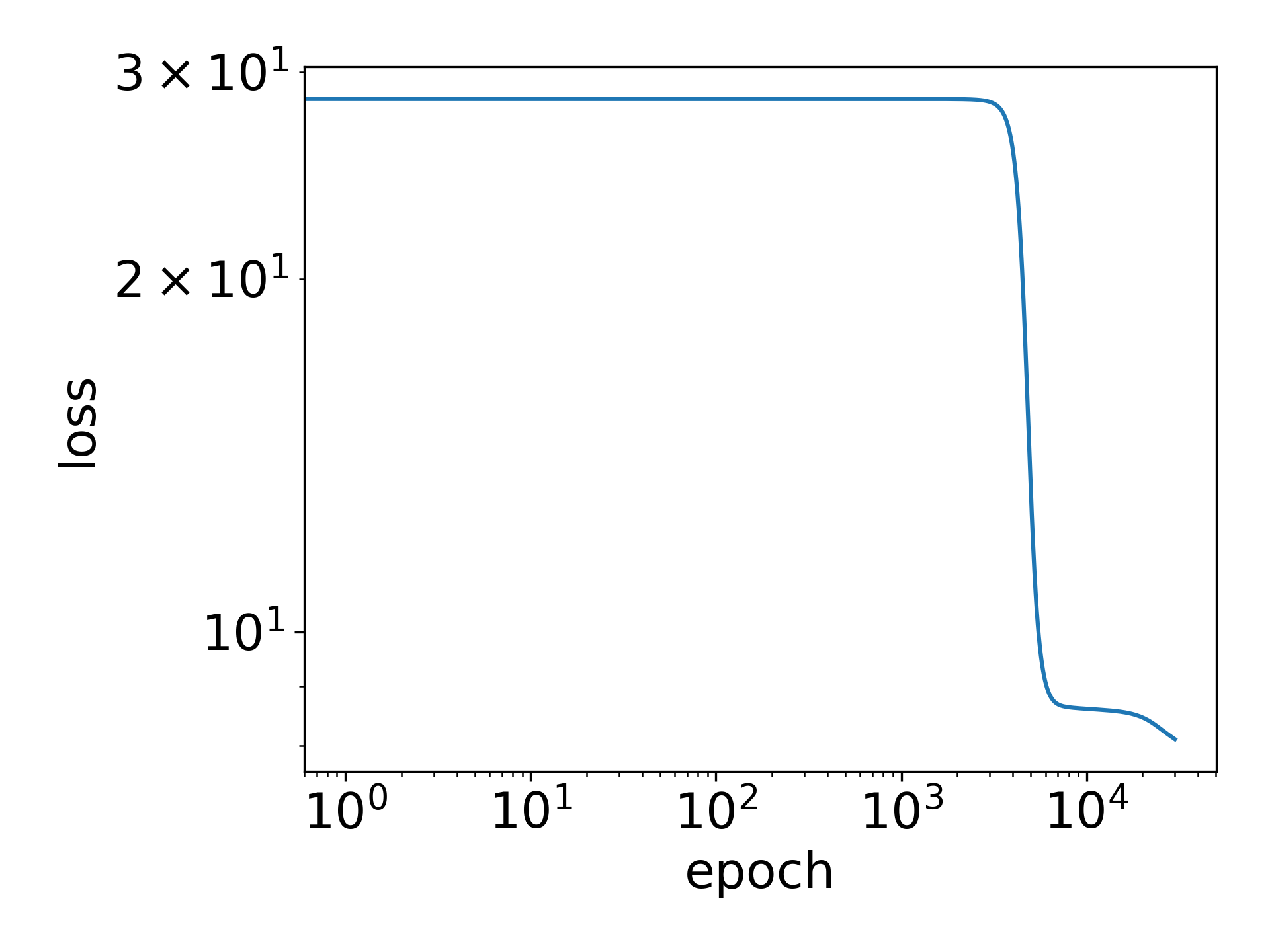}
        \caption{Fig.\ref{GDCifar10c}}
    \end{subfigure}
        \caption{Losses of the experiments on MNIST and CIFAR10 dataset. The original figures corresponding to each sub-picture are written in the sub-captions. }
        \label{Fig.loss}
\end{figure}

\section{Preliminaries}
\subsection{Some Notations}\label{subsection...Notations}
For a matrix $\mathbf{A}$, we use $\mathbf{A}_{i, j}$ to denote its $(i, j)$-th entry. We  also use $\mathbf{A}_{i,:}$ to denote the $i$-th row vector of $\mathbf{A}$ and define $\mathbf{A}_{i, j: k}:=\left(\mathbf{A}_{i, j}, \mathbf{A}_{i, j+1}, \cdots, \mathbf{A}_{i, k}\right)$ as part of the vector. Similarly $\mathbf{A}_{:, k}$ is the $k$-th column vector and $\mathbf{A}_{i: j, k}:=\left(\mathbf{A}_{i, k}, \mathbf{A}_{i+1, k}, \cdots, \mathbf{A}_{j, k}\right)^\T$ is   part of  the $k$-th column vector. 

We let $[n]=\{1,2, \ldots, n\}$. We set   $\fN(\vmu, \Sigma)$ as the normal distribution with mean $\vmu$ and covariance $\Sigma$.   For a vector $\mathbf{v}$, we use $\|\mathbf{v}\|_2$ to denote its Euclidean norm, and we use $\left<\cdot,\cdot\right>$ to denote the standard inner product between two vectors. For a matrix $\mathbf{A}$, we use $\|\mathbf{A}\|_{\mathrm{F}}$ to denote its Frobenius norm and $\|\mathbf{A}\|_{2\to 2}$ to denote its operator norm.  Finally, we use $\fO(\cdot)$ and $\Omega(\cdot)$ for the standard Big-O and Big-Omega notations.
\subsection{Problem Setup}
We focus on the empirical risk minimization problem given by the quadratic loss:
\begin{equation}\label{eq...text...Intro...LossFunction}
\min_{\vtheta}R_S(\vtheta)=\frac{1}{2n}\sum_{i=1}^n\left({f(\vx_i,\vtheta)-y_i}\right)^2.
\end{equation}
In the above, $n$ is the total number of training samples,  $\{ \vx_i\}_{i=1}^n$ are the training inputs, $\{ y_i\}_{i=1}^n$ are the labels,  $f(\vx_i,\vtheta)$ is the prediction function, and $\vtheta$ are the parameters to be optimized, and their dependence is modeled by a $(L+1)$-layer convolution neural network~(CNN) with filter size $m\times m$. We denote $\vx^{[l]}(i)$ as the output of the $l$-th layer with respect to the $i$-th sample for $l\geq 1$, and $\vx^{[0]}(i):=\vx_i$ is the $i$-th input. For any $l\in[0:L]$, we denote the size of width, height, channel of $\vx^{[l]}$ as $W_l,~H_l,$ and $C_l$ respectively, i.e., $\{\vx^{[l]}(i)\}_{i=1}^n\subset\sR^{W_l \times H_l\times C_l}$. 
We introduce a {\emph{filter}} operator $\chi(\cdot,\cdot)$  , which maps the width and height indices of the output of all layers to a binary variable,    i.e.,  for a filter of size $m\times m$, the filter operator reads 
\begin{equation}\label{eq...text...Intro...FilterOperator}
    \chi(p, q)=\left\{\begin{array}{l}1,\ \ \text{for~~} 0\leqslant p, q \leqslant m-1 
    \\ 0, \ \ \text{otherwise,} \end{array}\right. 
\end{equation}
 and  the   $(L+1)$-layer CNN with filter size $m\times m$ is recursively defined for  $l\in[2:L]$,
\begin{align*}
\vx_{u, v, \beta}^{[1]}&:=\left[\sum_{\alpha=1}^{C_{0}}\left(\sum_{p=-\infty}^{+\infty} \sum_{q=-\infty}^{+\infty} \vx^{[0]}_{u+p, v+q, \alpha} \cdot \mW_{p,q,\alpha,\beta}^{[1]} \cdot \chi(p, q)\right)\right]+\vb_{\beta}^{[1]}, \\
\vx_{u, v, \beta}^{[l]}&:=\left[\sum_{\alpha=1}^{C_{l-1}}\left(\sum_{p=-\infty}^{+\infty} \sum_{q=-\infty}^{\infty} \sigma\left(\vx_{u+p, v+q, \alpha}^{[l-1]} \right) \cdot \mW_{p,q,\alpha,\beta}^{[l]} \cdot \chi(p, q) \right)\right]+\vb_{\beta}^{[l]}, \\
f(\vx,\vtheta)& :=f_{\mathrm{CNN}}(\vx,\vtheta):=\left<\va, \sigma\left(\vx^{[L]}\right)\right>=\sum_{\beta=1}^{C_L}\sum_{u=1}^{W_L}\sum_{v=1}^{H_L} \va_{u, v, \beta} \cdot\sigma\left( \vx_{u, v, \beta}^{[L]}\right),
\end{align*} 
 where $\sigma(\cdot)$ is the activation function applied coordinate-wisely to its input, and for each layer $l\in[L]$, all parameters belonging to this layer are initialized by: For $p,q\in[m-1]$,  $\alpha\in[C_{l-1}]$ and $\beta\in[C_l]$,
\begin{equation}
\mW_{p,q,\alpha,\beta}^{[l]}\sim\fN(0, \beta_1^2),\quad  \vb_{\beta}^{[l]}  \sim\fN(0, \beta_1^2).
\end{equation}
Moreover, for $u\in[W_L]$ and $v\in[H_L]$,
\begin{equation}
\quad \va_{u, v, \beta}\sim\fN(0, \beta_2^2),
\end{equation}
 and for convenience we set  
$\beta_1=\beta_2=\eps,$ 
where $\eps>0$ is the scaling parameter. 
Finally, for all $i\in[n]$,  we denote hereafter that  
\[
e_i :=e_i(\vtheta) :=  f (\vx_i,\vtheta) - y_i,   
\]
and  
\[
\ve:=\ve(\vtheta) := {[e_1(\vtheta), e_2(\vtheta), \ldots, e_n(\vtheta)]}^{\T}\in\sR^n.
\]
\subsection{GD Dynamics}
In this paper, we train all layers of the neural network with continuous time gradient descent~(GD):  For any time $t \geq 0$,
\begin{equation}\label{eq...text...Intro...GDAbstract}
\frac{\D {\boldsymbol{\theta}}}{\D t}=-\nabla_{\boldsymbol{\theta}} R_S(\boldsymbol{\theta}).
\end{equation}
We remark that details of the dynamics \eqref{eq...text...Intro...GDAbstract} are hard to write out in matrix form, so  we turn to the alternative approach to derive the GD dynamics of each individual parameter. In order for that, it is natural for us to define a series of auxiliary variables: For each  $i\in[n]$ and $l\in[L]$, we define $\vz_{u,v,\beta}^{[l]}(i)$ as the partial derivative of $f(\vx_i, \vtheta)$ with respect to $\vx_{u,v,\beta}^{[l]}(i)$, i.e.,
\begin{equation}\label{eq...text...GDDynamics...AuxVariables}
    \vz_{u,v,\beta}^{[l]}(i):= \frac{\partial f(\vx_i,\vtheta)}{\partial \vx_{u,v,\beta}^{[l]}(i)},
\end{equation}
we obtain that for $l\in[L-1]$,
\begin{equation*}
\begin{aligned}
\vz_{u,v,\beta}^{[l]}(i)&= \sum_{\alpha=1}^{C_{l+1}} \sum_{s=1}^{W_{l+1}} \sum_{t=1}^{H_{l+1}}  \vz_{s, t, \alpha}^{[l+1]}(i) \cdot  \mW_{u-s,v-t,\beta,\alpha}^{[l+1]} \cdot \sigma^{(1)}\left(\vx_{u, v, \beta}^{[l]}(i)\right) \cdot \chi{(u-s, v-t)}, \\
\vz_{u,v,\beta}^{[L]}(i)&= \va_{u,v,\beta}\cdot\sigma^{(1)}\left(\vx_{u, v, \beta}^{[L]}(i)\right),
\end{aligned} 
\end{equation*}
hence we obtain that for $l\in[2:L],$
\begin{equation}\label{eqgroup...text...GDDynamics...W_partial}
\begin{aligned}
\frac{\partial f(\vx_i, \vtheta)}{\partial \mW_{p,q,\alpha, \beta}^{[1]}}&=\sum_{u=1}^{W_{1}} \sum_{v=1}^{H_{1}} \vz_{u,v,\beta}^{[1]}(i) \cdot \vx^{[0]}_{u+p,v+q,\alpha}(i),\\
\frac{\partial f(\vx_i, \vtheta)}{\partial \mW_{p,q,\alpha, \beta}^{[l]}}&=\sum_{u=1}^{W_{l}} \sum_{v=1}^{H_{l}} \vz_{u,v,\beta}^{[l]}(i) \cdot \sigma\left(\vx_{u+p,v+q,\alpha}^{[l-1]} (i)\right),\\
\frac{\partial f(\vx_i, \vtheta)}{\partial \vb_{\beta}^{[1]}}&=\sum_{u=1}^{W_{1}} \sum_{v=1}^{H_{1}} \vz_{u,v,\beta}^{[1]}(i),\\
\frac{\partial f(\vx_i, \vtheta)}{\partial \vb_{\beta}^{[l]}}&=\sum_{u=1}^{W_{l}} \sum_{v=1}^{H_{l}} \vz_{u,v,\beta}^{[l]}(i).
\end{aligned}
\end{equation}
With the above notations, for $l\in[2:L],$ the  dynamics \eqref{eq...text...Intro...GDAbstract} reads
\begin{equation}\label{eqgroup...text...GDDynamics...FullGDDyanmics}
\begin{aligned}
\frac{\D {\mW}_{p,q,\alpha,\beta}^{[1]}}{\D t}&=-\frac{1}{n}\sum_{i=1}^n e_i\cdot \left(\sum_{u=1}^{W_{1}} \sum_{v=1}^{H_{1}} \vz_{u,v,\beta}^{[1]}(i) \cdot \vx^{[0]}_{u+p,v+q,\alpha}(i)\right),\\
 \frac{\D {\mW}_{p,q,\alpha,\beta}^{[l]}}{\D t}&=-\frac{1}{n}\sum_{i=1}^n e_i\cdot \left(\sum_{u=1}^{W_{l}} \sum_{v=1}^{H_{l}} \vz_{u,v,\beta}^{[l]}(i) \cdot \sigma\left(\vx_{u+p,v+q,\alpha}^{[l-1]} (i)\right)\right),\\
\frac{\D \vb_{\beta}^{[1]}}{\D t}&=-\frac{1}{n}\sum_{i=1}^n e_i\cdot \left(\sum_{u=1}^{W_{1}} \sum_{v=1}^{H_{1}} \vz_{u,v,\beta}^{[1]}(i)\right),\\
\frac{\D \vb_{\beta}^{[l]}}{\D t}&=-\frac{1}{n}\sum_{i=1}^n e_i\cdot \left(\sum_{u=1}^{W_{l}} \sum_{v=1}^{H_{l}} \vz_{u,v,\beta}^{[l]}(i)\right),\\
 \frac{\D \va_{u,v,\beta}}{\D t}&=-\frac{1}{n}\sum_{i=1}^n e_i\cdot  \sigma\left(
 \vx_{u,v,\beta}^{[L]}(i)\right).
\end{aligned}
\end{equation}
\section{Two-Layer CNNs with Single Channel}
\subsection{Activation Function}
In this part, we shall impose some   technical conditions  on the activation function and input samples.
We start with a   technical condition~\cite[Definition 1]{zhou2022towards}  on the activation function $\sigma(\cdot)$ 
\begin{definition}[Multiplicity $r$]\label{def...Multiplicity} 
   $\sigma(\cdot):\sR\to\sR$ has multiplicity $r$
    if there exists an integer $r\geq 1$, such that  for all $0\leq s\leq r-1$, the $s$-th order derivative satisfies $\sigma^{(s)}(0)=0$, and $\sigma^{(r)}(0) \neq 0$.
\end{definition}
\noindent We   list out some examples of activation functions with different multiplicity.
\begin{remark}
~\\
\begin{itemize}
\item $\rm{tanh}(x):=\frac{\exp(x)-\exp(-x)}{\exp(x)+\exp(-x)}$ is with multiplicity $r=1$;\\
\item $\rm{SiLU}(x):=\frac{x}{1+ \exp(-x)}$ is with multiplicity $r=1$;\\
\item $\rm{xtanh}(x):=\frac{x\exp(x)-x\exp(-x)}{\exp(x)+\exp(-x)}$ is with multiplicity $r=2$.
\end{itemize}
\end{remark}
\begin{assumption}[Multiplicity $1$]\label{Assumption....ActivationFunctions}
The activation function $\sigma\in\fC^2(\sR)$, and there exists a universal constant $C_L>0$, such that   its first and second  derivatives   satisfy 
\begin{equation} 
   \Norm{\sigma^{(1)}(\cdot)}_{\infty}\leq C_L,\quad \Norm{\sigma^{(2)}(\cdot)}_{\infty}\leq C_L.
\end{equation}
Moreover,
\begin{equation} 
\sigma(0)=0,\quad \sigma^{(1)}(0)=1.
\end{equation}
\end{assumption}
\begin{remark}
We remark that $\sigma$ has multiplicity $1$.
$\sigma^{(1)}(0)=1$   can be replaced by  $\sigma^{(1)}(0)\neq 0$, and we set   $\sigma^{(1)}(0)=1$  for simplicity, and it can be easily satisfied by replacing the original activation $\sigma(\cdot)$ with $\frac{\sigma(\cdot)}{\sigma^{(1)}(0)}$.

We note that \Cref{Assumption....ActivationFunctions} can be satisfied by using the $\rm{tanh}$ activation:
\[
 \sigma(x)=\frac{\exp(x)-\exp(-x)}{\exp(x)+\exp(-x)},
\]
and the scaled SiLU activation
\[
 \sigma(x)=\frac{2x}{1+ \exp(-x)}.
\]
\end{remark}
\noindent
In the case of two-layer CNNs, as $L=2$,  then we shall set  ${\mW}_{p,q,\alpha,\beta}:= {\mW}_{p,q,\alpha,\beta}^{[1]}$,  $\vb_{\beta}:=\vb_{\beta}^{[1]}$, and $\vx_{u+p,v+q,\alpha}(i):=\vx^{[0]}_{u+p,v+q,\alpha}(i)$ for simplicity, then
 \eqref{eqgroup...text...GDDynamics...FullGDDyanmics} reads
\begin{align*}
\frac{\D {\mW}_{p,q,\alpha,\beta}}{\D t}&=-\frac{1}{n}\sum_{i=1}^n e_i\cdot \left(\sum_{u=1}^{W_{1}} \sum_{v=1}^{H_{1}}\va_{u,v,\beta}\cdot\sigma^{(1)}\left(\vx_{u, v, \beta}^{[1]}(i)\right)\cdot \vx_{u+p,v+q,\alpha}(i)\right),\\
\frac{\D \vb_{\beta}}{\D t}&=-\frac{1}{n}\sum_{i=1}^n e_i\cdot \left(\sum_{u=1}^{W_{1}} \sum_{v=1}^{H_{1}}\va_{u,v,\beta}\cdot\sigma^{(1)}\left(\vx_{u, v, \beta}^{[1]}(i)\right)\right),\\
 \frac{\D \va_{u,v,\beta}}{\D t}&=-\frac{1}{n}\sum_{i=1}^n e_i\cdot  \sigma\left(
 \vx_{u,v,\beta}^{[1]}(i)\right).
\end{align*}     
Moreover, since  the MNIST~\cite{deng2012mnist} images are black and white,  therefore we  do not need three different color-channels to represent the final color, and only one channel is enough, i.e., $C_0=1$, hence the above dynamics can be further simplified into
\begin{equation}\label{eq...text...2Layer...alpha=1Dynamics}
\begin{aligned}
\frac{\D {\mW}_{p,q,\beta}}{\D t}&=-\frac{1}{n}\sum_{i=1}^n e_i\cdot \left(\sum_{u=1}^{W_{1}} \sum_{v=1}^{H_{1}}\va_{u,v,\beta}\cdot\sigma^{(1)}\left(\vx_{u, v, \beta}^{[1]}(i)\right)\cdot \vx_{u+p,v+q}(i)\right),\\
\frac{\D \vb_{\beta}}{\D t}&=-\frac{1}{n}\sum_{i=1}^n e_i\cdot \left(\sum_{u=1}^{W_{1}} \sum_{v=1}^{H_{1}}\va_{u,v,\beta}\cdot\sigma^{(1)}\left(\vx_{u, v, \beta}^{[1]}(i)\right)\right),\\
 \frac{\D \va_{u,v,\beta}}{\D t}&=-\frac{1}{n}\sum_{i=1}^n e_i\cdot  \sigma\left(
 \vx_{u,v,\beta}^{[1]}(i)\right).
\end{aligned}     
\end{equation}
We identify the vectorized parameters $\vtheta $ as variables of order $1$  by setting 
$ 
 \vtheta=\eps\Bar{\vtheta},
$  
and 
\begin{equation}
\begin{aligned}
\vx_{u, v, \beta}^{[1]}&= \left(\sum_{p=-\infty}^{+\infty} \sum_{q=-\infty}^{+\infty} \vx_{u+p, v+q} \cdot \eps\overline{\mW}_{p,q,\beta} \cdot \chi(p, q)\right) +\eps\Bar{\vb}_{\beta}^{[1]}\\
&= \eps \Bar{\vx}_{u, v, \beta}^{[1]},
\end{aligned}
\end{equation}
where 
\[
\Bar{\vx}_{u, v, \beta}^{[1]}:= \left(\sum_{p=-\infty}^{+\infty} \sum_{q=-\infty}^{+\infty} \vx_{u+p, v+q} \cdot  \overline{\mW}_{p,q,\beta} \cdot \chi(p, q)\right) +\Bar{\vb}_{\beta}^{[1]},
\]
is also of order $1$,
and the rescaled dynamics can be written into
\begin{align*}
\frac{\D \overline{\mW}_{p,q,\beta}}{\D t}&=-\frac{1}{n}\sum_{i=1}^n e_i\cdot \left(\sum_{u=1}^{W_{1}} \sum_{v=1}^{H_{1}}\Bar{\va}_{u,v,\beta}\cdot\sigma^{(1)}\left(\eps \Bar{\vx}_{u, v, \beta}^{[1]}(i)\right)\cdot \vx_{u+p,v+q}(i)\right),\\
\frac{\D \Bar{\vb}_{\beta}}{\D t}&=-\frac{1}{n}\sum_{i=1}^n e_i\cdot \left(\sum_{u=1}^{W_{1}} \sum_{v=1}^{H_{1}}\Bar{\va}_{u,v,\beta}\cdot\sigma^{(1)}\left(\eps \Bar{\vx}_{u, v, \beta}^{[1]}(i)\right)\right),\\
 \frac{\D \Bar{\va}_{u,v,\beta}}{\D t}&=-\frac{1}{n}\sum_{i=1}^n e_i\cdot  \frac{\sigma\left(\eps \Bar{\vx}_{u, v, \beta}^{[1]}(i)\right)}{\eps},
\end{align*}  
with the following initialization 
\begin{equation}
\overline{\mW}^0_{p,q,\beta} \sim\fN(0, 1),\quad  \Bar{\vb}^0_{\beta}   \sim\fN(0, 1),\quad \Bar{\va}^0_{u, v, \beta}\sim\fN(0,1).
\end{equation}
In the following discussion throughout this paper, we always refer to the above rescaled dynamics and drop all the ``bar''s of $\Bar{\vtheta}$, $\overline{\mW}_{p,q,\beta}$, $\Bar{\vb}_{\beta}$, $\Bar{\vx}_{u, v, \beta}^{[1]}$    and  $\Bar{\va}_{u, v, \beta}$  for  notational simplicity. Moreover, we remark that $p\in[0:m-1]$, $q\in[0:m-1]$ and $\beta\in[M]$, where $m$ is the filter size, and $M:=C_1$, the number of channels in $\vx^{[1]}(i)$,  which can be heuristically understood as the `width' of the hidden layer in the case of two-layer neural networks~(NNs). Before we end this section, we assume hereafter that 
\begin{assumption} \label{assumption...Data}
The training inputs  $\{ \vx_i\}_{i=1}^n$  and labels  $\{ y_i\}_{i=1}^n$ satisfy  that   there exists a universal constant $c>0$, such that given any $i\in[n]$, then for each $u\in[W_0]$,  $v\in[H_0]$ and $\alpha\in [C_0]$, the following holds
\[  \frac{1}{c}\leq \Abs{\vx_{u,v,\alpha}(i)}, \quad\Abs{y_{i}}\leq c.\] 
\end{assumption}
\noindent
We assume further that 
\begin{assumption}\label{assump...LimitExistence}
The following limit exists
\begin{equation} \label{eq...assump...definition...LimitExistence}
{\gamma}:=\lim_{M\to\infty} -\frac{\log \eps^2}{\log M}.
\end{equation}

\end{assumption}
\subsection{Effective Linear Dynamics}
As   the normalized flow reads
\begin{align*}
\frac{\D {\mW}_{p,q,\beta}}{\D t}&=-\frac{1}{n}\sum_{i=1}^n e_i\cdot \left(\sum_{u=1}^{W_{1}} \sum_{v=1}^{H_{1}}{\va}_{u,v,\beta}\cdot\sigma^{(1)}\left(\eps {\vx}_{u, v, \beta}^{[1]}(i)\right)\cdot \vx_{u+p,v+q}(i)\right),\\
\frac{\D  {\vb}_{\beta}}{\D t}&=-\frac{1}{n}\sum_{i=1}^n e_i\cdot \left(\sum_{u=1}^{W_{1}} \sum_{v=1}^{H_{1}} {\va}_{u,v,\beta}\cdot\sigma^{(1)}\left(\eps  {\vx}_{u, v, \beta}^{[1]}(i)\right)\right),\\
 \frac{\D  {\va}_{u,v,\beta}}{\D t}&=-\frac{1}{n}\sum_{i=1}^n e_i\cdot  \frac{\sigma\left(\eps  {\vx}_{u, v, \beta}^{[1]}(i)\right)}{\eps},
\end{align*}  
since $e_i\approx -y_i$, and   by means of perturbation
expansion with respect to $\eps$ and keep the order $1$ term, we obtain that
\begin{equation}\label{eq...text...E-LinearDynamics....LinearDynamics}
\begin{aligned}
\frac{\D {\mW}_{p,q,\beta}}{\D t}&\approx\frac{1}{n}\sum_{i=1}^n y_i\cdot \left(\sum_{u=1}^{W_{1}} \sum_{v=1}^{H_{1}}{\va}_{u,v,\beta}\cdot \vx_{u+p,v+q}(i)\right),\\
\frac{\D  {\vb}_{\beta}}{\D t}&\approx \frac{1}{n}\sum_{i=1}^n y_i\cdot  \sum_{u=1}^{W_{1}} \sum_{v=1}^{H_{1}} {\va}_{u,v,\beta},\\
 \frac{\D  {\va}_{u,v,\beta}}{\D t}&\approx\frac{1}{n}\sum_{i=1}^n y_i\cdot   {\vx}_{u, v, \beta}^{[1]}(i)\\
 &=\frac{1}{n}\sum_{i=1}^n y_i\cdot   \left[\left(\sum_{p=0}^{m-1} \sum_{q=0}^{m-1} \vx_{u+p, v+q}(i) \cdot   {\mW}_{p,q,\beta}   \right) + {\vb}_{\beta}\right].
\end{aligned}  
\end{equation}
Given any $u\in[W_1]$ and $v\in[H_1]$, then for all $p, q\in[0:m-1]$, we set
\begin{equation}\label{eq...text...MultiChannels...SumofVectors}
\begin{aligned}
\vz_{u+p,v+q}&:=  \frac{1}{n}\sum_{i=1}^n y_i \vx_{u+p,v+q}(i),\\
z&:=\frac{1}{n}\sum_{i=1}^n y_i,
\end{aligned}
\end{equation}
then the dynamics \eqref{eq...text...E-LinearDynamics....LinearDynamics} can be further simplified into: For any $\beta\in[M]$,
\begin{equation}\label{eq...text...E-LinearDynamics....SimplifiedLinearDynamics}
\begin{aligned}
\frac{\D {\mW}_{p,q,\beta}}{\D t}&\approx \sum_{u=1}^{W_{1}} \sum_{v=1}^{H_{1}}{\va}_{u,v,\beta}\cdot \vz_{u+p,v+q},\\
\frac{\D  {\vb}_{\beta}}{\D t}&\approx   \sum_{u=1}^{W_{1}} \sum_{v=1}^{H_{1}} {\va}_{u,v,\beta}\cdot z,\\
 \frac{\D  {\va}_{u,v,\beta}}{\D t}&\approx  \left(\sum_{p=0}^{m-1} \sum_{q=0}^{m-1}  \vz_{u+p, v+q}  \cdot   {\mW}_{p,q,\beta}   \right) + {\vb}_{\beta}\cdot z.
\end{aligned}  
\end{equation}
We observe that \eqref{eq...text...E-LinearDynamics....SimplifiedLinearDynamics} reveals that the training  dynamics of two-layer CNNs at initial stage has a close relationship to power iteration  of a matrix $\mA$ that only depends on the input samples, i.e., the dynamics \eqref{eq...text...E-LinearDynamics....SimplifiedLinearDynamics} takes the form
\begin{equation}
   \frac{\D \vtheta_\beta}{\D t}=\mA\vtheta_\beta,  
\end{equation}
where 
\begin{align*}
\vtheta_\beta:=\Big(&\mW_{0,0,\beta},\mW_{0,1,\beta},\cdots,\mW_{0,m-1,\beta}; \mW_{1,0,\beta},\cdots,\mW_{1,m-1,\beta};\cdots\cdots\mW_{m-1,m-1,\beta};{\vb}_{\beta};\\
&{\va}_{1,1,\beta}, {\va}_{1,2,\beta},\cdots,{\va}_{1,H_1,\beta};{\va}_{2,1,\beta},\cdots,{\va}_{2,H_1,\beta};\cdots\cdots{\va}_{W_1,H_1,\beta}\Big)^\T,
\end{align*} 
and if we would like to simplify our notations
\begin{align*}
\vtheta_\beta=\Big(&\mW_{0,0:(m-1),\beta}; \mW_{1,0:(m-1),\beta};\cdots\cdots\mW_{m-1,0:(m-1),\beta};{\vb}_{\beta};\\
&{\va}_{1,1:H_1,\beta} ;{\va}_{2,1:H_1,\beta};\cdots\cdots{\va}_{W_1,1:H_1,\beta}\Big)^\T,
\end{align*} 

\begin{equation}
    \mA:=\left[\begin{array}{cc}
\mzero_{(m^2+1)\times (m^2+1)} & {\mZ}^\T \\
\mZ & \mzero_{W_1H_1\times W_1H_1} 
\end{array}\right],
\end{equation}
where $\mZ\in\sR^{W_1H_1\times (m^2+1)}$ and $\mZ$ depends sorely on the input samples  $\{ \vx_i\}_{i=1}^n$ and
$\{ y_i\}_{i=1}^n$, whose entries read
\begin{align}
 \mZ:=\left[\begin{array}{ccccccccccc }
\vz_{1,1}&\vz_{1,2}&\cdots &\vz_{1,m};&\vz_{2,1} &\cdots &\vz_{2,m};&\cdots\cdots&\vz_{m,m};&z\\
\vz_{1,2}&\vz_{1,3}&\cdots &\vz_{1,m+1};&\vz_{2,2} &\cdots &\vz_{2,m+1};&\cdots\cdots&\vz_{m,m+1};&z\\
\vdots&\vdots&\cdots &\vdots &\vdots&\cdots &\vdots&\cdots\cdots&\vdots&\vdots\\
\vz_{1,H_1}&\vz_{1,H_1+1}&\cdots &\vz_{1,H_0};&\vz_{2,H_1} &\cdots &\vz_{2,H_0};&\cdots\cdots&\vz_{m,H_0};&z\\
\vz_{2,1}&\vz_{2,2}&\cdots &\vz_{2,m};&\vz_{3,1} &\cdots &\vz_{3,m};&\cdots\cdots&\vz_{m+1,m};&z\\
\vdots&\vdots&\cdots &\vdots &\vdots&\cdots &\vdots&\cdots\cdots&\vdots&\vdots\\
\vz_{2,H_1}&\vz_{2,H_1+1}&\cdots &\vz_{2,H_0};&\vz_{3,H_1} &\cdots &\vz_{3,H_0};&\cdots\cdots&\vz_{m+1,H_0};&z\\
\vdots&\vdots&\cdots &\vdots &\vdots&\cdots &\vdots&\cdots\cdots&\vdots&\vdots\\
\vz_{W_1,H_1}&\vz_{W_1,H_1+1}&\cdots &\vz_{W_1,H_0};&\vz_{W_1+1,H_1} &\cdots &\vz_{W_1+1,H_0};&\cdots\cdots&\vz_{W_0,H_0};&z\\
\end{array}\right].
\label{Singlechannel_matrix}
\end{align}
If we would like to simplify our notations,
\begin{align*}
 \mZ:=\left[\begin{array}{cccccc}
\vz_{1,1:m};&\vz_{2,1:m};&\cdots\cdots&\vz_{m,1:m};&z\\
\vz_{1,2:(m+1)};&\vz_{2,2:(m+1)};&\cdots\cdots&\vz_{m,2:(m+1)};&z\\
\vdots&\vdots&\cdots\cdots &\vdots &\vdots& \\
\vz_{1,H_1:H_0};&\vz_{2,H_1:H_0};&\cdots\cdots&\vz_{m,H_1:H_0};&z\\
\vz_{2,1:m};&\vz_{3,1:m};&\cdots\cdots&\vz_{m+1,1:m};&z\\
\vdots&\vdots&\cdots  \cdots&\vdots&\vdots\\
\vz_{2,H_1:H_0};&\vz_{3,H_1:H_0};&\cdots\cdots&\vz_{m+1,H_1:H_0};&z\\
\vdots&\vdots& \cdots\cdots&\vdots&\vdots\\
\vz_{W_1,H_1:H_0};&\vz_{W_1+1,H_1:H_0};&\cdots\cdots&\vz_{W_0,H_1:H_0};&z\\
\end{array}\right].
\end{align*}

In order  to solve out the simplified dynamics \eqref{eq...text...E-LinearDynamics....SimplifiedLinearDynamics},  we need to perform Singular value decomposition~(SVD) on $\mZ$, i.e., 
\begin{equation}\label{eq...text...E-LinearDynamics....SVD}
\mZ=\mU\Lambda\mV^\T,    
\end{equation}
where
\[
\mU=\left[\vu_1,\vu_2,\cdots,\vu_{W_1H_1}\right],\quad\mV=\left[\bm{v}_1,\bm{v}_2,\cdots,\bm{v}_{m^2+1}\right],
\]
and as we denote $r:=\mathrm{rank}(\mZ)$, naturally, $r\leq \min\{W_1H_1, m^2+1\}$
 we have $r$ singular values,
\[\lambda_1\geq\lambda_2\geq \cdots \geq\lambda_r> 0,\]
 and WLOG, we assume that  
 \begin{assumption}[Spectral Gap of $\mZ$]\label{assump...SpectralGap}
The singular values $\{\lambda_k\}_{k=1}^r$ of $\mZ$ satisfy that
\begin{equation} \lambda_1>\lambda_2\geq \cdots \geq\lambda_r> 0,\end{equation}     
and we denote the spectral gap between $\lambda_1$ and $\lambda_2$ by 
\[\Delta\lambda:= \lambda_1-\lambda_2.\]
 \end{assumption}
Moreover, as we denote further that
\begin{align*}
\vtheta_{\mW,\beta}&:=\left(\mW_{0,0,\beta},\mW_{0,1,\beta},\cdots,\mW_{0,m-1,\beta}; \mW_{1,0,\beta},\cdots,\mW_{1,m-1,\beta};\cdots\cdots\mW_{m-1,m-1,\beta};{\vb}_{\beta}  \right)^\T,\\
\vtheta_{\va,\beta}&:= \left({\va}_{1,1,\beta}, {\va}_{1,2,\beta},\cdots,{\va}_{1,H_1,\beta};{\va}_{2,1,\beta},\cdots,{\va}_{2,H_1,\beta};\cdots\cdots{\va}_{W_1,H_1,\beta}\right)^\T,
\end{align*}
hence 
\[
\vtheta_\beta=\left(\vtheta_{\mW,\beta}^\T, \vtheta_{\va,\beta}^\T\right)^\T,
\]
and the linear dynamics \eqref{eq...text...E-LinearDynamics....SimplifiedLinearDynamics} read
\begin{equation}\label{eq...text...E-LinearDynamics....SimplifiedLinearDynamicsMatrixForm}
\begin{aligned}
\frac{\D \vtheta_{\mW,\beta}}{\D t}&=\mZ^\T\vtheta_{\va,\beta},\quad \vtheta_{\mW,\beta}(0)=\vtheta_{\mW,\beta}^0,  \\
\frac{\D \vtheta_{\va,\beta}}{\D t}&=\mZ\vtheta_{\mW,\beta}, \quad\vtheta_{\va,\beta}(0)=\vtheta_{\va,\beta}^0,
\end{aligned}
\end{equation}
hence \eqref{eq...text...E-LinearDynamics....SimplifiedLinearDynamicsMatrixForm} can be simplified further into two separate second order differential equations,  
\begin{equation}\label{eq...text...E-LinearDynamics....SecondOrder...ThetaW} 
\frac{\D^2 \vtheta_{\mW,\beta}}{\D t^2}=\mZ^\T\mZ \vtheta_{\mW,\beta},~~\vtheta_{\mW,\beta}(0)=\vtheta_{\mW,\beta}^0,~~\frac{\D \vtheta_{\mW,\beta}}{\D t}(0)=\mZ^\T\vtheta_{\va,\beta}^0,
\end{equation}
and 
\begin{equation}\label{eq...text...E-LinearDynamics....SecondOrder...ThetaA} 
\frac{\D^2 \vtheta_{\va,\beta}}{\D t^2}=\mZ \mZ^\T\vtheta_{\va,\beta},~~\vtheta_{\va,\beta}(0)=\vtheta_{\va,\beta}^0,~~\frac{\D \vtheta_{\va,\beta}}{\D t}(0)=\mZ\vtheta_{\mW,\beta}^0.
\end{equation}
We observe that 
\begin{align*}
\mZ^\T\mZ&=\sum_{k=1}^r \lambda_k^2\bm{v}_k \bm{v}_k^\T,\\
\mZ\mZ^\T&=\sum_{k=1}^r \lambda_k^2\vu_k \vu_k^\T,
\end{align*}
hence the solutions to \eqref{eq...text...E-LinearDynamics....SecondOrder...ThetaW} and \eqref{eq...text...E-LinearDynamics....SecondOrder...ThetaA} respectively reads
\begin{equation}
\begin{aligned}
\vtheta_{\mW,\beta}(t)&=\left(\sum_{k=1}^r\left[c_{\lambda_k,\mW,\beta}\exp(\lambda_kt)+d_{\lambda_k,\mW,\beta}\exp(-\lambda_kt)\right]\bm{v}_k\right)\\
&~~+\fP_{(r+1):(m^2+1)}\vtheta_{\mW,\beta}(0),\\
\vtheta_{\va,\beta}(t)&=\left(\sum_{k=1}^r\left[c_{\lambda_k,\va,\beta}\exp(\lambda_kt)+d_{\lambda_k,\va,\beta}\exp(-\lambda_kt)\right]\vu_k\right)\\
&~~+\fP_{(r+1):(W_1H_1)}\vtheta_{\va,\beta}(0),
\end{aligned}
\end{equation}
where for each $k\in[r]$ and $\beta\in[M]$, the constants $c_{\lambda_k,\mW,\beta}$, $d_{\lambda_k,\mW,\beta}$, $c_{\lambda_k,\va,\beta}$ and $d_{\lambda_k,\va,\beta}$ depend on   $\left<\vtheta_{\mW,\beta}^0, \bm{v}_k\right>$ and $\left<\vtheta_{\va,\beta}^0,\vu_k\right>$, i.e.,     the constants $c_{\lambda_k,\mW,\beta}$ and $d_{\lambda_k,\mW,\beta}$ are determined by,
\begin{equation}\label{eq...text...E-LinearDynamics...Constants}
\begin{aligned}
c_{\lambda_k,\mW,\beta}+d_{\lambda_k,\mW,\beta}&=\left<\vtheta_{\mW,\beta}^0, \bm{v}_k\right>,\\
\lambda_k c_{\lambda_k,\mW,\beta}-\lambda_kd_{\lambda_k,\mW,\beta}&=\left<\mZ^\T\vtheta_{\va,\beta}^0,\bm{v}_k\right>=\lambda_k\left<\vtheta_{\va,\beta}^0, \bm{u}_k\right>,
\end{aligned}
\end{equation}
thus 
\begin{align*}
c_{\lambda_k,\mW,\beta}&=\frac{1}{2}\left(\left<\vtheta_{\mW,\beta}^0, \bm{v}_k\right>+\left<\vtheta_{\va,\beta}^0, \bm{u}_k\right>\right),\\
d_{\lambda_k,\mW,\beta}&=\frac{1}{2}\left(\left<\vtheta_{\mW,\beta}^0, \bm{v}_k\right>-\left<\vtheta_{\va,\beta}^0, \bm{u}_k\right>\right),
\end{align*}
which matches the same constants as  the   ones in two-layer NNs~\cite{chen2023phase}, a special case where $r=1$.
Then with slight misuse of notations, $\fP_{1:r}\vtheta_{\mW,\beta}(0)$ refers to the projection of $\vtheta_{\mW,\beta}^0$ towards $\mathrm{span}\left\{\bm{v}_k\right\}_{k=1}^{r}$, $\fP_{(r+1):(m^2+1)}\vtheta_{\mW,\beta}(0)$ refers to the projection of $\vtheta_{\mW,\beta}^0$ towards $\mathrm{span}\left\{\bm{v}_k\right\}_{k=r+1}^{m^2+1}$, and  $\fP_{(r+1):(W_1H_1)}\vtheta_{\va,\beta}(0)$ refers to the projection of $\vtheta_{\va,\beta}^0$ towards $\mathrm{span}\left\{\vu_k\right\}_{k=r+1}^{W_1H_1}$. 
\begin{proposition}\label{prop...LinearODESolution}
The solution to the linear differential equation
\begin{equation}
\begin{aligned}
\frac{\D \vtheta_{\mW,\beta}}{\D t}&=\mZ^\T\vtheta_{\va,\beta},\quad \vtheta_{\mW,\beta}(0)=\vtheta_{\mW,\beta}^0,  \\
\frac{\D \vtheta_{\va,\beta}}{\D t}&=\mZ\vtheta_{\mW,\beta}, \quad\vtheta_{\va,\beta}(0)=\vtheta_{\va,\beta}^0,
\end{aligned}
\end{equation}   
reads
\begin{equation}
\begin{aligned}
\vtheta_{\mW,\beta}(t)&=\left(\sum_{k=1}^r\left[c_{\lambda_k,\mW,\beta}\exp(\lambda_kt)+d_{\lambda_k,\mW,\beta}\exp(-\lambda_kt)\right]\bm{v}_k\right)\\
&~~+\fP_{(r+1):(m^2+1)}\vtheta_{\mW,\beta}(0),\\
\vtheta_{\va,\beta}(t)&=\left(\sum_{k=1}^r\left[c_{\lambda_k,\va,\beta}\exp(\lambda_kt)+d_{\lambda_k,\va,\beta}\exp(-\lambda_kt)\right]\vu_k\right)\\
&~~+\fP_{(r+1):(W_1H_1)}\vtheta_{\va,\beta}(0).
\end{aligned}    
\end{equation}
\end{proposition}
\begin{remark}\label{rmk...Projection}
It is noteworthy that $\vtheta_{\mW,\beta}$ shall be understood as two components, one is 
the projection of  $\vtheta_{\mW,\beta}$ into  $\mathrm{span}\left\{\bm{v}_k\right\}_{k=1}^{r}$, 
\[
\fP_{1:r}\vtheta_{\mW,\beta}(t):=\left(\sum_{k=1}^r\left[c_{\lambda_k,\mW,\beta}\exp(\lambda_kt)+d_{\lambda_k,\mW,\beta}\exp(-\lambda_kt)\right]\bm{v}_k\right),
\]
which evolves with respect to time $t$, and the other is  the projection of  $\vtheta_{\mW,\beta}$ into  $\left(\mathrm{span}\left\{\bm{v}_k\right\}_{k=1}^{r}\right)^\perp=\mathrm{span}\left\{\bm{v}_k\right\}_{k=r+1}^{m^2+1}$,
\[
\fP_{(r+1):(m^2+1)}\vtheta_{\mW,\beta}(t)=\fP_{(r+1):(m^2+1)}\vtheta_{\mW,\beta}(0),
\]
which remains frozen as $t$ evolves.
\end{remark}
\subsection{Difference between Real  and Linear Dynamics}
For any $\beta\in[M]$,  the real dynamics \eqref{eq...text...2Layer...alpha=1Dynamics} can be written into  
\begin{align*}
\frac{\D {\mW}_{p,q,\beta}}{\D t}&=-\frac{1}{n}\sum_{i=1}^n e_i\cdot \left(\sum_{u=1}^{W_{1}} \sum_{v=1}^{H_{1}}\va_{u,v,\beta}\cdot\sigma^{(1)}\left(\eps\vx_{u, v, \beta}^{[1]}(i)\right)\cdot \vx_{u+p,v+q}(i)\right)\\
&~~-\frac{1}{n}\sum_{i=1}^n y_i\cdot \left(\sum_{u=1}^{W_{1}} \sum_{v=1}^{H_{1}}\va_{u,v,\beta}\cdot  \vx_{u+p,v+q}(i)\right)\\
&~~+\sum_{u=1}^{W_{1}} \sum_{v=1}^{H_{1}}{\va}_{u,v,\beta}\cdot \vz_{u+p,v+q},\\
\frac{\D \vb_{\beta}}{\D t}&=-\frac{1}{n}\sum_{i=1}^n e_i\cdot \left(\sum_{u=1}^{W_{1}} \sum_{v=1}^{H_{1}}\va_{u,v,\beta}\cdot\sigma^{(1)}\left(\eps\vx_{u, v, \beta}^{[1]}(i)\right)\right)\\
&~~-\frac{1}{n}\sum_{i=1}^n y_i\cdot \left(\sum_{u=1}^{W_{1}} \sum_{v=1}^{H_{1}}\va_{u,v,\beta}\right)\\
&~~+\sum_{u=1}^{W_{1}} \sum_{v=1}^{H_{1}} {\va}_{u,v,\beta}\cdot z,\\
 \frac{\D \va_{u,v,\beta}}{\D t}&=-\frac{1}{n}\sum_{i=1}^n e_i\cdot  \frac{\sigma\left(\eps
 \vx_{u,v,\beta}^{[1]}(i)\right)}{\eps}-\frac{1}{n}\sum_{i=1}^n y_i\cdot 
 \vx_{u,v,\beta}^{[1]}(i)\\
&~~+\left(\sum_{p=0}^{m-1} \sum_{q=0}^{m-1}  \vz_{u+p, v+q}  \cdot   {\mW}_{p,q,\beta}   \right) +  {\vb}_{\beta}\cdot z.
\end{align*}
Hence the difference between the real and linear dynamics is characterized by $\{f_{p,q,\beta}, f_\beta, g_{u,v,\beta}\}_{p,q\in[0:m-1],u\in[W_1], v\in[H_1],\beta\in[M]}$, where
\begin{align*}
f_{p,q,\beta}&:=\frac{1}{n}\sum_{i=1}^n e_i\cdot \left(\sum_{u=1}^{W_{1}} \sum_{v=1}^{H_{1}}\va_{u,v,\beta}\cdot\sigma^{(1)}\left(\eps\vx_{u, v, \beta}^{[1]}(i)\right)\cdot \vx_{u+p,v+q}(i)\right)\\
&~~+\frac{1}{n}\sum_{i=1}^n y_i\cdot \left(\sum_{u=1}^{W_{1}} \sum_{v=1}^{H_{1}}\va_{u,v,\beta}\cdot \vx_{u+p,v+q}(i)\right),\\
f_{\beta}&:=\frac{1}{n}\sum_{i=1}^n e_i\cdot \left(\sum_{u=1}^{W_{1}} \sum_{v=1}^{H_{1}}\va_{u,v,\beta}\cdot\sigma^{(1)}\left(\eps\vx_{u, v, \beta}^{[1]}(i)\right)\right)\\
&~~+\frac{1}{n}\sum_{i=1}^n y_i\cdot \left(\sum_{u=1}^{W_{1}} \sum_{v=1}^{H_{1}}\va_{u,v,\beta} \right),\\
g_{u,v,\beta}&:= \frac{1}{n}\sum_{i=1}^n e_i\cdot  \frac{\sigma\left(\eps
 \vx_{u,v,\beta}^{[1]}(i)\right)}{\eps}+\frac{1}{n}\sum_{i=1}^n y_i\cdot  
 \vx_{u,v,\beta}^{[1]}(i),
\end{align*}
and  for each $\beta\in[M]$, we set
\begin{align*}
\vf_{\beta}&:=\left(f_{0,0,\beta},f_{0,1,\beta},\cdots,f_{0,m-1,\beta}; f_{1,0,\beta},\cdots,f_{1,m-1,\beta};\cdots\cdots f_{m-1,m-1,\beta};f_{\beta}  \right)^\T,\\
\vg_{\beta}&:= \left({g}_{1,1,\beta}, {g}_{1,2,\beta},\cdots,{g}_{1,H_1,\beta};{g}_{2,1,\beta},\cdots,{g}_{2,H_1,\beta};\cdots\cdots{g}_{W_1,H_1,\beta}\right)^\T,
\end{align*}
and we observe further that for any $\beta\in[M]$, the real dynamics read
\begin{equation}\label{eq...text...Difference...RealDynamics}
\frac{\D \begin{pmatrix}\vtheta_{\mW,\beta}\\
					\vtheta_{\va,\beta}\end{pmatrix}}{\D t}=\begin{pmatrix}\vf_\beta\\
					\vg_\beta \end{pmatrix}+\mA\begin{pmatrix}\vtheta_{\mW,\beta}\\
					\vtheta_{\va,\beta}\end{pmatrix}.
\end{equation}
\begin{definition}[Neuron energy]
In real dynamics, we define the  energy at time $t$ for each $\beta\in[M]$, 
\begin{equation}
        E_\beta(t):=\left( \Norm{\vtheta_{\mW,\beta}(t)}_2^2+\Norm{\vtheta_{\va,\beta}(t)}_2^2\right)^{\frac{1}{2}},
\end{equation}
and we denote 
\begin{equation}
E_{\max}(t):=\max_{\beta\in[M]}E_\beta(t).
\end{equation}
\end{definition}
\noindent
For simplicity, we hereafter drop the $(t)$s for all $E_\beta(t)$ and $E_{\max}(t)$.  
Then the estimates on $\{\vf_\beta, \vg_\beta\}_{\beta=1}^M$ read
\begin{proposition}\label{prop...CellProblemEstimates}
For any $\eps>0$ and any time $t>0$,
\begin{equation}
\begin{aligned}
\Norm{\vf_{\beta}}_2&\leq \left(M\eps^2 E_{\max}^2+\eps E_{\max} \right) \Norm{\vtheta_{\va,\beta}}_2,\\
\Norm{\vg_{\beta}}_2&\leq \left(M\eps^2 E_{\max}^2+\eps E_{\max} \right) \Norm{\vtheta_{\mW,\beta}}_2.
\end{aligned}    
\end{equation}
Moreover,  we obtain that 
\begin{equation*}  \left(\Norm{\vf_{\beta}}_2+\Norm{\vg_{\beta}}_2^2\right)^{\frac{1}{2}}  \leq \left(M\eps^2 E_{\max}^2+\eps E_{\max} \right) E_\beta\leq \left(M\eps^2 E_{\max}^2+\eps E_{\max} \right) E_{\max}.
\end{equation*}
\end{proposition}
\begin{proof}
We obtain that for each $i\in[n]$,
\begin{align*}
\Abs{e_i+y_i}&=\Abs{\sum_{\beta=1}^{M}\sum_{u=1}^{W_1}\sum_{v=1}^{H_1} \eps\va_{u, v, \beta} \cdot\sigma\left(\eps \vx_{u, v, \beta}^{[1]}(i)\right)}\leq\eps\sum_{\beta=1}^{M}\Abs{ \sum_{u=1}^{W_1}\sum_{v=1}^{H_1} \va_{u, v, \beta} \cdot\sigma\left(\eps \vx_{u, v, \beta}^{[1]}(i)\right)}\\
&\leq\eps^2\sum_{\beta=1}^{M} \Norm{\vtheta_{\va,\beta}}_1\max_{u\in[W_1],v\in[H_1]}\Abs{\vx_{u, v, \beta}^{[1]}(i)}\\
&\leq \eps^2\sum_{\beta=1}^{M} \Norm{\vtheta_{\va,\beta}}_1\max_{u\in[W_1],v\in[H_1]}\Abs{\left(\sum_{p=0}^{m-1} \sum_{q=0}^{m-1} \vx_{u+p, v+q}(i) \cdot   {\mW}_{p,q,\beta}   \right) + {\vb}_{\beta}}\\
&\leq\eps^2 c\sum_{\beta=1}^{M} \Norm{\vtheta_{\va,\beta}}_1\Norm{\vtheta_{\mW,\beta}}_1\\
&\leq \eps^2c\sqrt{W_1H_1}\sqrt{m^2+1}\sum_{\beta=1}^{M}\Norm{\vtheta_{\va,\beta}}_2\Norm{\vtheta_{\mW,\beta}}_2,
\end{align*}
for simplicity we omit the constant  $ \sqrt{W_1H_1}\sqrt{m^2+1}$ since it is a universal constant.
Hence, we obtain further that 
\begin{align*}
\Norm{\vf_{\beta}}_2&\leq c\sqrt{m^2+1}\Abs{\frac{1}{n}\sum_{i=1}^n (e_i+y_i)\cdot \left(\sum_{u=1}^{W_{1}} \sum_{v=1}^{H_{1}}\va_{u,v,\beta}\cdot\sigma^{(1)}\left(\eps\vx_{u, v, \beta}^{[1]}(i)\right)\right)}\\
&~~+c\sqrt{m^2+1}\Abs{\frac{1}{n}\sum_{i=1}^n y_i\cdot \left(\sum_{u=1}^{W_{1}} \sum_{v=1}^{H_{1}}\va_{u,v,\beta}\cdot\left(\sigma^{(1)}\left(\eps\vx_{u, v, \beta}^{[1]}(i)\right)-1\right) \right)} \\
&\leq  \frac{c\sqrt{m^2+1}\sqrt{W_1H_1}}{n}\sum_{i=1}^n \left(\left(\sum_{\beta=1}^M \eps^2 \Norm{\vtheta_{\mW,\beta}}_2 \Norm{\vtheta_{\va,\beta}}_2 \right)\Norm{\vtheta_{\va,\beta}}_2\right)\\
&~~+ c^2\sqrt{m^2+1}\sqrt{W_1H_1} \Norm{ \vtheta_{\va,\beta}}_2 \Norm{\eps\vtheta_{\mW,\beta}}_2  \\
&\leq \left(M\eps^2 E_{\max}^2+\eps E_{\max} \right) \Norm{\vtheta_{\va,\beta}}_2, 
\end{align*}
where we also   omit the constant  $ \sqrt{W_1H_1}\sqrt{m^2+1}$, and similarly
\begin{align*}
\Norm{\vg_{\beta}}_2&\leq c\sqrt{W_1H_1}\max_{u\in[W_1], v\in[H_1]}\Abs{\frac{1}{n}\sum_{i=1}^n (e_i+y_i)\cdot\frac{\sigma\left(\eps
\vx_{u,v,\beta}^{[1]}(i)\right)}{\eps}}\\
&~~+c\sqrt{W_1H_1}\max_{u\in[W_1], v\in[H_1]}\Abs{\frac{1}{n}\sum_{i=1}^n y_i\cdot\left(\frac{\sigma\left(\eps
\vx_{u,v,\beta}^{[1]}(i)\right)}{\eps}-\vx_{u,v,\beta}^{[1]}(i)\right)}\\
&\leq  \frac{c\sqrt{m^2+1}\sqrt{W_1H_1}}{n}\sum_{i=1}^n \left(\left(\sum_{\beta=1}^M \eps^2 \Norm{\vtheta_{\mW,\beta}}_2 \Norm{\vtheta_{\va,\beta}}_2 \right)\Norm{\vtheta_{\mW,\beta}}_2\right)\\
&~~+ c^2\sqrt{m^2+1}\sqrt{W_1H_1}  \eps\Norm{\vtheta_{\mW,\beta}}_2^2  \\
&\leq \left(M\eps^2 E_{\max}^2+\eps E_{\max} \right) \Norm{\vtheta_{\mW,\beta}}_2.
\end{align*}
\end{proof}
\subsection{Several Estimate on the Initial Parameters}
We begin this part by an estimate on standard Gaussian vectors
\begin{lemma}[Bounds on initial parameters]\label{lemma..Initialization}
Given any $\delta\in(0,1)$, we have with probability at least $1-\delta$ over the choice of $\vtheta^0$,
\begin{equation} 
\max_{\beta\in[M]}\left\{\Norm{\vtheta_{\mW,\beta}^0}_{\infty}, \Norm{\vtheta_{\va,\beta}^0}_{\infty}\right \}\leq \sqrt{2\log\frac{2M(m^2+1+W_1H_1)}{\delta}}.
\end{equation}
\end{lemma}
\begin{proof}
If $\rX \sim \fN(0, 1)$, then   for any $\eta > 0$,
\[\Prob(\abs{\rX} > \eta) \leq 2\exp\left(-\frac{1}{2}\eta^2\right).\] 
 Since given any $\beta\in[M]$, for  each $u\in[W_1]$,  $v\in[H_1]$, $p\in[0:m-1]$ and $q\in[0:m-1]$,
\[	{\mW}^0_{p,q,\beta} \sim\fN(0, 1),\quad  {\vb}^0_{\beta}   \sim\fN(0, 1),\quad {\va}^0_{u, v, \beta}\sim\fN(0,1),\]
and they are all independent with each other. 
As we set
\begin{equation*}
\eta = \sqrt{2\log\frac{2M(m^2+1+W_1H_1)}{\delta}},
\end{equation*}
we obtain that
\begin{align*}
&~~\Prob\left(\max_{\beta\in[M]}\left\{\Norm{\vtheta_{\mW,\beta}^0}_{\infty}, \Norm{\vtheta_{\va,\beta}^0}_{\infty}\right \}>\eta\right)\\
& = \Prob\left(\max_{\beta\in[M],u\in[W_1],v\in[H_1],p\in[0:m-1],q\in[0:m-1]}\left\{\Abs{{\mW}^0_{p,q,\beta}}, \Abs{{\vb}^0_{\beta}},\Abs{{\va}^0_{u, v, \beta}}\right \}>\eta\right)  \\
& \leq 2M(m^2+1) \exp\left( -\frac{1}{2}\eta^2\right) + 2MW_1H_1 \exp\left( -\frac{1}{2}\eta^2 \right)\\
& = 2M(m^2+1+W_1H_1)\exp\left({-\frac{1}{2}\eta^2}\right) = \delta.
\end{align*}
\end{proof}
\noindent
Next we would like to introduce the sub-exponential norm~\cite{Vershynin2010Introduction} of a random variable and  Bernstein's Inequality.
\begin{definition}[Sub-exponential norm]
The sub-exponential norm of a random variable $\rX$ is defined as
    \begin{equation}
        \norm{\rX}_{\psi_1} := \inf\left\{s>0 \mid \Exp_{\rX}\left[\exp\left(\frac{\abs{\rX}}{s}\right)\right]\leq 2\right\}.
    \end{equation}
    In particular, we denote $\rX:=\chi^2(d)$ as   a   chi-square distribution with  $d$ degrees of freedom~\cite{Laurent2000AdaptiveEstimationQuadratic}, and its 
   sub-exponential norm by \[C_{\psi,d}:=\norm{\rX}_{\psi_1}.\] 
\end{definition}

\begin{remark}
As the  probability density function of  $\rX=\chi^2(d)$ reads
    \begin{equation*}
        f_{\rX}(z):=\frac{1}{2^{\frac{d}{2}}\Gamma(\frac{d}{2})}z^{\frac{d}{2}-1}\exp\left({-\frac{z}{2}}\right),\quad z\geq 0,
    \end{equation*}
    we note that
\begin{align*}
        \Exp_{\rX\sim\chi^2(1)}\exp\left(\frac{\abs{\rX}}{s}\right)
         & =\int_0^{+\infty}\frac{1}{2^{\frac{1}{2}}\Gamma(\frac{1}{2})}z^{-\frac{1}{2}}\exp\left(-\left({\frac{1}{2}-\frac{1}{s}}\right)z\right)\diff{z}=\frac{1}{\sqrt{1-\frac{2}{s}}},
\end{align*}
Then we obtain that \[\frac{8}{3}\leq C_{\psi,1}<3.\] Moreover, we notice that 
\begin{align*}
        \Exp_{\rX\sim\chi^2(d)}\exp\left(\frac{\abs{\rX}}{s}\right)
         & =\left(\Exp_{\rY\sim\chi^2(1)}\exp\left(\frac{\abs{\rY}}{s}\right)\right)^d,
\end{align*}
as we set 
\begin{align*}
       \frac{1}{\sqrt{1-\frac{2}{s}}}=2^{\frac{1}{d}},
\end{align*}
then \[s=\frac{2}{1-2^{-\frac{2}{d}}},\]
hence 
\[\frac{2}{1-2^{-\frac{2}{d}}}\leq C_{\psi,d}<3,\]
and 
\[ C_{\psi,d}\geq C_{\psi,1},\]
for $d\geq 1$.
\end{remark}
\begin{theorem}[Bernstein's inequality]\label{thm...BernsteinInequality}
Let $\{\rX_k\}_{k=1}^m$ be i.i.d.\ sub-exponential random variables satisfying \[\Exp\rX_1=\mu,\] then for any $\eta\geq 0$, we have
\begin{equation*}
\Prob\left(\Abs{\frac{1}{m}\sum_{k=1}^m\rX_k-\mu}\geq \eta\right)\leq 2\exp\left(-C_0 m \min\left(\frac{\eta^2}{\norm{\rX_1}^2_{\psi_1}},\frac{\eta}{\norm{\rX_1}_{\psi_1}}\right)\right),
\end{equation*}
for some absolute constant $C_0$.
\end{theorem}
\noindent
In order to study the phenomenon  of condensation, we need to concatenate the vectors $\left\{ \vtheta_{\mW,\beta}\right\}_{\beta=1}^M$ into
\[
\vtheta_{\mW}:=\mathrm{vec}\left(\left\{ \vtheta_{\mW,\beta}\right\}_{\beta=1}^M\right),
\]
and we obtain that 
\begin{proposition}[Upper and lower bounds of initial parameters]\label{prop..UpperBoundandLowerBoundInitial}
Given any $\delta\in(0,1)$,   if
    \[M=\Omega\left(\log \frac{2}{\delta} \right),\]
then with probability at least $1-\delta$ over the choice of $\vtheta^0$,
\begin{align}
\sqrt{\frac{M(m^2+1)}{2}} & \leq \Norm{\vtheta_{\mW}^0}_2\leq \sqrt{\frac{3M(m^2+1)}{2}}.\label{append...eqgroup..InitialWNorm}     \end{align}
\end{proposition}
\begin{proof}
Since given any $\beta\in[M]$, for  each   $p\in[0:m-1]$ and $q\in[0:m-1]$,
\[\left(\mW_{p,q,\beta}^0\right)^2,  \quad\left(\vb_{\beta}^0\right)^2\sim\chi^2(1)\] 
are  sub-exponential  
random variables   with 
\[\Exp\left(\mW_{p,q,\beta}^0\right)^2=1,\quad\Exp\left(\vb_{\beta}^0\right)^2=1.\] 
 Since $C_{\psi,1}\geq \frac{8}{3}>2$, then  for any $0\leq \eta\leq 2$, it is obvious that
\[
\min\left(\frac{\eta^2}{C^2_{\psi,1}},\frac{\eta}{C_{\psi,1}}\right)=\frac{\eta^2}{C^2_{\psi,1}}.
\]
Hence, by application of  \Cref{thm...BernsteinInequality},
\begin{align*}
&\Prob\left(\Abs{\frac{1}{M(m^2+1)}\sum_{\beta=1}^M\left[\left(\sum_{p=0}^{m-1}\sum_{q=0}^{m-1}\left(\mW_{p,q,\beta}^0\right)^2\right)+\left(\vb_\beta\right)^2\right]-1}\geq \eta\right)\\
&~~\leq 2\exp\left(-\frac{C_0 M(m^2+1)  \eta^2}{C^2_{\psi,1}} \right),
\end{align*}
as we  set 
\[
2\exp\left(-\frac{C_0 M(m^2+1)  \eta^2}{C^2_{\psi,1}} \right)= {\delta},
\]
and consequently,
\[
\eta=\sqrt{\frac{C^2_{\psi,1}}{   C_0M(m^2+1)  }\log\frac{2}{\delta}},
\]
then  with probability at least $1- {\delta} $ over the choice of $\vtheta^0$,
\begin{align*}
   &\Norm{\vtheta_{\mW}^0}^2_2\geq M(m^2+1)\left(1-\sqrt{\frac{C^2_{\psi,1}}{   C_0M(m^2+1)  }\log\frac{2}{\delta}}\right), \\
   &\Norm{\vtheta_{\mW}^0}^2_2\leq M(m^2+1)\left(1+\sqrt{\frac{C^2_{\psi,1}}{   C_0M(m^2+1)  }\log\frac{2}{\delta}}\right),
\end{align*}
and by choosing 
\[
M\geq{\frac{4C^2_{\psi,1}}{ C_0(m^2+1) }\log\frac{2}{\delta}},
\]
we obtain that 
\[ \sqrt{\frac{M(m^2+1)}{2}}
\leq \Norm{\vtheta_{\mW}^0}_2\leq \sqrt{\frac{3M(m^2+1)}{2}}.\]
\end{proof}
\subsection{Lower Bound on Effective Time}
We denote a useful quantity
\begin{equation} 
\phi(t):= \sup_{0\leq s\leq t} {E}_{\max}(s),
\end{equation} 
then directly from \Cref{lemma..Initialization}, we have with probability at least $1-\delta$ over the choice of $\vtheta^0$,
\begin{equation*}
\max_{\beta\in[M]}\left\{\Norm{\vtheta_{\mW,\beta}^0}_{\infty}, \Norm{\vtheta_{\va,\beta}^0}_{\infty}\right \}\leq \sqrt{2\log\frac{2M(m^2+1+W_1H_1)}{\delta}},
\end{equation*}
hence
\begin{equation} 
 \phi(0)\leq\sqrt{2(m^2+1+W_1H_1)\log\frac{2M(m^2+1+W_1H_1)}{\delta}}.
\end{equation}
We define
\begin{equation}
T_{\mathrm{eff}}: = \inf\left\{t>0 \mid M\eps^2\phi^3(t)>M^{-\tau},\quad \tau=\frac{\gamma-1}{4}\right\},
\end{equation}
then for $M$ large enough,
\begin{align*}
M\eps^2\phi^3(0)&\leq  M\eps^2\left({2(m^2+1+W_1H_1)\log\frac{2M(m^2+1+W_1H_1)}{\delta}}\right)^{\frac{3}{2}} \leq M^{-\frac{\gamma-1}{2}},
\end{align*}
hence $T_{\mathrm{eff}}\geq 0$.
We observe further that as the real dynamics read
\begin{equation} 
\frac{\D \begin{pmatrix}\vtheta_{\mW,\beta}\\
					\vtheta_{\va,\beta}\end{pmatrix}}{\D t}=\begin{pmatrix}\vf_\beta\\
					\vg_\beta \end{pmatrix}+\mA\begin{pmatrix}\vtheta_{\mW,\beta}\\
					\vtheta_{\va,\beta}\end{pmatrix}.
\end{equation}
then by taking the $2$-norm on both sides
\begin{align*}
E_\beta(t)&\leq\exp\left(t\Norm{\mA}_{2\to 2}\right)E_\beta(0)\\
&~~+\int_{0}^t\exp\left((t-s)\Norm{\mA}_{2\to 2}\right)\left(M\eps^2E_{\max}^2(s)+\eps E_{\max}(s)\right)E_\beta(s)\D s,     
\end{align*}
by taking supreme over the  index $\beta$ and  time $0\leq t\leq T_{\mathrm{eff}}$   on both sides,   for $M$ large enough,  
\begin{equation}
\begin{aligned}
\phi(t)&\leq \phi(0)\exp(\lambda_1 t)+ 2M^{-\min\{1,\tau\}}\int_0^t\exp(\lambda_1 (t-s))\D s \\ 
&\leq \phi(0)\exp(\lambda_1t)+2M^{-\min\{1,\tau\}}\frac{\exp(\lambda_1t)-1}{\lambda_1}\\
&\leq \phi(0)\exp(\lambda_1t)+2M^{-\min\{1,\tau\}}\frac{\exp(\lambda_1t)}{\lambda_1},
\end{aligned}  
\end{equation}
then based on \Cref{lemma..Initialization}, with probability $1-\delta$ over the choice of $\vtheta^0$, for sufficiently large $M$,
\begin{equation}
\begin{aligned}
\phi(t)  
&\leq \phi(0)\exp(\lambda_1t)+\frac{2}{\lambda_1}M^{-\min\{1,\tau\}}\exp(\lambda_1t)\\
&\leq 2\phi(0)\exp(\lambda_1t)\\
&\leq 2\sqrt{2(m^2+1+W_1H_1)\log\frac{2M(m^2+1+W_1H_1)}{\delta}}\exp(\lambda_1t),  
\end{aligned}  
\end{equation}
we set $t_0$ as the time satisfying
\begin{equation}
2\sqrt{2(m^2+1+W_1H_1)\log\frac{2M(m^2+1+W_1H_1)}{\delta}}\exp(\lambda_1t) =\frac{1}{2} M^{\frac{\gamma-1}{4}},  
\end{equation}
then we obtain that, for any $\eta_0>\frac{\gamma-1}{100}>0$,
\begin{equation}
T_{\mathrm{eff}}\geq t_0>\frac{1}{\lambda_1}\left[ \log\left(\frac{1}{4}\right)+\left({\frac{\gamma-1}{4}}-\eta_0\right)\log(M)\right].
\end{equation} 
Recall that
\[
\vtheta_{\mW}=\mathrm{vec}\left(\left\{ \vtheta_{\mW,\beta}\right\}_{\beta=1}^M\right),
\]
and we denote further that 
\[
\vtheta_{\mW,\bm{v}_1}:=\fP_{1}\vtheta_{\mW}:= \left( \left<\vtheta_{\mW,1},\bm{v}_1\right>, \left<\vtheta_{\mW,2},\bm{v}_1\right>, \cdots \left<\vtheta_{\mW,M},\bm{v}_1\right>\right)^\T,
\]
where $\bm{v}_1$ is the  eigenvector of the largest eigenvalue of $\mZ^\T\mZ$, or the first column vector of $\mV$ in \eqref{eq...text...E-LinearDynamics....SVD}.

\begin{theorem}\label{thm..Condense}
Given any $\delta\in(0,1)$, under \Cref{Assumption....ActivationFunctions}, \Cref{assumption...Data}, \Cref{assump...LimitExistence} and \Cref{assump...SpectralGap}, if $\gamma> 1$,
then	with probability at least $1-\delta$ over the choice of $\vtheta^0$,
we have  
\begin{equation}\label{eq...thm...Condense...PartOne}
\lim_{M\to+\infty} \sup\limits_{t\in[0,T_{\mathrm{eff}}]} \frac{\Norm{\vtheta_{\mW}(t)-\vtheta_{\mW}(0)}_2}{\Norm{\vtheta_{\mW}(0)}_2}=+\infty,
\end{equation}
and
\begin{equation}\label{eq...thm...Condense...PartTwo}
\lim_{M\to+\infty} \sup\limits_{t\in[0,T_{\mathrm{eff}}]}\frac{\Norm{\vtheta_{\mW, \bm{v}_1}(t) }_2}{\Norm{\vtheta_{\mW}(t)}_2} =1.
\end{equation}
\end{theorem}
\begin{proof}
Since   for each $\beta\in[M]$, the real dynamics read
\begin{align*}
\frac{\D \begin{pmatrix} \vtheta_{\mW,\beta}\\
\vtheta_{\va,\beta} \end{pmatrix}}{\D t}
=\mA\begin{pmatrix} \vtheta_{\mW,\beta}\\
\vtheta_{\va,\beta} \end{pmatrix}
+
\begin{pmatrix}\vf_\beta\\
\vg_\beta \end{pmatrix},\quad  \begin{pmatrix} \vtheta_{\mW,\beta}(0)\\
\vtheta_{\va,\beta} (0)\end{pmatrix}= \begin{pmatrix} \vtheta_{\mW,\beta}^0\\
\vtheta_{\va,\beta}^0 \end{pmatrix},
\end{align*}
and  
\begin{align*}
\begin{pmatrix} \vtheta_{\mW,\beta}\\
\vtheta_{\va,\beta} \end{pmatrix}
&=\exp\left(t\mA\right)\begin{pmatrix} \vtheta_{\mW,\beta}^0\\
\vtheta_{\va,\beta}^0 \end{pmatrix}
+\int_{0}^t\exp\left((t-s)\mA\right)
\begin{pmatrix}\vf_\beta\\
\vg_\beta \end{pmatrix}\D s.    
\end{align*}
As we notice that for any $\beta\in[M]$, $\begin{pmatrix} \vtheta_{\mW,\beta}\\
\vtheta_{\va,\beta} \end{pmatrix}$ 
can be written into two parts,  the first one is the linear part, the second one is the residual part. For simplicity of proof, we need to introduce some further notations, i.e., as we denote 
\begin{align*}
\begin{pmatrix} \Bar{\vtheta}_{\mW,\beta}\\
\Bar{\vtheta}_{\va,\beta} \end{pmatrix}&:=\exp\left(t\mA\right)\begin{pmatrix} \vtheta_{\mW,\beta}^0\\
\vtheta_{\va,\beta}^0 \end{pmatrix},\\
\begin{pmatrix} \widetilde{\vtheta}_{\mW,\beta}\\
\widetilde{\vtheta}_{\va,\beta} \end{pmatrix}&:=\int_{0}^t\exp\left((t-s)\mA\right)
\begin{pmatrix}\vf_\beta\\
\vg_\beta \end{pmatrix}\D s,
\end{align*}
then
\[
\begin{pmatrix} \vtheta_{\mW,\beta}\\
\vtheta_{\va,\beta} \end{pmatrix}=\begin{pmatrix} \Bar{\vtheta}_{\mW,\beta}\\
\Bar{\vtheta}_{\va,\beta} \end{pmatrix}+\begin{pmatrix} \widetilde{\vtheta}_{\mW,\beta}\\
\widetilde{\vtheta}_{\va,\beta}\end{pmatrix},
\]
directly from \Cref{prop...LinearODESolution}, we obtain that 
\begin{equation}
\begin{aligned}
\Bar{\vtheta}_{\mW,\beta}(t)&=\left(\sum_{k=1}^r\left[c_{\lambda_k,\mW,\beta}\exp(\lambda_kt)+d_{\lambda_k,\mW,\beta}\exp(-\lambda_kt)\right]\bm{v}_k\right)\\
&~~+\fP_{(r+1):(m^2+1)}\vtheta_{\mW,\beta}^0,\\
\Bar{\vtheta}_{\va,\beta}(t)&=\left(\sum_{k=1}^r\left[c_{\lambda_k,\va,\beta}\exp(\lambda_kt)+d_{\lambda_k,\va,\beta}\exp(-\lambda_kt)\right]\vu_k\right)\\
&~~+\fP_{(r+1):(W_1H_1)}\vtheta_{\va,\beta}^0.
\end{aligned}    
\end{equation}
We are hereby to prove \eqref{eq...thm...Condense...PartOne}.  Firstly, we observe that 
\[
{\vtheta}_{\mW}(0)=\Bar{\vtheta}_{\mW}(0),
\]
hence
\begin{align*}
&\Norm{\vtheta_{\mW}(t)-\vtheta_{\mW}(0)}_2^2\\
=&\Norm{\vtheta_{\mW}(t)-\Bar{\vtheta}_{\mW}(0)}^2_2\\     
=&\Norm{\fP_{1:r}\left(\vtheta_{\mW}(t)-\Bar{\vtheta}_{\mW}(0)\right)+\fP_{(r+1):(m^2+1)}\left(\vtheta_{\mW}(t)-\Bar{\vtheta}_{\mW}(0)\right) }_2^2\\
=&\Norm{\fP_{1:r}\left(\vtheta_{\mW}(t)-\Bar{\vtheta}_{\mW}(0)\right)}^2_2+\Norm{\fP_{(r+1):(m^2+1)}\left(\vtheta_{\mW}(t)-\Bar{\vtheta}_{\mW}(0)\right)}^2_2\\
=&\Norm{\fP_{1:r}\left(\Bar{\vtheta}_{\mW}(t)-\Bar{\vtheta}_{\mW}(0)\right)+\fP_{1:r}\widetilde{\vtheta}_{\mW}(t)}^2_2+\Norm{\fP_{(r+1):(m^2+1)} \widetilde{\vtheta}_{\mW}(t) }^2_2,
\end{align*} 
 by choosing $\eta_0=\frac{\gamma-1}{8}$, then  for time $0\leq t\leq \bar{t}_0:=\frac{1}{\lambda_1}\left[\left({\frac{\gamma-1}{8}}\right)\log(M)-\log(2)\right]$ and any $\beta\in[M]$,
\begin{align*}
\Norm{\begin{pmatrix} \widetilde{\vtheta}_{\mW,\beta}\\
\widetilde{\vtheta}_{\va,\beta} \end{pmatrix}}_2&=\Norm{\int_{0}^{t}\exp\left((t-s)\mA\right)
 \begin{pmatrix}\vf_k\\
 \vg_k \end{pmatrix}
 \D s}_2\\
&\leq \left(M\eps^2\phi^3(t)+  \eps\phi^2(t)\right) \int_{0}^{t}\exp(\lambda_1(t-s))  \D s\\
&\leq 2M^{-\min\left\{\tau, \frac{1}{2}\right\}}\int_{0}^{t}\exp(\lambda_1(t-s)) \D s\\
&\leq 2M^{-\frac{\gamma-1}{4}}  \frac{\exp(\lambda_1t)}{\lambda_1}\leq 2M^{-\frac{\gamma-1}{4}}  \exp(\lambda_1\bar{t}_0)= M^{-\frac{\gamma-1}{8}}.
\end{align*}
We  conclude that for $t\leq \bar{t}_0$, the following holds
\begin{align*}
\Norm{\widetilde{\vtheta}_{\mW}(t)}_2&\leq \sqrt{M}\Norm{\begin{pmatrix} \widetilde{\vtheta}_{\mW,\beta}\\
\widetilde{\vtheta}_{\va,\beta} \end{pmatrix}}_2\leq\sqrt{M} M^{-\frac{\gamma-1}{8}},
\end{align*}
thus the ratio reads
\begin{align*}
&~~\left(\frac{\Norm{{\vtheta}_{\mW}(t)-{\vtheta}_{\mW}(0)}_2}{\Norm{{\vtheta}_{\mW}(0)}_2}\right)^2\\
&=\frac{\Norm{\fP_{1:r}\left(\Bar{\vtheta}_{\mW}(t)-\Bar{\vtheta}_{\mW}(0)\right)+\fP_{1:r}\widetilde{\vtheta}_{\mW}(t)}^2_2+\Norm{\fP_{(r+1):(m^2+1)} \widetilde{\vtheta}_{\mW}(t) }^2_2}{\Norm{{\vtheta}_{\mW}(0)}_2^2}\\
&=\underbrace{\frac{\Norm{\fP_{1:r}\left(\Bar{\vtheta}_{\mW}(t)-\Bar{\vtheta}_{\mW}(0)\right)+\fP_{1:r}\widetilde{\vtheta}_{\mW}(t)}^2_2}{\Norm{{\vtheta}_{\mW}(0)}_2^2}}_{\textrm{I}}+\underbrace{\frac{\Norm{\fP_{(r+1):(m^2+1)} \widetilde{\vtheta}_{\mW}(t) }^2_2}{\Norm{{\vtheta}_{\mW}(0)}_2^2}}_{\textrm{II}}.   
\end{align*}
For part II, we obtain that 
\[
\frac{\Norm{\fP_{(r+1):(m^2+1)} \widetilde{\vtheta}_{\mW}(t) }_2}{\Norm{{\vtheta}_{\mW}(0)}_2}\leq \frac{\Norm{  \widetilde{\vtheta}_{\mW}(t) }_2}{\Norm{{\vtheta}_{\mW}(0)}_2},
\]
then directly from \Cref{prop..UpperBoundandLowerBoundInitial}, with probability at least $1-\delta$ over the choice of $\vtheta^0$ and   large enough $M$,  for any $0\leq t \leq \bar{t}_0=\frac{1}{\lambda_1}\left[\left({\frac{\gamma-1}{8}}\right)\log(M)-\log(2)\right]$, the following holds:
\begin{align*}
 \frac{\Norm{\fP_{(r+1):(m^2+1)} \widetilde{\vtheta}_{\mW}(t) }_2}{\Norm{{\vtheta}_{\mW}(0)}_2}\leq \frac{\Norm{  \widetilde{\vtheta}_{\mW}(t) }_2}{\Norm{{\vtheta}_{\mW}(0)}_2}\leq \sqrt{\frac{2}{m^2+1}}M^{-\frac{\gamma-1}{8}},
\end{align*}
by taking  the limit, we obtain that 
\[
\lim_{M\to\infty}\sup_{t\in[0,\bar{t}_0]}\frac{\Norm{\fP_{(r+1):(m^2+1)} \widetilde{\vtheta}_{\mW}(t) }_2}{\Norm{{\vtheta}_{\mW}(0)}_2}=0.
\]
As for part I, we notice that 
\begin{align*}
&\frac{\Norm{\fP_{1:r}\left(\Bar{\vtheta}_{\mW}(t)-\Bar{\vtheta}_{\mW}(0)\right)+\fP_{1:r}\widetilde{\vtheta}_{\mW}(t)}_2}{\Norm{{\vtheta}_{\mW}(0)}_2}\\
&\geq \frac{\Norm{\fP_{1:r}\left(\Bar{\vtheta}_{\mW}(t)-\Bar{\vtheta}_{\mW}(0)\right)}_2-\Norm{\fP_{1:r}\widetilde{\vtheta}_{\mW}(t)}_2}{\Norm{{\vtheta}_{\mW}(0)}_2}  \\
&\geq \underbrace{\frac{\Norm{\fP_{1:r}\left(\Bar{\vtheta}_{\mW}(t)-\Bar{\vtheta}_{\mW}(0)\right)}_2 }{\Norm{{\vtheta}_{\mW}(0)}_2}}_{\textrm{III}}-\underbrace{\frac{ \Norm{\widetilde{\vtheta}_{\mW}(t)}_2}{\Norm{{\vtheta}_{\mW}(0)}_2}}_{\textrm{IV}},
\end{align*}
by similar reasoning as shown in part II, for any time $t\in[0,\bar{t}_0]$, part IV tends to zero as $M\to\infty$, i.e.,
\[
\lim_{M\to\infty}\sup_{t\in[0,\bar{t}_0]}\frac{\Norm{\fP_{1:r} \widetilde{\vtheta}_{\mW}(t) }_2}{\Norm{{\vtheta}_{\mW}(0)}_2}\leq \lim_{M\to\infty}\sup_{t\in[0,\bar{t}_0]}\frac{ \Norm{\widetilde{\vtheta}_{\mW}(t)}_2}{\Norm{{\vtheta}_{\mW}(0)}_2}=0.
\]
For part III, we observe that since
\begin{align*}
&\Norm{\fP_{1:r}\left(\Bar{\vtheta}_{\mW}(t)-\Bar{\vtheta}_{\mW}(0)\right)}_2^2\\
=&\sum_{\beta=1}^M\sum_{k=1}^r \left[c_{\lambda_k,\mW,\beta}\left(\exp(\lambda_kt)-1\right)+d_{\lambda_k,\mW,\beta}\left(\exp(-\lambda_kt)-1\right)\right]^2,
\end{align*}
where
\begin{align*}
c_{\lambda_k,\mW,\beta}&=\frac{1}{2}\left(\left<\vtheta_{\mW,\beta}^0, \bm{v}_k\right>+\left<\vtheta_{\va,\beta}^0, \bm{u}_k\right>\right),\\
d_{\lambda_k,\mW,\beta}&=\frac{1}{2}\left(\left<\vtheta_{\mW,\beta}^0, \bm{v}_k\right>-\left<\vtheta_{\va,\beta}^0, \bm{u}_k\right>\right),
\end{align*}
we observe that given $\vu_k$ and $\bm{v}_k$,  $\rY_{k,\beta}:=\left<\vtheta_{\va,\beta}^0, \bm{u}_k\right>\sim \fN(0, 1)$ and  $\rX_{k,\beta}:\left<\vtheta_{\mW,\beta}^0, \bm{v}_k\right> \sim \fN(0, 1)$. Moreover,    $\left\{\rX_{k,\beta}\right\}_{\beta=1}^M\sim \fN(0, 1)$ and  $\left\{\rY_{k,\beta}\right\}_{\beta=1}^M\sim \fN(0, 1)$  are i.i.d.\ Gaussian variables, and they are independent with each other. 
We denote further that $r_k(t):=\exp\left(\frac{1}{2}\lambda_kt\right)$, and  by application of \Cref{thm...BernsteinInequality}, with probability $1-\frac{\delta}{4}$ over the choice of $\vtheta^0$, for $M$ large enough,
\[
{\frac{(m^2+1)}{2}}
\leq \frac{1}{M}\Norm{\vtheta_{\mW}(0)}_2^2\leq  {\frac{3 (m^2+1)}{2}},
\]
and with probability $1-\frac{\delta}{4}$ over the choice of $\vtheta^0$, for $M$ large enough,
\[
\frac{1}{2}\leq\frac{1}{M}\sum_{\beta=1}^M\rX^2_{k,\beta}\leq \frac{3 }{2},
\]
and with probability $1-\frac{\delta}{4}$ over the choice of $\vtheta^0$, for $M$ large enough,
\[
\frac{1}{2}\leq\frac{1}{M}\sum_{\beta=1}^M\rY^2_{k,\beta}\leq \frac{3 }{2},
\]
and with probability $1-\frac{\delta}{4}$ over the choice of $\vtheta^0$, for $M$ large enough,
\[
-\frac{1}{4}\leq\frac{1}{M}\sum_{\beta=1}^M\rX_{k,\beta}\rY_{k,\beta}\leq\frac{1}{4},
\]
then we obtain that, with probability at least $1-\delta$ over the choice of $\vtheta^0$
\begin{align*}
&\frac{1}{M}\Norm{\fP_{1:r}\left(\Bar{\vtheta}_{\mW}(t)-\Bar{\vtheta}_{\mW}(0)\right)}_2^2\\
=&\frac{1}{M}\sum_{\beta=1}^M \sum_{k=1}^r\left[c_{\lambda_k,\mW,\beta}\left(\exp(\lambda_kt)-1\right)+d_{\lambda_k,\mW,\beta}\left(\exp(-\lambda_kt)-1\right)\right]^2\\
=& \frac{1}{4M}\sum_{\beta=1}^M \sum_{k=1}^r \left[\rX_{k,\beta}\left(r_k^2(t)+r_k^{-2}(t)-2\right)+\rY_{k,\beta}\left(r_k^2(t)-r_k^{-2}(t)\right)\right]^2\\
=&\frac{1}{4M}\sum_{\beta=1}^M \sum_{k=1}^r \left(r_k(t)-r_k^{-1}(t)\right)^2\left[\rX_{k,\beta}\left(r_k(t)-r_k^{-1}(t)\right)+\rY_{k,\beta}\left(r_k(t)+r_k^{-1}(t)\right)\right]^2\\
\geq &\frac{1}{8}\sum_{k=1}^r \left(r_k(t)-r_k^{-1}(t)\right)^2\left(r_k^2(t)+3r_k^{-2}(t)\right) \geq  \frac{1}{8}\sum_{k=1}^r \left(r_k(t)-r_k^{-1}(t)\right)^4. 
\end{align*}
Hence,  with probability at least $1-\delta$ over the choice of $\vtheta^0$ and   large enough $M$,  for any $0\leq t \leq \bar{t}_0=\frac{1}{\lambda_1}\left[\left({\frac{\gamma-1}{8}}\right)\log(M)-\log(2)\right]$, 
\begin{align*}
 &\frac{1}{M}\Norm{\fP_{1:r}\left(\Bar{\vtheta}_{\mW}(t)-\Bar{\vtheta}_{\mW}(0)\right)}_2^2 \\
\geq &  \frac{1}{8}\sum_{k=1}^r \left(r_k(\bar{t}_0)-r_k^{-1}(\bar{t}_0)\right)^4  \geq   \frac{1}{8}\sum_{k=1}^r \left(r_k(\bar{t}_0)-1\right)^4\\
\gtrsim &\frac{1}{8}\sum_{k=1}^r \exp\left(\frac{4}{2}\frac{\lambda_k}{\lambda_1}\left({\frac{\gamma-1}{8}}\right)\log(M)\right)=\frac{1}{8}\sum_{k=1}^rM^{\frac{\lambda_k}{\lambda_1}\frac{\gamma-1}{4}},
\end{align*}
then for part III, we obtain that with probability at least $1-\delta$ over the choice of $\vtheta^0$ and   large enough $M$,  for any $0\leq t \leq \bar{t}_0=\frac{1}{\lambda_1}\left[\left({\frac{\gamma-1}{8}}\right)\log(M)-\log(2)\right]$, 
\begin{align*}
\frac{\Norm{\fP_{1:r}\left(\Bar{\vtheta}_{\mW}(\bar{t}_0)-\Bar{\vtheta}_{\mW}(0)\right)}_2^2 }{\Norm{{\vtheta}_{\mW}(0)}_2^2}&=\frac{\frac{1}{M}\Norm{\fP_{1:r}\left(\Bar{\vtheta}_{\mW}(\bar{t}_0)-\Bar{\vtheta}_{\mW}(0)\right)}_2^2 }{\frac{1}{M}\Norm{{\vtheta}_{\mW}(0)}_2^2} \\
&\geq{\frac{2}{3 (m^2+1)}}{\frac{1}{M}\Norm{\fP_{1:r}\left(\Bar{\vtheta}_{\mW}(t)-\Bar{\vtheta}_{\mW}(0)\right)}_2^2} \\
&\gtrsim \frac{2}{3 (m^2+1)}\frac{1}{8}\sum_{k=1}^rM^{\frac{\lambda_k}{\lambda_1}\frac{\gamma-1}{4}},
\end{align*}
by taking  the limit, we obtain that 
\[
\lim_{M\to\infty}\sup_{t\in[0,\bar{t}_0]}\frac{\Norm{\fP_{1:r}\left(\Bar{\vtheta}_{\mW}(\bar{t}_0)-\Bar{\vtheta}_{\mW}(0)\right)}_2}{\Norm{{\vtheta}_{\mW}(0)}_2}=\infty.
\]
To sum up, since $\bar{t}_0\leq T_{\mathrm{eff}}$, we have that 
\begin{equation}
\lim_{m\to+\infty} \sup\limits_{t\in[0,T_{\mathrm{eff}}]} \frac{\Norm{\vtheta_{\mW}(t)-\vtheta_{\mW}(0)}_2}{\Norm{\vtheta_{\mW}(0)}_2}=+\infty+0=+\infty,
\end{equation}
which finishes the proof of \eqref{eq...thm...Condense...PartOne}.

In order to prove \eqref{eq...thm...Condense...PartTwo}, firstly we have
\[
\frac{\Norm{\vtheta_{\mW, \bm{v}_1}(t) }_2}{\Norm{\vtheta_{\mW}(t)}_2}\leq 1,
\]
moreover, we observe that 
\begin{align*}
\left(\frac{\Norm{\vtheta_{\mW, \bm{v}_1}(t) }_2}{\Norm{\vtheta_{\mW}(t)}_2}\right)^2&=\frac{\Norm{\vtheta_{\mW, \bm{v}_1}(t) }_2^2}{\Norm{\vtheta_{\mW}(t)}_2^2}=\frac{\Norm{\vtheta_{\mW, \bm{v}_1}(t) }_2^2}{\Norm{\fP_{1:r}\vtheta_{\mW}(t)}_2^2+\Norm{\fP_{(r+1):(m^2+1)}\vtheta_{\mW}(t)}_2^2}\\
&=\frac{\Norm{\Bar{\vtheta}_{\mW, \bm{v}_1}(t)+\widetilde{\vtheta}_{\mW, \bm{v}_1}(t)}^2_2}{\Norm{\fP_{1:r}\left(\Bar{\vtheta}_{\mW}(t)+\widetilde{\vtheta}_{\mW}(t)\right)}_2^2+\Norm{\fP_{(r+1):(m^2+1)}\left(\Bar{\vtheta}_{\mW}(t)+\widetilde{\vtheta}_{\mW}(t)\right)}_2^2},
\end{align*}
where 
\begin{align*}
\Bar{\vtheta}_{\mW, \bm{v}_1}(t)&:=\fP_1  \Bar{\vtheta}_{\mW}(t),\\  
\widetilde{\vtheta}_{\mW, \bm{v}_1}(t)&:=\fP_1  \widetilde{\vtheta}_{\mW}(t).
\end{align*}
Then  with probability at least $1-\delta$ over the choice of $\vtheta^0$ and   large enough $M$,  for any $0\leq t \leq \bar{t}_0=\frac{1}{\lambda_1}\left[\left({\frac{\gamma-1}{8}}\right)\log(M)-\log(2)\right]$, the following holds:
\begin{align*}
\Norm{\Bar{\vtheta}_{\mW, \bm{v}_1}(t)+\widetilde{\vtheta}_{\mW, \bm{v}_1}(t)}_2&\geq \Norm{\Bar{\vtheta}_{\mW, \bm{v}_1}(t)-\Bar{\vtheta}_{\mW, \bm{v}_1}(0)+ \widetilde{\vtheta}_{\mW, \bm{v}_1}(t)}_2-   \Norm{\Bar{\vtheta}_{\mW, \bm{v}_1}(0)}_2\\ 
&\geq \Norm{\Bar{\vtheta}_{\mW, \bm{v}_1}(t)-\Bar{\vtheta}_{\mW, \bm{v}_1}(0)}_2-\Norm{\widetilde{\vtheta}_{\mW, \bm{v}_1}(t)}_2-   \Norm{\Bar{\vtheta}_{\mW, \bm{v}_1}(0)}_2\\
&\geq \underbrace{\Norm{\Bar{\vtheta}_{\mW, \bm{v}_1}(t)-\Bar{\vtheta}_{\mW, \bm{v}_1}(0)}_2}_{\textrm{V}}-\Norm{\widetilde{\vtheta}_{\mW}(t)}_2-   \Norm{\Bar{\vtheta}_{\mW, \bm{v}_1}(0)}_2\\
&\geq  \sqrt{\frac{M}{8}} \left(r_1(t)-{r^{-1}_1(t)}\right)^2-\sqrt{M} M^{-\frac{\gamma-1}{8}}-\sqrt{\frac{3M}{2}},
\end{align*}
hence part V is the term of dominance,
and by similar reasoning
\begin{align*}
\Norm{\fP_{1:r}\left(\Bar{\vtheta}_{\mW}(t)+\widetilde{\vtheta}_{\mW}(t)\right)}_2&\geq \underbrace{\Norm{\fP_{1:r}\left(\Bar{\vtheta}_{\mW}(t)-\Bar{\vtheta}_{\mW}(0)\right)}_2}_{\textrm{VI}}-\Norm{\widetilde{\vtheta}_{\mW}(t)}_2-   \Norm{\fP_{1:r}\Bar{\vtheta}_{\mW}(0)}_2\\
&\geq  \sqrt{\frac{M}{8}\sum_{k=1}^r \left(r_k(t)-r_k^{-1}(t)\right)^4}-\sqrt{M} M^{-\frac{\gamma-1}{8}}-\sqrt{\frac{3Mr}{2}},
\end{align*}
hence part VI is the term of dominance, and finally
\begin{align*}
&\Norm{\fP_{(r+1):(m^2+1)}\left(\Bar{\vtheta}_{\mW}(t)+\widetilde{\vtheta}_{\mW}(t)\right)}_2\\
\leq& \Norm{\fP_{(r+1):(m^2+1)}\left(\Bar{\vtheta}_{\mW}(t)-\Bar{\vtheta}_{\mW}(0)\right)}_2+\Norm{\widetilde{\vtheta}_{\mW}(t)}_2 +\Norm{\fP_{(r+1):(m^2+1)}\Bar{\vtheta}_{\mW}(0)}_2\\
=& \Norm{\widetilde{\vtheta}_{\mW}(t)}_2 +\Norm{\fP_{(r+1):(m^2+1)}\Bar{\vtheta}_{\mW}(0)}_2 \leq \sqrt{M} M^{-\frac{\gamma-1}{8}}+\sqrt{\frac{3M(m^2+1)}{2}},
\end{align*}
which is at most of order $\sqrt{M}$. Then for $M$ large enough, the majority part of  the ratio 
$  
\frac{\Norm{\vtheta_{\mW, \bm{v}_1}(t) }_2^2}{\Norm{\vtheta_{\mW}(t)}_2^2}
$ 
is 
\[
\frac{\Norm{\Bar{\vtheta}_{\mW, \bm{v}_1}(t)-\Bar{\vtheta}_{\mW, \bm{v}_1}(0)}_2^2}{\Norm{\fP_{1:r}\left(\Bar{\vtheta}_{\mW}(t)-\Bar{\vtheta}_{\mW}(0)\right)}_2^2},
\]
where
\begin{align*}
&\frac{1}{M}\Norm{\Bar{\vtheta}_{\mW, \bm{v}_1}(t)-\Bar{\vtheta}_{\mW, \bm{v}_1}(0)}_2^2\\
=& \frac{1}{M}\sum_{\beta=1}^M  \left[c_{\lambda_1,\mW,\beta}\left(\exp(\lambda_1t)-1\right)+d_{\lambda_1,\mW,\beta}\left(\exp(-\lambda_1t)-1\right)\right]^2 \\
=& \frac{\exp(2\lambda_1t)}{M}\sum_{\beta=1}^M \left[c_{\lambda_1,\mW,\beta}\left(1-\exp(-\lambda_1t)\right)+d_{\lambda_1,\mW,\beta}\left(\exp(-2\lambda_1t)-\exp(-\lambda_1t)\right)\right]^2,\\
\text{and}\\
&\frac{1}{M}\Norm{\fP_{1:r}\left(\Bar{\vtheta}_{\mW}(t)-\Bar{\vtheta}_{\mW}(0)\right)}_2^2\\
=& \frac{1}{M}\sum_{\beta=1}^M \sum_{k=1}^r \left[c_{\lambda_k,\mW,\beta}\left(\exp(\lambda_kt)-1\right)+d_{\lambda_k,\mW,\beta}\left(\exp(-\lambda_kt)-1\right)\right]^2 \\
=& \frac{\exp(2\lambda_1t)}{M}\sum_{\beta=1}^M \sum_{k=1}^r \Big[c_{\lambda_k,\mW,\beta}\left(\exp((\lambda_1-\lambda_k)t)-\exp(-\lambda_1t)\right)\\
&~~~~~~~~~~~~~~~~~~~~~~~~+d_{\lambda_1,\mW,\beta}\left(\exp(-(\lambda_1+\lambda_k)t)-\exp(-\lambda_1t)\right)\Big]^2.
\end{align*}
By taking $t=\bar{t}_0$, we observe that  as  the spectral gap $\Delta \lambda>0$, then for any $k\in[2:r]$,
\begin{equation} \label{equ:...spectral gap}
\exp((\lambda_1-\lambda_k)\bar{t}_0)\leq \exp(-\Delta \lambda\bar{t}_0)\lesssim M^{-\left(\frac{\gamma-1}{8}\right)\frac{\Delta \lambda}{\lambda_1}},
\end{equation}
which tends to zero as $M\to\infty$, hence
\begin{align*}
&\lim_{M\to\infty}\frac{\Norm{\Bar{\vtheta}_{\mW, \bm{v}_1}(\bar{t}_0)-\Bar{\vtheta}_{\mW, \bm{v}_1}(0)}_2^2}{\Norm{\fP_{1:r}\left(\Bar{\vtheta}_{\mW}(\bar{t}_0)-\Bar{\vtheta}_{\mW}(0)\right)}_2^2}   \\
=&\lim_{M\to\infty}\frac{\frac{1}{M}\Norm{\Bar{\vtheta}_{\mW, \bm{v}_1}(\bar{t}_0)-\Bar{\vtheta}_{\mW, \bm{v}_1}(0)}_2^2}{\frac{1}{M}\Norm{\fP_{1:r}\left(\Bar{\vtheta}_{\mW}\widetilde{\vtheta}-\Bar{\vtheta}_{\mW}(0)\right)}_2^2}\\
=&\lim_{M\to\infty}\frac{\frac{\exp(2\lambda_1\bar{t}_0)}{M}\sum_{\beta=1}^M  c_{\lambda_1,\mW,\beta}^2 }{\frac{\exp(2\lambda_1\bar{t}_0)}{M}\sum_{\beta=1}^M  c_{\lambda_1,\mW,\beta}^2 +\underbrace{0+0+\cdots+0}_{r-1~\text{zeros}}}=1,
\end{align*}
and in combination with  $\bar{t}_0\leq T_{\mathrm{eff}}$, 
we finish the proof of \eqref{eq...thm...Condense...PartTwo}.
\end{proof}
\section{Two-Layer CNNs with Multi Channels}
In the case of two-layer CNNs with multi channels, as $L=2$,  then we still set  ${\mW}_{p,q,\alpha,\beta}:= {\mW}_{p,q,\alpha,\beta}^{[1]}$,  $\vb_{\beta}:=\vb_{\beta}^{[1]}$, and $\vx_{u+p,v+q,\alpha}(i):=\vx^{[0]}_{u+p,v+q,\alpha}(i)$ for simplicity, then the GD dynamics reads
\begin{align*}
\frac{\D {\mW}_{p,q,\alpha,\beta}}{\D t}&=-\frac{1}{n}\sum_{i=1}^n e_i\cdot \left(\sum_{u=1}^{W_{1}} \sum_{v=1}^{H_{1}}\va_{u,v,\beta}\cdot\sigma^{(1)}\left(\vx_{u, v, \beta}^{[1]}(i)\right)\cdot \vx_{u+p,v+q,\alpha}(i)\right),\\
\frac{\D \vb_{\beta}}{\D t}&=-\frac{1}{n}\sum_{i=1}^n e_i\cdot \left(\sum_{u=1}^{W_{1}} \sum_{v=1}^{H_{1}}\va_{u,v,\beta}\cdot\sigma^{(1)}\left(\vx_{u, v, \beta}^{[1]}(i)\right)\right),\\
 \frac{\D \va_{u,v,\beta}}{\D t}&=-\frac{1}{n}\sum_{i=1}^n e_i\cdot  \sigma\left(
 \vx_{u,v,\beta}^{[1]}(i)\right).
\end{align*}     
since $e_i\approx -y_i$, and   by means of perturbation
expansion with respect to $\eps$ and keep the order $1$ term, we obtain that
\begin{equation}
\begin{aligned}
\frac{\D {\mW}_{p,q,\alpha,\beta}}{\D t}&\approx\frac{1}{n}\sum_{i=1}^n y_i\cdot \left(\sum_{u=1}^{W_{1}} \sum_{v=1}^{H_{1}}{\va}_{u,v,\beta}\cdot \vx_{u+p,v+q,\alpha}(i)\right),\\
\frac{\D  {\vb}_{\beta}}{\D t}&\approx \frac{1}{n}\sum_{i=1}^n y_i\cdot  \sum_{u=1}^{W_{1}} \sum_{v=1}^{H_{1}} {\va}_{u,v,\beta},\\
 \frac{\D  {\va}_{u,v,\beta}}{\D t}&\approx\frac{1}{n}\sum_{i=1}^n y_i\cdot   {\vx}_{u, v, \beta}^{[1]}(i)\\
 &=\frac{1}{n}\sum_{i=1}^n y_i\cdot   \left[\left(\sum_{p=0}^{m-1} \sum_{q=0}^{m-1} \vx_{u+p, v+q}(i) \cdot   {\mW}_{p,q,\alpha,\beta}   \right) + {\vb}_{\beta}\right].
\end{aligned}  
\end{equation}
Given any $u\in[W_1]$, $v\in[H_1]$ and $\alpha\in[C_0]$, then for all $p, q\in[0:m-1]$, we set
\begin{equation}\label{eq...text...E-LinearDynamics...SumofVectors}
\begin{aligned}
\vz_{u+p,v+q,\alpha}&:=  \frac{1}{n}\sum_{i=1}^n y_i \vx_{u+p,v+q,\alpha}(i),\\
z&:=\frac{1}{n}\sum_{i=1}^n y_i,
\end{aligned}
\end{equation}
then the above dynamics   can be further simplified into: For any   $\beta\in[M]$,
\begin{equation}
\begin{aligned}
\frac{\D {\mW}_{p,q,\alpha,\beta}}{\D t}&\approx \sum_{u=1}^{W_{1}} \sum_{v=1}^{H_{1}}{\va}_{u,v,\beta}\cdot \vz_{u+p,v+q,\alpha},\\
\frac{\D  {\vb}_{\beta}}{\D t}&\approx   \sum_{u=1}^{W_{1}} \sum_{v=1}^{H_{1}} {\va}_{u,v,\beta}\cdot z,\\
 \frac{\D  {\va}_{u,v,\beta}}{\D t}&\approx \sum_{\alpha=1}^{C_0} \left(\sum_{p=0}^{m-1} \sum_{q=0}^{m-1}  \vz_{u+p, v+q, \alpha}  \cdot   {\mW}_{p,q,\alpha,\beta}   \right) + {\vb}_{\beta}\cdot z.
\end{aligned}  
\end{equation}
We observe that   the training  dynamics still takes the form
\begin{equation}
   \frac{\D \vtheta_\beta}{\D t}=\mA\vtheta_\beta,  
\end{equation}
except that in this case, 
\begin{align*}
\vtheta_\beta:=\Big(&\mW_{0,0,1,\beta},\mW_{0,1,1,\beta},\cdots,\mW_{0,m-1,1,\beta}; \mW_{1,0,1,\beta},\cdots,\mW_{1,m-1,1,\beta};\cdots\cdots\mW_{m-1,m-1,1,\beta};\\
&\mW_{0,0,2,\beta},\mW_{0,1,2,\beta},\cdots,\mW_{0,m-1,2,\beta}; \mW_{1,0,2,\beta},\cdots,\mW_{1,m-1,2,\beta};\cdots\cdots\mW_{m-1,m-1,2,\beta};\\
&\cdots\cdots\cdots\cdots\cdots\cdots\cdots\cdots\cdots\cdots\cdots\cdots\cdots\cdots\cdots\cdots\cdots\cdots\cdots\cdots\cdots\cdots\cdots\cdots\cdots\\
&\mW_{0,0,C_0,\beta},\mW_{0,1,C_0,\beta},\cdots,\mW_{0,m-1,C_0,\beta};  \cdots,\mW_{1,m-1,C_0,\beta};\cdots\cdots\mW_{m-1,m-1,C_0,\beta};\vb_{\beta};\\
&{\va}_{1,1,\beta}, {\va}_{1,2,\beta},\cdots,{\va}_{1,H_1,\beta};{\va}_{2,1,\beta},\cdots,{\va}_{2,H_1,\beta};\cdots\cdots{\va}_{W_1,H_1,\beta}\Big)^\T,
\end{align*} 
or in more simplified notations,
\begin{align*}
\vtheta_\beta:=\Big(&\mW_{0,0:(m-1),1,\beta}; \mW_{1,0:(m-1),1,\beta};\cdots\cdots\mW_{m-1,0:(m-1),1,\beta};\\
&\mW_{0,0:(m-1),2,\beta}; \mW_{1,0:(m-1),2,\beta};\cdots\cdots\mW_{m-1,0:(m-1),2,\beta};\\
&\cdots\cdots\cdots\cdots\cdots\cdots\cdots\cdots\cdots\cdots\cdots\cdots\cdots\cdots\cdots\cdots \\
&\mW_{0,0:(m-1),C_0,\beta};   \mW_{1,0:(m-1),C_0,\beta};\cdots\cdots\mW_{m-1,0:(m-1),C_0,\beta};\vb_{\beta};\\
&{\va}_{1,1:H_1,\beta};{\va}_{2,1:H_1,\beta};\cdots\cdots{\va}_{W_1,1:H_1,\beta}\Big)^\T,
\end{align*} 
and 
\begin{equation}
    \mA:=\left[\begin{array}{cc}
\mzero_{(C_0 m^2+1)\times (C_0m^2+1)} & {\mZ}^\T \\
\mZ & \mzero_{W_1H_1\times W_1H_1} 
\end{array}\right],
\end{equation}
where $\mZ\in\sR^{W_1H_1\times (m^2+1)}$ and $\mZ$ depends sorely on the input samples  $\{ \vx_i\}_{i=1}^n$ and
$\{ y_i\}_{i=1}^n$, whose entries read
\begin{align}
 \left[\begin{array}{ccccccccc }
\vz_{1,1:m,1};&\cdots &\vz_{m,1:m,1};&\vz_{1,1:m,2};&\cdots &\vz_{m,1:m,2};&\cdots\cdots&\vz_{m,1:m,C_0};&z\\
\vz_{1,2:(m+1),1};&\cdots &\vz_{m,2:(m+1),1};&\vz_{1,2:(m+1),2};&\cdots &\vz_{m,2:(m+1),2};&\cdots\cdots&\vz_{m,2:(m+1),C_0};&z\\
\vdots&\cdots &\vdots &\vdots&\cdots &\vdots&\cdots\cdots&\vdots&\vdots\\
\vz_{1,H_1:H_0,1};&\cdots &\vz_{m,H_1:H_0,1};&\vz_{1,H_1:H_0,2} &\cdots &\vz_{m,H_1:H_0,2};&\cdots\cdots&\vz_{m,H_1:H_0,C_0};&z\\
\vz_{2,1:m,1};&\cdots &\vz_{m+1,1:m,1};&\vz_{2,1:m,2};&\cdots &\vz_{m+1,1:m,2};&\cdots\cdots&\vz_{m+1,1:m,C_0};&z\\
\vdots&\cdots &\vdots &\vdots&\cdots &\vdots&\cdots\cdots&\vdots&\vdots\\
\vz_{2,H_1:H_0,1};&\cdots &\vz_{m+1,H_1:H_0,1};&\vz_{2,H_1:H_0,2} &\cdots &\vz_{m+1,H_1:H_0,2};&\cdots\cdots&\vz_{m+1,H_1:H_0,C_0};&z\\
\vdots&\cdots &\vdots &\vdots&\cdots &\vdots&\cdots\cdots&\vdots&\vdots\\
\vz_{W_1,H_1:H_0,1};&\cdots &\vz_{W_0,H_1:H_0,1};&\vz_{W_1,H_1:H_0,2} &\cdots &\vz_{W_0,H_1:H_0,2};&\cdots\cdots&\vz_{W_0,H_1:H_0,C_0};&z\\
\end{array}\right].
\label{multichannel_matrix}
\end{align}
We remark that all results in the case of single channel CNNs can be reproduced for multi channel CNNs, and we state a theorem without proof 
\begin{theorem}\label{thm..Condense....rep}
Given any $\delta\in(0,1)$, under \Cref{Assumption....ActivationFunctions}, \Cref{assumption...Data}, \Cref{assump...LimitExistence} and \Cref{assump...SpectralGap}, if $\gamma> 1$,
then	with probability at least $1-\delta$ over the choice of $\vtheta^0$,
we have  
\begin{equation} 
\lim_{M\to+\infty} \sup\limits_{t\in[0,T_{\mathrm{eff}}]} \frac{\Norm{\vtheta_{\mW}(t)-\vtheta_{\mW}(0)}_2}{\Norm{\vtheta_{\mW}(0)}_2}=+\infty,
\end{equation}
and
\begin{equation} 
\lim_{M\to+\infty} \sup\limits_{t\in[0,T_{\mathrm{eff}}]}\frac{\Norm{\vtheta_{\mW, \bm{v}_1}(t) }_2}{\Norm{\vtheta_{\mW}(t)}_2} =1.
\end{equation}
\end{theorem}
\noindent
To sum up, the weight vectors condense toward the unit eigenvector equipped with the largest singular value.

\end{document}